\PassOptionsToPackage{sort&compress}{natbib}
\documentclass{article}

\usepackage[dvipsnames,svgnames,x11names, table]{xcolor}
\usepackage[final]{corl_2026} 

\usepackage{todonotes}
\usepackage{soul}
\usepackage{xspace}
\usepackage{amsmath}
\usepackage{amssymb}
\usepackage{mdframed}
\usepackage{verbatim}
\usepackage{wrapfig}
\usepackage{colortbl}
\usepackage{array}
\usepackage{pifont}
\usepackage{booktabs}
\usepackage{graphicx}
\usepackage[font=footnotesize]{caption}

\usepackage{multirow}
\usepackage{longtable}
\usepackage{float} 
\usepackage{arydshln} 
\usepackage{tikz}
\usetikzlibrary{positioning,arrows.meta,calc,fit,backgrounds}
\definecolor{claudeorange}{HTML}{D97757} 
\usepackage[font=footnotesize]{caption}
\usepackage{subcaption} 
\newcommand{\pms}[2]{#1{\,\tiny\color{black!55}$\pm$#2}}

\newcommand{\promptfile}[1]{%
  \par\smallskip\noindent\textcolor{black!40}{\rule{\linewidth}{0.4pt}}\par\nobreak
  {\scriptsize\verbatiminput{#1}}%
  \nobreak\noindent\textcolor{black!40}{\rule{\linewidth}{0.4pt}}\par\smallskip}

\makeatletter
\renewcommand{\paragraph}[1]{\noindent\textbf{#1}}
\makeatother

\definecolor{flodarkpurple}{rgb}{0,0,0}

\newcommand{\algname}{\textsc{Direct}\xspace}
\makeatletter
\newif\ifpublic
\if@conferencefinal\publictrue\fi
\if@preprinttype\publictrue\fi
\makeatother

\begin{document}

\title{\algname: When and Where Should You Allocate Test-Time Compute in Embodied Planners?}

\makeatletter
\DeclareRobustCommand\onedot{\futurelet\@let@token\@onedot}
\def\@onedot{\ifx\@let@token.\else.\null\fi\xspace}
\newcommand{\eg}{\textit{e.g.},\xspace}
\newcommand{\ie}{\textit{i.e.},\xspace}
\newcommand{\cf}{\textit{cf}\onedot}
\newcommand{\etc}{\textit{etc}\onedot}
\newcommand{\wrt}{\textit{w.r.t}\onedot}
\newcommand{\iid}{\textit{i.i.d}\onedot}
\makeatother

\author{
    \bfseries
    Jadelynn Dao\textsuperscript{1,\ding{228}},\,\,
    Milan Ganai\textsuperscript{1,\ding{228}},\,\,
    Yasmina Abukhadra\textsuperscript{1},\,\,
    Ajay Sridhar\textsuperscript{1},\\ \bfseries
    Mozhgan Nasr Azadani\textsuperscript{1,2},\,\,
    Katie Luo\textsuperscript{1},\\ \bfseries 
    Clark Barrett\textsuperscript{1},\,\,
    Jiajun Wu\textsuperscript{1},\,\, 
    Chelsea Finn\textsuperscript{1},\,\,
    Marco Pavone\textsuperscript{1,3}\\\vspace{-6pt}\\
    \textcolor{flodarkpurple}{\textsuperscript{1}Stanford University, \textsuperscript{2}University of Waterloo, \textsuperscript{3}NVIDIA}
    \vspace{-8pt}
}

\maketitle

\ifpublic
\begingroup
\renewcommand\thefootnote{\textsuperscript{\ding{228}}}
\footnotetext{Equal contribution.}
\renewcommand\thefootnote{\textsuperscript{}}
\phantomsection\footnotetext{Correspondence to: \texttt{\{jadelynn, mganai\}@cs.stanford.edu}}
\endgroup
\fi


\begin{figure*}[h]
    \vspace{-15pt}
    \centering
    \includegraphics[width=\linewidth]{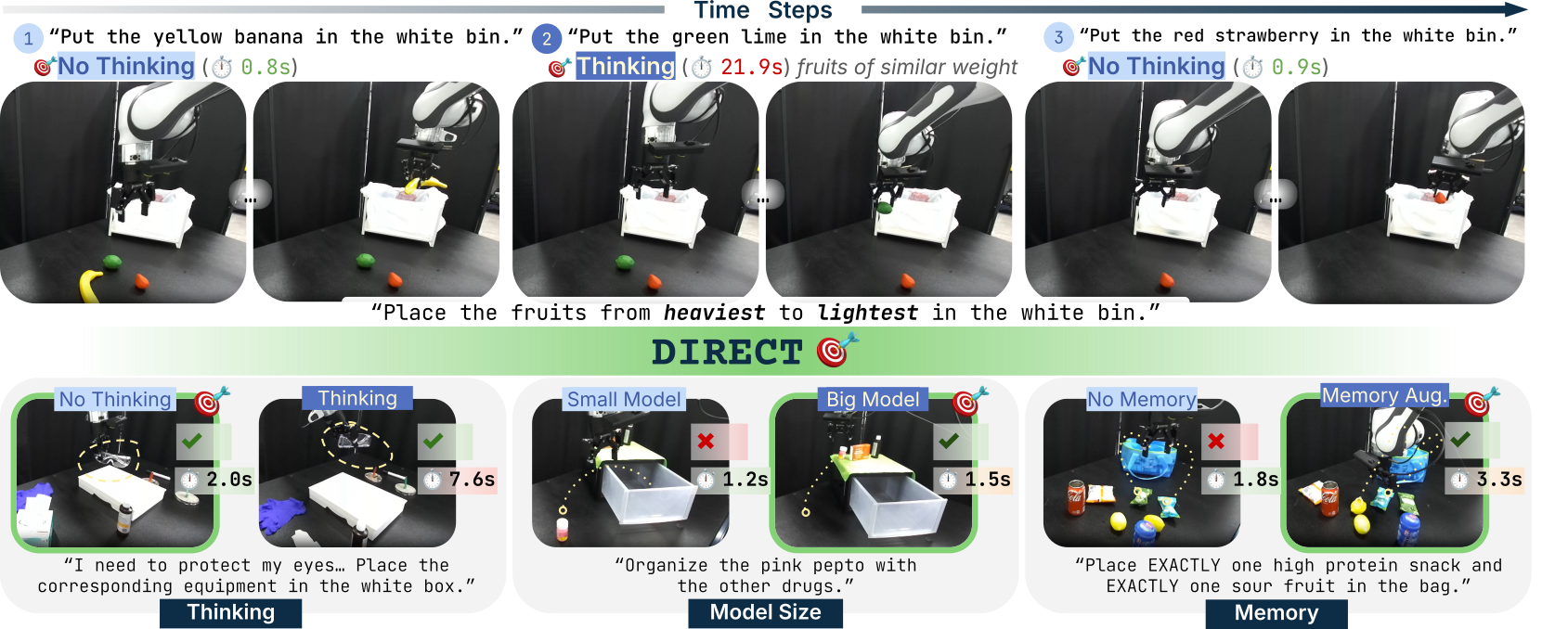}
    \vspace{-8pt}
    \caption{\textbf{\algname routes test-time compute per task, matching frontier planning at a fraction of the cost.} \textbf{Top:} on a multi-step task, \algname uses a non-thinking planner on trivial steps ($0.8$--$0.9$s) and escalates to thinking ($21.9$s) on ambiguous steps. \textbf{Bottom:} across three axes---chain-of-thought, model size, and memory---\algname routes to the cheapest yet capable model per task.}
    \label{fig:hero}
    \vspace{-12pt}
\end{figure*}

\begin{abstract} 
Vision-Language Models (VLMs) are increasingly deployed as high-level planners for embodied agents, with an emerging strategy of scaling test-time compute to improve capability. However, we observe that doing so increases latency, token usage, and FLOPs while yielding uneven, often diminishing gains in downstream success, limiting where embodied agents can be deployed. We argue that choosing when and where to spend test-time compute is central to bringing frontier performance to the real world, and we can budget inference at the planning level. We introduce \algname, a routing framework that uses multimodal scene context to allocate compute at each step of a closed-loop agent, selecting a planner for every subtask, improving the success--cost Pareto frontier over fixed model selection. Across three dominant scaling axes, namely chain-of-thought depth, model size, and memory history, our experiments on VLABench and RoboMME show that test-time compute is not a uniform lever: different axes yield qualitatively distinct capability gains. We validate these insights on a physical Franka arm in a DROID setup spanning zero-shot manipulation and long-horizon chaining, where our router matches or exceeds a stronger model's success rate at up to 65\% lower average latency. Ultimately, our results show that naively scaling test-time compute is wasteful, and  \algname can provide frontier-level embodied planning at a fraction of the cost. Project page can be found at \href{https://jadee-dao.github.io/direct/}{jadee-dao.github.io/direct/}.

\end{abstract}


\vspace{-5pt}
\section{Introduction}
\vspace{-5pt}

Vision-Language Models (VLMs) are increasingly deployed as high-level planners that decompose abstract instructions into executable sub-skills for embodied agents~\cite{Ahn2022SayCan, zitkovich2023rt2, driess2023plame, shi2025hirobot}, enabling semantic flexibility for open-world robotic interaction. In this hierarchy, an orchestrating VLM planner generates skill plans that are enacted by a low-level policy. An emerging strategy to improve performance is scaling compute cost expended at inference~\cite{Ahn2022SayCan, team2025gemini, sridhar2025memer}--this cost is known as test-time compute~\cite{snell2024scaling}.

But this scaling for embodied planners creates a growing friction with real-world viability: frontier VLM latencies can reach tens of seconds. In Fig.~\ref{fig:hero} (bottom left) on the Franka, a Thinking model is $3.8\times$ slower, yet has the same success as its non-thinking counterpart. The inference cost compounds as many planning calls are chained over long-horizon tasks, narrowing where they can be deployed. So, deciding \emph{when and where} to spend compute at each decision point is critical for robot systems. Model routing has emerged as one such approach for language models~\cite{Chen2023FrugalGPT, Ong2024RouteLLM} by directing queries to the cheapest capable model. In embodied planning, compute is not uniform--different techniques confer distinct capabilities, like deeper reasoning and longer-horizon memory. Effective allocation requires routing policies that infer task demands from scene and instruction context and match them to the compute configuration with the best quality–cost tradeoff.

We propose \algname: \textbf{D}ynamic \textbf{I}nference \textbf{R}outer for \textbf{E}mbodied \textbf{C}ompute \textbf{T}radeoffs, a routing framework over VLM planners for hierarchical embodied planning. Our investigations show scaling test-time compute is not a uniform lever: CoT depth helps on tasks with implicit semantic, physical, or spatial constraints; VLM size governs breadth of skills a planner can command; and memory-oriented compute helps on history-dependent tasks but can hurt elsewhere. Building on this, we develop a lightweight router that conditions on multimodal context, directing each task to the planner whose capability profile matches inferred demands. Our contributions are: (1) diagnostic analyses of task difficulty, planner capability, and test-time compute tradeoffs in embodied settings; (2) our framework \algname that dynamically allocates test-time compute for embodied planning; and (3) experiments on VLABench, RoboMME, and Franka DROID hardware showing \algname can match the strongest model's performance at up to 65\% lower latency.

\vspace{-5pt}
\section{Related Works}
\label{sec:citations}
\vspace{-5pt}

\paragraph{Hierarchical Vision-Language-Action Models.}
A growing line of work positions vision-language models as high-level planners that decompose abstract instructions into sub-skills for a low-level policy. Early work grounded LLM reasoning in robotic affordances \cite{Ahn2022SayCan, Huang2022InnerMonologue}, while recent hierarchical VLAs adopt an explicit System 1–System 2 decomposition~\cite{kahneman2011thinking}, where a high-level VLM emits semantic subtasks that a low-level policy enacts~\cite{shi2025hirobot, black2025pi05, li2025hamster, sridhar2025memer}. Variants explore skill-centric scheduling~\cite{robomatrix} and dedicated task-management VLMs~\cite{lohomanip, han2025robocerebralargescalebenchmarklonghorizon}. Existing methods commit to a single high-level model with fixed capability and cost. We instead treat the high-level planner as a dynamic variable, dispatching to different VLMs based on each task's inferred cognitive demands.

\paragraph{Dynamic Model Routing for Language and Multimodal Models.}
Dynamic routing optimizes inference by directing queries to the most efficient model in an ensemble. Frameworks like FrugalGPT \cite{Chen2023FrugalGPT} and RouteLLM \cite{Ong2024RouteLLM} predict performance from linguistic features \cite{ding2024hybridllm, aggarwal2024automix}. However, these frameworks route purely on text. A nascent multimodal line \cite{huang2025vlrouterbenchbenchmarkvisionlanguagemodel, ma2026mmr, liu2026adaptive, tang2025ecvl} shows that text features alone underestimate routing-relevant complexity. In embodied AI, difficulty further involves physical scene understanding and trajectory feasibility~\cite{Zhang_2025_ICCV}. The closest analog in robotics, DA-SIP \cite{chun2026dynamic}, routes \emph{within} a low-level diffusion controller, selecting integration-step budgets by difficulty. We extend multimodal routing to the high-level planning stack, where a broad ecosystem of interchangeable candidate models span orders of magnitude in compute cost and capability.

\paragraph{Test-Time Compute and Adaptive Inference.}
Scaling test-time compute can improve capability, even outperforming a 14$\times$ larger model \cite{snell2024scaling}, via CoT prompting \cite{Wei2022CoT}, deliberate search \cite{yao2023tree, bi2024forest}, and parallel sampling with majority voting or verifier-guided selection \cite{wang2022self, zhao2025sample}. Yet longer chains can hurt performance \cite{thinkingoptimal, dontoverthink} and saturate early \cite{wang2025entropy}. In embodied settings, grounded multi-step reasoning improves VLA success \cite{zawalski2024robotic,zhao2025cot,ganai2026self}, but its latency is prohibitive on a robot~\cite{chen2025training}; recent work amortizes this cost via thought reuse \cite{duan2025fast} or latent reasoning \cite{liu2026last, bai2026latent, Chen25-ecot-lite}. Scaling up the memory history length of the high-level planner has also been shown to improve performance on long-horizon memory-based tasks \cite{sridhar2025memer}. SayCan \cite{Ahn2022SayCan} shows scaling model size of the high-level LLM planner can improve task success, but performance saturates at some point. Since these strategies are typically applied uniformly \cite{Huang2022InnerMonologue, zawalski2024robotic}, we instead ask \emph{when} compute is worth its cost, identifying high-signal cases via perception-aware classifiers conditioned on scene and instruction context.

\vspace{-10pt}

\section{Investigating How Test-Time Compute Impacts Embodied Planning}
\label{sec:casestudies}
\vspace{-5pt}

Scaling test-time compute is not free in robotics: in a hierarchical stack the planner is invoked repeatedly, so per-call costs from CoT, larger models, or denser memory compound can multiply base model latencies. Cutting compute uniformly mitigates latency but risks failure on harder tasks. We therefore study where compute actually yields gains across three axes---reasoning depth, model size, and memory history---motivating our framework in Section~\ref{sec:method}. We study each of these axes in isolation to attribute gains to a single, qualitatively distinct form of compute.

\vspace{-7pt}
\subsection{Spending Thinking Tokens: Depth vs. Delay.}
\vspace{-5pt}
\begin{wrapfigure}{r}{0.55\textwidth}
\vspace{-7pt}
    \centering
    \includegraphics[width=\linewidth]{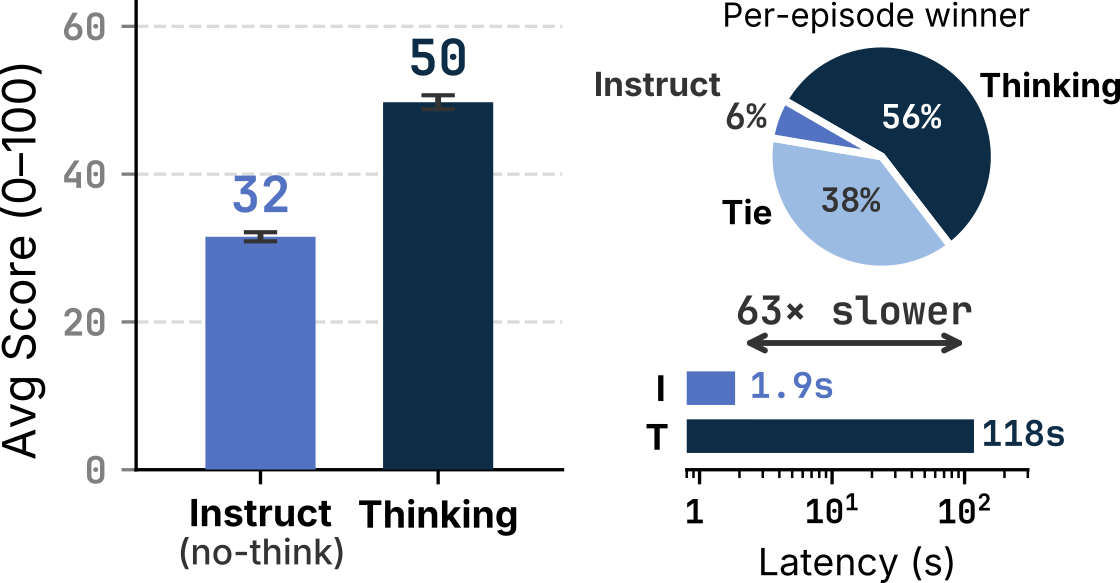}
    \vspace{-15pt}
    \caption{\footnotesize Across VLABench~\cite{Zhang_2025_ICCV}, Thinking significantly outperforms Instruct. However, 44\% of cases Qwen3-VL 8B Instruct matches/surpasses Thinking at $<$2\% of the latency. Scores normalized to 0--100.}
    \label{fig:case_think}
    \vspace{-8pt}
\end{wrapfigure}
We first examine the effect of CoT reasoning on planning quality on VLABench~\cite{Zhang_2025_ICCV}, a multi-step reasoning benchmark, where a high-level planner decomposes abstract instructions (\eg ``sort these books on the shelf so the library looks tidy'') into an executable skill sequence. High-level planners that employ deliberate thinking chains can reason through subtle physical, semantic, and spatial constraints, yet this reasoning comes at a heavy cost in token generation latency. Comparing Qwen3-VL 8B Instruct versus Qwen3-VL 8B Thinking (Fig.~\ref{fig:case_think}), we find while thinking improves average performance, on 44\% of tasks the Instruct model matches or outperforms the Thinking model with orders of magnitude faster inference (see Appendix \ref{app:results:expanded}, \ref{app:results:permodel}). This suggests relying on reasoning models can be wasteful as many scenes can be solved efficiently for real-time viability.

\vspace{-7pt}
\subsection{Scaling Model Size: Command on More Skills.}
\vspace{-5pt}
\begin{wrapfigure}{r}{0.6\textwidth}
    \centering
    \vspace{-5pt}
    \includegraphics[width=\linewidth]{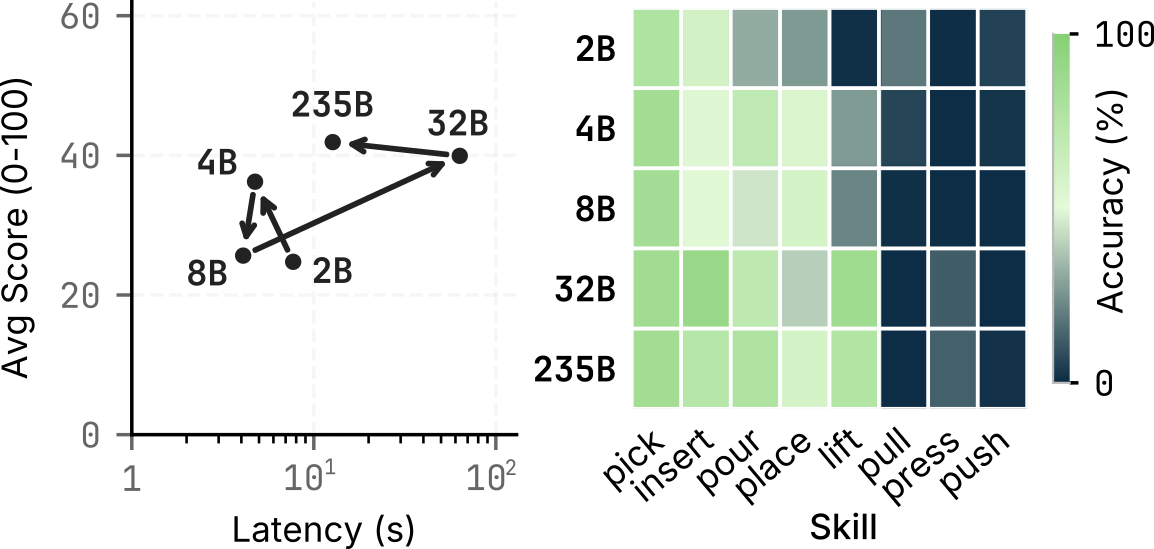}
    \vspace{-12pt}
    \caption{\footnotesize Scaling Model size (2B--235B) on VLABench~\cite{Zhang_2025_ICCV}. \textbf{Left:} score and latency scale non-monotonically with model size. \textbf{Right:} model size broadens the skill set a planner can command.}
    \label{fig:case_size}
    \vspace{-5pt}
\end{wrapfigure}
We investigate how planner performance scales with model size. Evaluating Qwen3-VL Instruct from 2B to 235B parameters~\cite{bai2025qwen3vltechnicalreport}, we find brute-force scaling traces no clean curve in either score or latency (Fig.~\ref{fig:case_size} left, Appendix \ref{app:results:expanded}, \ref{app:results:permodel}). Score is non-monotonic, and since verbosity is not tied to parameter count, smaller variants sometimes generate too many tokens and run slower than larger ones (\eg 32B is slower than 235B). Decomposing by skill reveals size primarily governs the \emph{breadth} of skills a planner can reliably command (Fig.~\ref{fig:case_size} right). The larger model's edge is correctly commanding a broader set of skills, paying off only on tasks that demand them. Uniform deployment of the large model wastes compute on other tasks, motivating per-task allocation.

\subsection{Varying Memory Architectures: Contextual Overhead.}
\vspace{-5pt}
\begin{wrapfigure}{r}{0.6\textwidth}
    \centering
    \includegraphics[width=\linewidth]{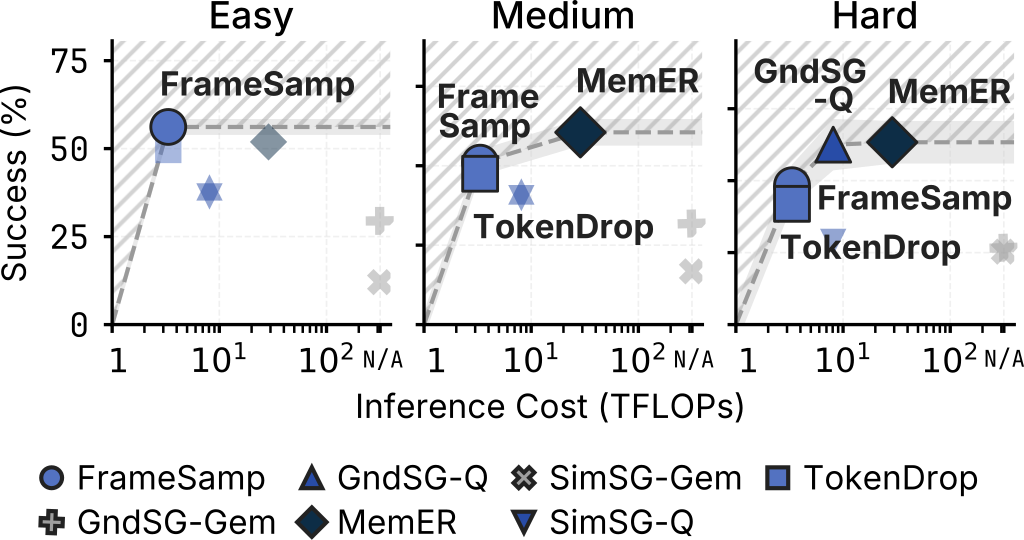}
    \caption{\footnotesize Memory architectures across difficulty tiers on RoboMME~\cite{dai2026robomme}. Lightweight schemes win cheaply on easy tasks; \texttt{MemER} and \texttt{GroundSG}
lead on hard tasks.}
\label{fig:case_mem}
\vspace{-10pt}
\end{wrapfigure}
Long-horizon tasks can require memory, but a planner VLM can compress history in very different ways in context. \texttt{FrameSamp} and \texttt{TokenDrop} reduce raw visual tokens, \texttt{SimpleSG} and  \texttt{GroundSG} summarize history as language subgoals, and \texttt{MemER}~\cite{sridhar2025memer} recalls prior keyframes. On RoboMME~\cite{dai2026robomme}, we observe each trades accuracy against overhead, with the most efficient choice shifting with difficulty (Fig.~\ref{fig:case_mem}, Appendix \ref{app:results:expanded}). On easy tasks needing short recall, the lightweight \texttt{FrameSamp} outperforms \texttt{MemER} at an order of magnitude fewer FLOPs; as tasks demand distant history, \texttt{MemER} and \texttt{GroundSG} have higher success. No single architecture dominates as each demonstrates benefits in particular levels of difficulty.

\noindent \textbf{The Case for Adaptive Allocation.} These investigations yield three trends: (1) large capability gaps separate cheap and expensive configurations; (2) the gaps are non-uniform, hinging on task-specific demands like skill command or recall horizon; and (3) lightweight models suffice on many tasks. Since any static model fixes a single latency–capability compromise, these observations motivate a framework that infers each task's demands from the scene context and allocates compute per task.

\vspace{-5pt}
\section{Proposed Embodied Routing Framework: \algname}
\label{sec:method}
\vspace{-5pt}

\begin{figure*}[t]
    \centering
    \includegraphics[width=\linewidth]{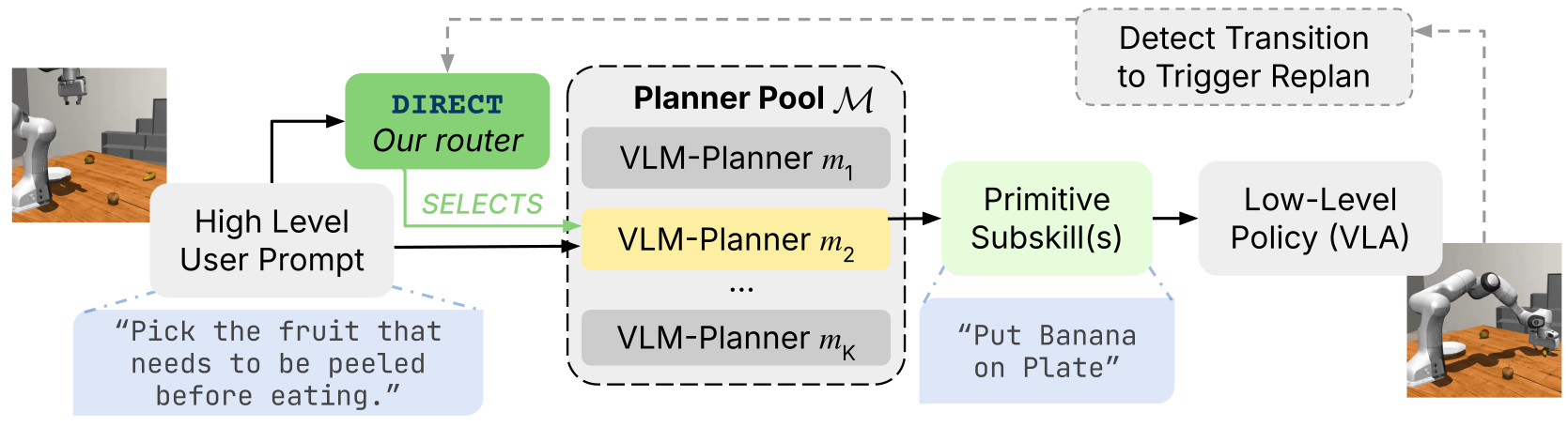}
    \caption{Given a user instruction and scene observation, our lightweight router \algname selects the planner with the best quality--cost tradeoff from a fixed pool of VLM planners $\mathcal{M} = \{m_1, \dots, m_K\}$. The chosen VLM decomposes the instruction into contextual commands (e.g., primitive subskills or free-form language instructions) executed by a single low-level VLA policy, while a transition detector triggers replanning for multi-stage tasks. By conditioning on multimodal context, \algname allocates test-time compute per task rather than using a single fixed-capability planner.}
    \label{fig:router_framework}
\vspace{-7pt}
\end{figure*}

We introduce our routing framework \algname, which aims to learn a mapping from a task's
scene and instruction to the planner best suited to solve it. \algname chooses among a
fixed pool of candidates that differ in capability and cost. Concretely, we formalize
embodied test-time compute allocation as a selection problem over a fixed pool
$\mathcal{M} = \{m_1, \ldots, m_K\}$ of VLM embodied planners, each of which maps
$m_k: x \mapsto (a_1, \ldots, a_T)$ from a task to a sequence of contextual commands
that a single downstream language-conditioned policy executes on the robot.
Because routing operates at the planner level, \algname is agnostic to the form of these
commands: they can be primitive sub-skills from a fixed vocabulary, as in VLABench
(App.~\ref{app:bench:vlabench}), or free-form language sub-goals grounded in the capabilities of a generalist
policy such as $\pi_{0.5}$-DROID; only the planners' output
format changes, not the routing mechanism (App.~\ref{app:prompts}).
A task input $x = (I, \ell)$ pairs an initial scene observation $I$ (an RGB image, or a
short multi-view stack where available) with a natural-language instruction $\ell$.
Given $x$, a lightweight router predicts which planner offers the best quality--cost
tradeoff and selects a single $m_k \in \mathcal{M}$ (Fig.~\ref{fig:router_framework}).

\paragraph{Data Collection.} For every training task $x_i \in \mathcal{D}_\text{train}$ and planner $m_k$ we measure two quantities. The first is a quality score $q_{i,k} \in [0,1]$: the task success rate (or progress score where available) when $m_k$'s sub-skill sequence $(a_1, \ldots, a_T)$ is executed end-to-end. The second is an inference cost $c_{i,k} \in \mathbb{R}_{> 0}$, the resource measure that constrains deployment. We use latency when planners differ along an autoregressive axis (e.g., reasoning depth), since wall-clock time is the operative constraint on sequential generation. We use FLOPs when they differ in parallel structure (\eg memory architectures), since latency there conflates compute with hardware-level parallelism. Collecting these across all $N$ tasks and $K$ planners yields $Q \in [0,1]^{N \times K}$ and $C \in \mathbb{R}_{> 0}^{N \times K}$, the quality and cost matrices over the task--planner grid. In simulation, we score rollouts directly from the benchmark.

\paragraph{Synthetic Data Generation for Hardware.} On hardware, collecting data to train the router by exhaustively rolling out every planner on every physical scene is impractical. So, we synthesize tasks offline by sampling scenes disjoint from evaluation, prompting a large VLM to propose candidate instructions with reference skill decompositions, running each planner $m_k$ on those synthetic tasks to record its sequence and latency, and scoring quality with an LLM judge against the reference. This yields $Q$ and $C$ training data for physical-style tasks without requiring robot hardware rollouts, and the resulting trained router is deployed zero-shot on hardware (details in Appendix \ref{app:bench:synthetic}).

\paragraph{Router Architectures.} The router $r(\cdot)$ maps encoded task features $\phi(x) \in \mathbb{R}^d$ to a planner index $\hat{k}\in\{1, ...K\}$, introduced at the start of the section. We construct $\phi$ by encoding $I$ with a frozen SigLIP-family~\cite{zhai2023sigmoid, tschannen2025siglip} vision encoder and $\ell$ with a frozen BGE-M3 \cite{multi2024m3} text encoder, then concatenating the two embeddings into a single vector. Additional details and ablations over various embedding fusion strategies, vision encoders, and embedding types are reported in Appendix \ref{app:router:embeddings}. We evaluate many parameter-efficient router architectures, in which router inference cost and a single embedding pass remains negligible (20-50 total millisec) against any VLM planner in $\mathcal{M}$ ($>$1 sec). Router candidates include linear models, $k$-nearest-neighbors (KNN), pairwise preference aggregated KNN (PRkNN)~\cite{zheng2023efficient}, $k$-means, one-versus-rest (OVR)~\cite{rifkin2004defense}, and two-layer multilayer perceptrons. Our extensive architecture sweep is detailed in the Appendix \ref{app:router:arch}.

\paragraph{Routing Objective.} As established in Section~\ref{sec:casestudies}, no single planner dominates across tasks, and quality and cost objectives conflict. To choose among planners, we require a scalar criterion that combines objectives into a single quantity, since the routing decision is ultimately a one-dimensional argmax over $\mathcal{M}$. We capture this through a utility function $U(q, c)$, defining the per-task target as
\vspace{1pt}
\begin{equation}
    k_i^\star \;=\; \arg\max_{k} \;\; U\!\left(q_{i,k},\, c_{i,k}\right).
    \label{eq:target}
\end{equation}
\vspace{1pt}
The framework is agnostic to the specific form of $U$, and we expose the tradeoff through a family of routing objectives, each corresponding to a different instantiation of Equation~\ref{eq:target}. 
One strategy is to predict $\hat{q}_{i,k}$ and $\hat{c}_{i,k}$ via regression heads against the entries of $Q$ and $C$, so for $\alpha \geq 0$ weighting cost against quality, the utility can be framed as $U_{\text{reg}}(\hat{q}_{i,k},\, \hat{c}_{i,k}) \;=\; \frac{1}{\hat{c}_{i,k}} \cdot \mathbf{1}[ \max_{k'}\hat{q}_{i,k'} - \hat{q}_{i,k} < \alpha\,]$. We detail and ablate on other utility functions and heads (see Appendix \ref{app:router:heads}). We generally found regression as the best head likely since it leverages the most expressive signal of quality and cost.

\paragraph{Evaluation Metric.} Each router configuration, defined by its choice of head, hyperparameters, and architecture, realizes a different point in the quality-cost plane. We compare routers via a harmonic-mean efficiency score ($\eta$) inspired by LLM/VLM routing benchmark metrics~\cite{lu2025routerarena, huang2025vlrouterbenchbenchmarkvisionlanguagemodel}:
\vspace{1pt}
\begin{equation}
    \eta(\beta) = \frac{(1 + \beta)\,q_s\,(1 - c_s)}{\beta\,q_s + (1 - c_s)},
    \label{eq:harmonic_score}
\end{equation}
\vspace{1pt}
where $q_s$ and $c_s$ are the router's mean quality and cost min-max-scaled on a hold-out set across all routers and baselines in the comparison, as well as an oracle quality ceiling, and $\beta$ weighs relative emphasis on quality. This evaluation criterion is decoupled from any specific head's objective.

\paragraph{Deployment.} At inference, we observe a fresh task $x = (I, \ell)$, compute $\phi(x)$, and the router predicts a planner index $\hat{k} = r(\phi(x))$. The router is invoked once per task, or repeatedly for multi-stage tasks, with each call deciding independently on the current $(I, \ell)$.

\vspace{-5pt}
\section{Experimental Results}
\label{sec:result}
\vspace{-5pt}

We evaluate the effectiveness of \algname across three distinct axes of test-time compute: CoT reasoning depth, model size, and memory history. Each axis corresponds to a substantively different way in which VLMs in the pool $\mathcal{M}$ can apply more compute at test time to increase capability, and each gives rise to a routing problem whose structure we characterize empirically. We train and evaluate routers in simulation and in Franka DROID hardware (Appendix \ref{app:results:hardware}, \ref{app:results:hardware:pareto}, \ref{app:bench:droid}), validating how \algname improves test time compute scaling across all axes. In total our results contain over 270,000 simulated routing decisions and 245 hardware trajectories.

\paragraph{Baselines.} \label{par:setup:baselines} We hypothesize the benefit of test-time compute is task-contingent, so \algname will recover the strongest/expensive model's quality at a fraction of its cost by invoking it only when needed. In each regime we compare the learned router against: \emph{Cheap} and \emph{Expensive}, which are the static choices with, respectively, the lowest and highest mean cost, marking the two deployable allocation extremes; \emph{Random}, which allocates uniformly at random; and \emph{OOD detection}, in which tasks whose embeddings are out-of-distribution (detected via conformal prediction~\cite{sinha2024realtime}) of a calibration set are routed to progressively more expensive variants. We report results for the router with the optimal configuration and hyperparameters in the sweeps for \emph{OOD detection} and \algname. Distinct from the other baselines, the \emph{oracle} selects the per-task optimal variant with full knowledge of $Q$ and $C$. The \emph{oracle} is not deployable, but bounds the achievable frontier given the pool; the gap to any policy quantifies the headroom available to selective allocation.

\vspace{-3pt}
\subsection{Routing on the Chain-of-Thought Test-Time Compute Axis}
\label{sec:result:reasoning}
\vspace{-2pt}

\begin{table*}[t]
\centering 
\rowcolors{1}{}{}
\caption{\footnotesize \textbf{VLABench routing comparison} ($N = 900$). Comparing learned routing against baselines on VLABench. Columns report mean success score (Q, \%) and average planning inference latency (Lat, s); routing methods additionally report routing efficiency ($\eta$, \%) where $\beta=0.1$. More extensive results can be found in Appendix \ref{app:ablations}. Pair labels: I/T = Instruct/Thinking, M/T = Minimal/Thinking. $^\dagger$ These models were evaluated on a 20\% subsample due to substantially higher inference time.}
\label{tab:learned_vs_oracle_full}
\setlength{\tabcolsep}{4pt}
\renewcommand{\arraystretch}{1.15}
\setlength{\dashlinedash}{0.5pt}
\setlength{\dashlinegap}{1.5pt}
\resizebox{\textwidth}{!}{
\begin{tabular}{|l||cc|cc||cc:c|cc:c||>{\columncolor{lime!10}}c>{\columncolor{lime!10}}c:>{\columncolor{lime!10}}c|}
\hline
& \multicolumn{2}{c|}{\textit{Cheap}} & \multicolumn{2}{c||}{\textit{Expensive}} & \multicolumn{3}{c|}{\textit{Random}} & \multicolumn{3}{c||}{\textit{OOD Detection}} & \multicolumn{3}{c|}{\cellcolor{lime!10}\algname} \\
\textbf{Pair (Cheap / Expensive)} & Q & Lat & Q & Lat & Q & Lat & $\eta$ & Q & Lat & $\eta$ & Q & Lat & $\eta$ \\ \hline
\rowcolor{black!15} \multicolumn{14}{|l|}{\textit{Homogeneous Open-Weight Pairs}} \\
GLM-4.6 I / 4.1 T       & 27.24 & 2.11  & 54.02 & 45.10 & 40.02 & 22.39 & 46.33 & 49.83 & 39.60 & 54.52 & 51.79 & 31.82 & \textbf{75.26}\\
GLM-4.6 I / T      &  27.24    &  2.11   &  46.09  & 24.08 &  35.97    &  13.24   &  43.68    &   42.73   &  20.85   &    55.34  &  44.69   &   16.97   &    \textbf{74.92} \\
Qwen3-VL 8B I / T $^\dagger$ &  31.64    &  1.85   &  48.28   &  120.76   &  40.62   &  64.40    &  48.48  &  46.89   & 106.22    &  54.20   & 47.48    &  81.72   & \textbf{ 74.78}   \\
\rowcolor{black!15} \multicolumn{14}{|l|}{\textit{Homogeneous Closed-Weight Pairs}} \\
Gemini 3 Flash M / T    &  61.22   &  1.88    &  64.22    &  15.86   &  62.11    &  8.82   &  15.53   &  61.83    & 6.70    & 10.81    &  64.83   &  8.47    &  \textbf{57.99}    \\
Gemini-ER 1.6 M / T     &  62.88    &  2.31  &  62.62   &  7.02    &  62.46   &  4.67    &  0.0   &  63.19   &  3.13    &  19.68    & 63.92    & 4.72    & \textbf{37.09}   \\
\rowcolor{black!15} \multicolumn{14}{|l|}{\textit{Heterogeneous Model Pairs}} \\
Cosmos-R2 / Qwen3-VL 8B T $^\dagger$     & 29.56    & 2.02    & 48.28    & 120.76    & 38.93    &  64.49   & 45.03    &  37.32   &  49.37   &  38.47    & 46.98    &  86.36   &  \textbf{71.16}    \\
RynnBrain-Plan / GLM 4.1 T   &  45.25    &  1.67    &  54.02   &  45.10    &  49.44   & 22.17    & 32.51    &  52.56   &  32.72    & 50.42    &  54.33    &  27.50    &  \textbf{63.94}    \\
\hline
\end{tabular}
}
\vspace{-10pt}

\end{table*}

\paragraph{Simulation Results.}
\label{sec:result:reasoning:sim-thinking}
We evaluate routing over No-Thinking (\eg instruct), Thinking model pairs drawn from both open-weight and closed-weight families, along with heterogeneous open-weight model pairs. As seen in Table~\ref{tab:learned_vs_oracle_full}, \algname achieves the highest routing efficiency $\eta$ in every configuration. Open-weight pairs reveal substantial CoT headroom that \algname exploits most aggressively. On the three homogeneous open-weight pairs, the cheap-to-expensive quality gap is large, and \algname recovers near-expensive quality while cutting latency by around $30\%$, with $\eta \approx 75\%$ across all three pairs. On closed-weight pairs, \algname still achieves the best routing efficiency $\eta$, and furthermore \algname's success score \textit{exceeds both} the cheap and expensive static choices, evidence that our routing framework extracts complementary strengths even when neither variant dominates on average. The advantage persists under heterogeneous pools, where cheap and expensive are drawn from different model families. \algname recovers most of the expensive variant's quality at $\sim$30-40\% lower latency, indicating the learned signal is not tied to within-family stylistic cues. Together, these results indicate that only a subset of VLABench's language-conditioned manipulation tasks benefit from CoT, and \algname can identify these tasks from the initial scene and instruction across model families and capability gaps.

\begin{wrapfigure}{l}{0.35\textwidth}
    \centering
    \includegraphics[width=\linewidth]{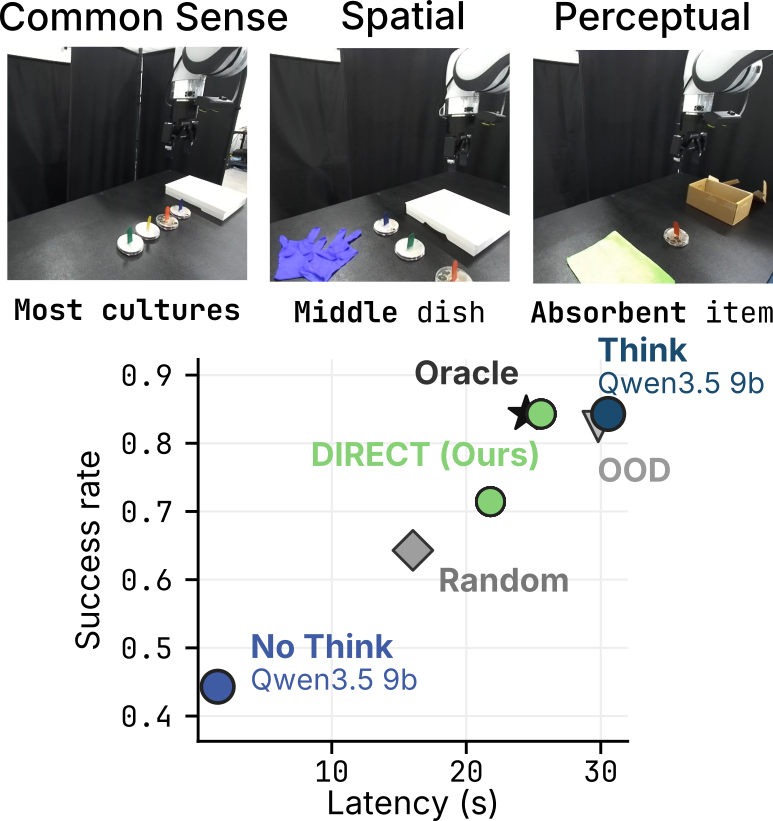}
    \caption{\footnotesize Qwen3.5-9B Instruct vs. Think on DROID. \algname recovers Think's success at fraction of latency. }
\label{fig:exp_reasoning_physical}
\vspace{-10pt}
\end{wrapfigure}

\paragraph{Physical Validation.}
\label{sec:result:reasoning:physical}
We validate these findings on physical robot tasks using Qwen3.5-9B with and without CoT reasoning, evaluating each variant across five tasks spanning three cognitive categories: 
Common Sense / World Knowledge (selecting objects that require background knowledge to identify), Spatial reasoning (placing objects under relational constraints), and Perceptual understanding (identifying objects by their properties). As shown in Fig.~\ref{fig:exp_reasoning_physical}, the non-thinking planner is fast but low in success, while CoT reasoning substantially raises success at a 20$\times$ latency cost. 
Across our five tasks, the non-thinking planner matches the Thinking variant on two tasks, while thinking yields a significant gain on the remaining three. Because this benefit is task-contingent, \algname escalates to the thinking model only where it helps, recovering the Thinking variant's success at a significantly lower latency (details in Appendix Table \ref{tab:app:physical_routers_intelligence}). \algname achieves close to the Oracle on both success and latency, and outperforms both OOD- and random-based routing.

\begin{wraptable}{r}{0.4\textwidth}
\centering
\vspace{-12pt}
\caption{\footnotesize \textbf{Multi-step grocery bagging (heaviest to lightest).} After each completed step, \algname routes to the appropriate model to plan the next, matching Thinking planner's success at lower latency.}
\vspace{-3pt}
\label{tab:exp_size_physical}
\setlength{\tabcolsep}{4pt}
\renewcommand{\arraystretch}{1.15}
\resizebox{0.4\textwidth}{!}{%
\begin{tabular}{|l||cc|}
\hline
\textbf{Planner} & Success (\%) & Lat (s) \\ \hline
Qwen3.5-VL 9B I  & 47.62 & 2.19 \\
Qwen3.5-VL 9B T & 90.48 & 19.58 \\
\rowcolor{lime!10} \algname & \textbf{95.24} & 6.85 \\
\hline
\end{tabular}%
}
\vspace{-10pt}
\end{wraptable}
We further evaluate \algname on a multi-step grocery-bagging task that bags items from heaviest to lightest, where each completed step exposes a new scene, and the planner is re-invoked to plan the next (see Fig.~\ref{fig:hero}, top).
Since steps vary in reasoning demand, this experiment tests per-step allocation: \algname decides at each step whether routing to the thinking model is worth its cost. As shown in Table~\ref{tab:exp_size_physical}, \algname, which routes step-wise, attains the highest progress success, matches the Thinking model's performance with much lower latency, while the non-thinking model alone generally fails to complete the task successfully.
\vspace{-10pt}
\subsection{Routing on the Model Size Test-Time Compute Axis}
\label{sec:result:capacity}

\begin{wrapfigure}{l}{0.45\textwidth}
    \centering
    \includegraphics[width=\linewidth]{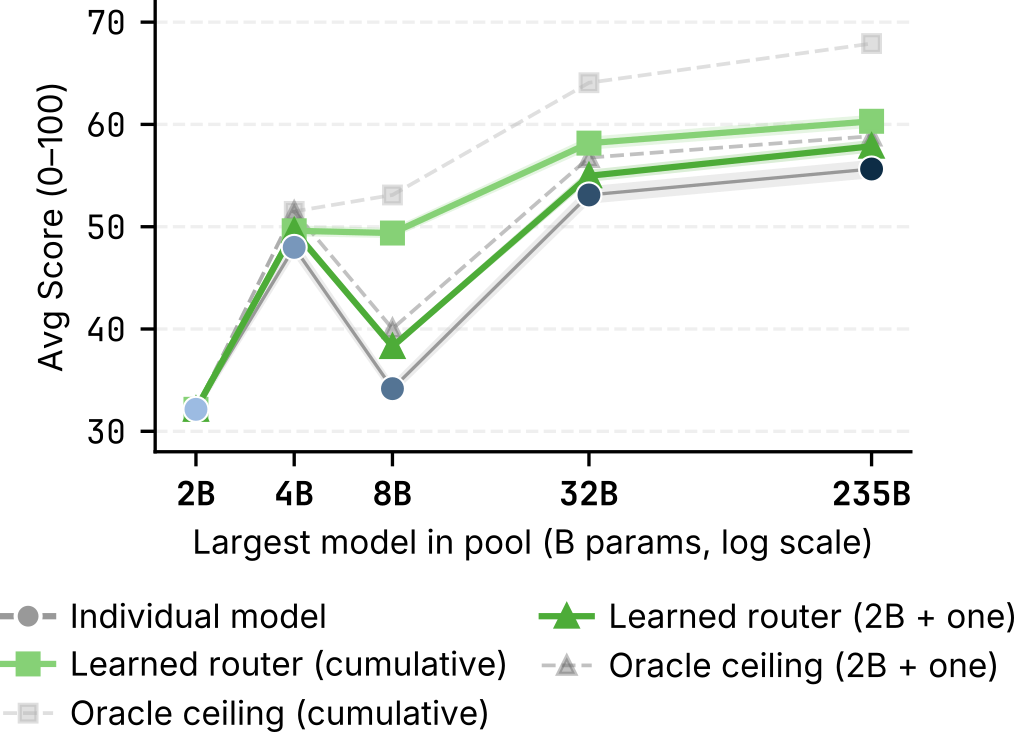}
    \caption{\footnotesize Per-task routing across Qwen3-VL Instruct 2B--235B on VLABench~\cite{Zhang_2025_ICCV}.}
\label{fig:exp_size}
\vspace{-10pt}
\end{wrapfigure}

\paragraph{Simulation Results.} \label{sec:result:reasoning:sim-size}
Because scale helps unevenly across skills, simply deploying the largest planner overspends on the many tasks where it adds little benefit. We use \algname to route over Qwen3-VL Instruct variants from 2B to 235B on VLABench, comparing per-task routing against the individual models and an oracle ceiling (Fig.~\ref{fig:exp_size}). Consistent with Section~\ref{sec:casestudies}, individual-model performance is non-monotonic in scale. \emph{Pairwise} routing between the 2B variant and a single
larger model (Fig.~\ref{fig:exp_size}) improves over either constituent at each size but inherits the pool's non-uniformity: with only a weak large model to escalate to, the 8B dip persists, though the curve still tracks its oracle ceiling. \emph{Cumulative} routing over all sizes up to each point instead becomes monotonic: since each added variant only enlarges the collective skill set, the router never loses an earlier capability, maintaining $49.3\%$ at 8B by falling back on the 4B model while reserving larger ones for tasks that demand them. For the most expensive model (32B), the cumulative router (2B+4B+8B+32B) improves performance by 5.1 points while reducing average latency by 32.4 seconds. \algname thus converts an erratic scaling curve into monotonic improvement, recovering each model's distinct skill coverage while discarding its weaknesses.

\vspace{5pt}

\begin{wrapfigure}{r}{0.45\textwidth}
    \centering
    \includegraphics[width=\linewidth]{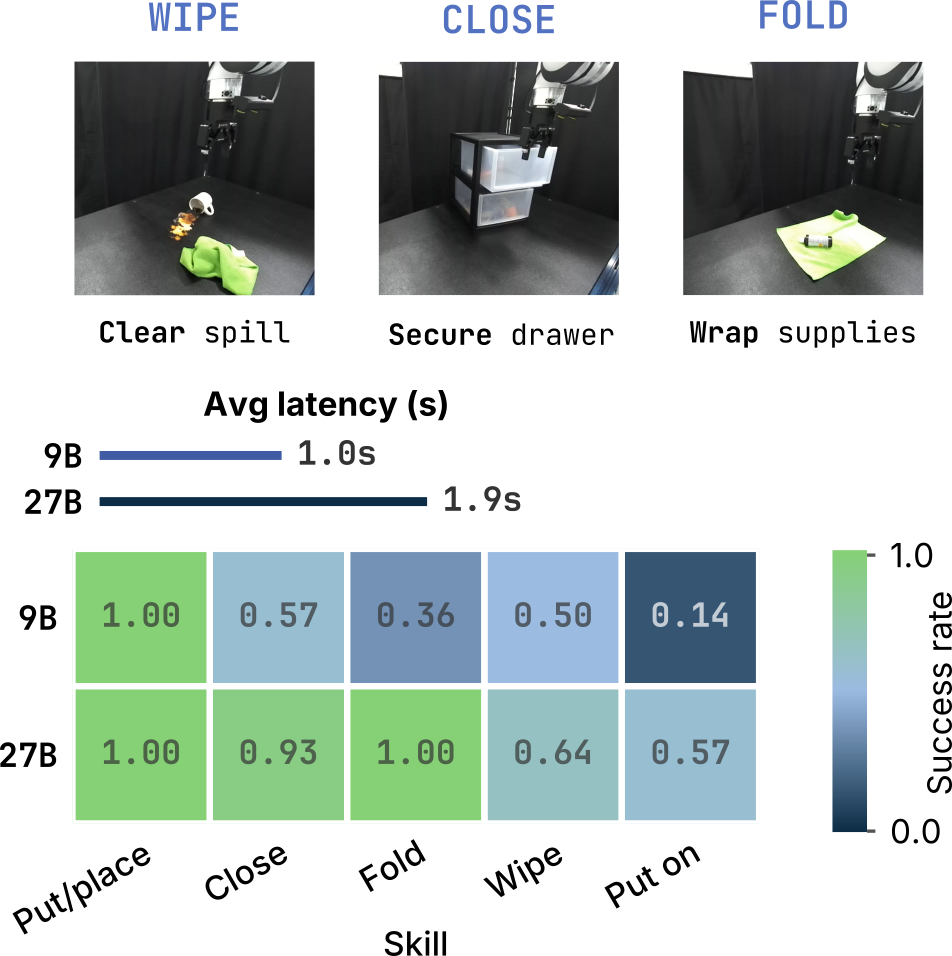}
    \caption{\footnotesize Qwen3.5-VL 9B vs. 27B on DROID. Apart from \texttt{Put/Place}, the 27B planner raises success on every skill at $\sim2\times$ latency.}
    \label{fig:exp_size_physical}
\vspace{-5pt}
\end{wrapfigure}
\paragraph{Physical Validation.} We validate our observation from Section~\ref{sec:casestudies} on scaling model size using Qwen3.5-VL at 9B and 27B Instruct as embodied planners on the Franka DROID hardware across five household skills. As seen in Fig.~\ref{fig:exp_size_physical}, the 27B planner improves command success on every skill except for \texttt{Put/Place} at double the latency of the smaller model. This mirrors the simulation finding that model size primarily broadens the set of skills a planner can reliably command: while a 9B model could command basic skills like \texttt{Put/Place} ($1.00$) and is mediocre at \texttt{Wipe} ($0.50$), improved command on a broader set of skills including \texttt{Fold} ($0.36 \rightarrow 1.00$) and \texttt{Close} ($0.57 \rightarrow 0.93$) depend almost entirely on the larger 27B model. Even where the 27B helps, the gains are not uniform: \texttt{Wipe} only climbs to 0.64 and \texttt{Put on} to 0.57, while \texttt{Fold} and \texttt{Close} jump to near-saturation. The benefit of a larger planner is therefore concentrated on the specific skills a task demands rather than spread uniformly across the skill set (see Appendix Table \ref{tab:app:physical_routers_skill}).

\subsection{Routing on the Memory Context Test-Time Compute Axis}
\label{sec:result:memory}

\begin{figure}[ht]
    \centering
    \includegraphics[width=0.9\linewidth]{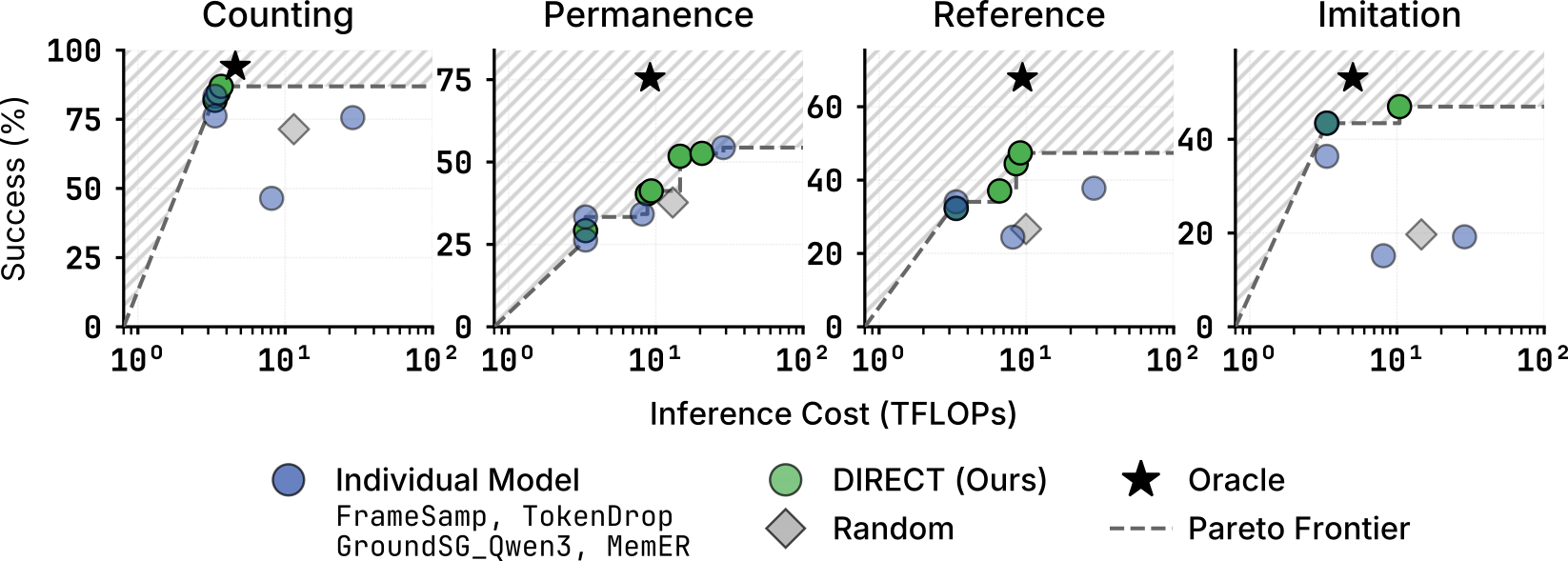}
    \caption{\footnotesize RoboMME memory routing across difficulty suites. Success rate versus inference cost (TFLOPs) for individual memory architectures and \algname's routed Pareto frontier (dashed). In Reference the routed frontier outperforms the strongest single architecture at far lower cost than the most expensive specialist (\texttt{MemER}), evidence that no single model is optimal suite-wide.}
    \label{fig:exp_mem_sim}
\vspace{-5pt}
\end{figure}

\paragraph{Simulation Results.}
\label{sec:result:memory:sim}
The per-suite Pareto comparison (Fig.~\ref{fig:exp_mem_sim}) shows that routing's value is itself non-uniform across memory families. In each suite, \algname generally achieves a success–cost point better than the most performant architecture, at a fraction of the most expensive specialist's (\texttt{MemER}) cost. This is only possible when a family decomposes into sub-types with different optimal recall structures: if one memory model were optimal family-wide, the router could at best match it by always selecting it. \algname's margin over the best specialist is therefore evidence that it has discovered \emph{multiple successful decompositions} within a family. A residual gap to the oracle persists across families, whose source we leave open: it may be \emph{unlearnable noise} that no router could predict, or \emph{learnable structure} that current routers in \algname miss, implying a stronger framework could recover it. Resolving this is left for future work on the learning theory of embodied routing.

\paragraph{Physical Validation.}
\label{sec:result:memory:physical}
We test memory routing on the physical Franka DROID setup across two task types that stress recall differently: \emph{Unobservable Goal} tasks, whose target is specified only by task history no longer visible in the current frame, and \emph{Observable Goal} tasks, whose constraints can be resolved from the present scene. The non-memory model is Gemini-ER 1.6 with a single frame and task as context. We construct a memory planner by providing Gemini-ER 1.6 with both a 40-frame
\begin{wrapfigure}{l}{0.4\textwidth}
    \centering
    \includegraphics[width=\linewidth]{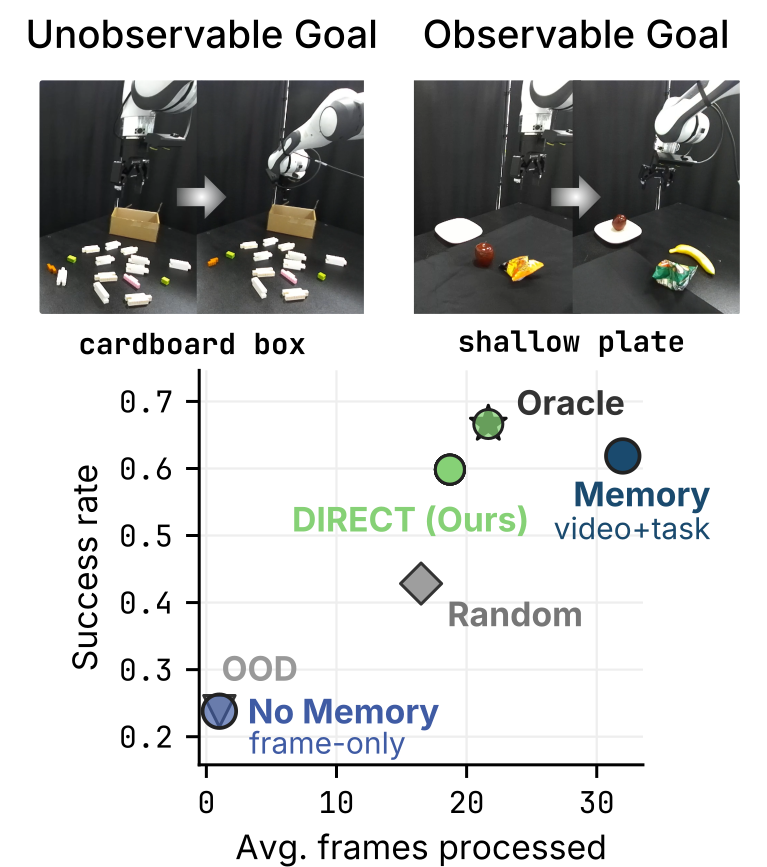}
    \caption{Routing between No-memory and Memory planners on DROID. \algname achieve close to oracle performance.}
\label{fig:exp_mem_physical}
\vspace{-10pt}
\end{wrapfigure}
video history along with a summary of task list histories. As shown in Fig.~\ref{fig:exp_mem_physical}, \algname recovers near-oracle success rate while operating at substantially lower average frames processed than the full memory-augmented planner and outperforms the memory-free variant as well as the OOD and random routing baselines (details in Appendix Table \ref{tab:app:physical_routers_memory}). This confirms the simulation finding in physical execution: memory-oriented compute is only worth its cost when a task genuinely depends on distant history. Furthermore, \algname can allocate test-time compute in the form of historical context selectively rather than paying the recall overhead on every task.

\paragraph{Routing across axes.}
\label{sec:cross_axis}
Although we study each axis separately to isolate its contribution, \algname is
agnostic to what differentiates planners in $\mathcal{M}$, so it applies directly to
pools that vary along multiple axes at once. We route over a joint pool varying both
model size and CoT (Qwen3-VL-8B Instruct, 8B Thinking, and 32B Instruct;
App.~\ref{app:cross_axis}). The best router recovers most of the Thinking planner's
performance ($38.29$ vs.\ $39.81$) at roughly $40\%$ of its cost.
\vspace{-10pt}
\section{Conclusion}
\vspace{-10pt}
\label{sec:conclusion}

We studied when and where test-time compute pays off in VLM embodied planning, showing across reasoning depth, model size, and memory history that its benefit is task-contingent rather than uniform: each axis yields a qualitatively distinct capability gain, and cheap configurations already suffice on a large fraction of tasks. Building on this, \algname routes each task to a planner whose capability profile matches its inferred demands, recovering near-frontier success at substantially lower cost on VLABench, RoboMME, and on Franka DROID hardware experiments. Our synthesis suggests we can infer task difficulty from scene and instruction context, suggesting practitioners can build effective routers from lightweight, embedding-based features. On the VLM embodied planners and benchmarks evaluated, these results suggest that treating the high-level planner as a dynamic variable is a practical path to recovering frontier-level success at meaningfully lower cost. 

\paragraph{Limitations.} The router is trained offline over a fixed pool, matching the standard deployment setting where candidate planners are known in advance. Adapting to a changing pool requires recollecting $Q,C$ and retraining, and by construction the router allocates among pool members rather than synthesizing new capabilities. We treat cost as a deployment-bound proxy (\eg latency or FLOPs), so reported crossover points are representative of these settings rather than universal constants. Our per-call formulation conditions on features $\phi(x)$ from the initial observation, to keep router overhead negligible relative to any VLM planner, so we do not model cross-stage dependencies.


\acknowledgments{
The authors would like to thank Zen Yasukawa and Steve Cousins for help with hardware setup.
JD's work is supported by NSF Graduate Research Fellowship Program.
MG's work is supported by NASA University Leadership Initiative, Stanford Center for Automated Reasoning, Stanford Google Cloud Platform, and Thinking Machines Lab.
YA is currently employed by Wing; however, this research was conducted independently and not in connection with her employment.
AS's work is supported by NSF Graduate Research Fellowship Program.
MNA's work is supported by the Natural Sciences and Engineering Research Council of Canada (NSERC)'s Postdoctoral Fellowship and Mitacs Globalink Research Award.
KL's work is supported by the Defense Advanced Research Projects Agency (DARPA) under Agreement Number 00011869.
This project is also supported by Stanford Robotics Center, Cornell University Gemini Academic Program, ONR grant N00014-22-1-2621, DoD High Performance Computing Modernization Program, Stanford Marlowe GPU Cluster~\cite{kapfer2025marlowe}, the National Artificial Intelligence Research Resource Pilot and the Anvil supercomputer~\cite{anvil} (award NSF-OAC 2005632). Any opinions, findings, and conclusions or recommendations expressed in this material are those of the authors and do not necessarily reflect the views of any aforementioned supporting entity.
}



\bibliography{ref}  

\newpage


\begin{center}
    {\Large\bfseries \algname: When and Where Should You Allocate \\[2pt]
    Test-Time Compute in Embodied Planners?}\\[6pt]
    {\large Appendix}
\end{center}
\vspace{1em}

\setcounter{tocdepth}{-1} 

{\hypersetup{linkcolor=black}\tableofcontents}
\vspace{1em}

\appendix
\addtocontents{toc}{\string\setcounter{tocdepth}{2}}



\section{Notation and Model References}
\label{app:notation}

\subsection{Notation}
Table~\ref{tab:app:notation} (below) collects the central symbols used in the main paper and
this supplement; minor, section-local notation is defined where it appears.

\begingroup
\footnotesize
\setlength{\LTpre}{3pt}\setlength{\LTpost}{3pt}
\renewcommand{\arraystretch}{1.1}
\begin{longtable}{@{}p{0.17\linewidth}p{0.77\linewidth}@{}}
\label{tab:app:notation}\\
\toprule
\textbf{Symbol} & \textbf{Meaning} \\
\midrule
\endfirsthead
\toprule
\textbf{Symbol} & \textbf{Meaning} \\
\midrule
\endhead
\bottomrule
\endlastfoot

\multicolumn{2}{@{}l}{\emph{Routing formulation (Sec.~4 of the main paper)}} \\
\addlinespace[2pt]
$\mathcal{M}$ & Planner pool: the $K$ candidate model variants the router selects among. \\
$N$ & Number of tasks / evaluation samples. \\
$x=(I,\ell)$ & Task input: scene observation $I$ $+$ language instruction $\ell$. \\
$Q,\,C$ & Per-(task, variant) quality and cost matrices ($N\times K$); entries $q_{i,k}\in[0,1]$, $c_{i,k}\geq 0$. \\
$\bar{c}_k$ & Training-time average cost of variant $k$. \\
$U$ & Preference (utility) function. \\
$k_i^\star$ & Routing target for task $i$: $\arg\max_k U(q_{i,k},c_{i,k})$. \\
$\lambda$ & Quality tolerance defining the classification target. \\
$\phi(x)$ & Fused multimodal embedding of $x$; the router's input features. \\
$r_\theta(x)$ & Variant the trained router selects for input $x$. \\
$\hat{q}_{i,k}$ & Regression head's predicted quality for sample $i$ and planner $k$. \\
$\hat{c}_{i,k}$ & Regression head's predicted cost for sample $i$ and planner $k$. \\

\addlinespace[3pt]
\multicolumn{2}{@{}l}{\emph{Evaluation metric}} \\
\addlinespace[2pt]
$\eta(\beta)$ & Routing efficiency: preference-weighted harmonic mean of scaled quality $q_s$ and cost-savings $(1-c_s)$. \\
$\beta$ & Cost-emphasis weight in $\eta$ \emph{and} also re-used for the rank-score cost-emphasis (in App.~\ref{app:router:preference} and App.~\ref{app:baselines:list}) . \\
$\alpha$ & Regression tie window (Base $\epsilon$, Tie $0.1 + \epsilon$). \\

\end{longtable}
\endgroup

\noindent\emph{Conventions.} Scores and $\eta$ values in the ablation tables are
$\times100$; cost is the VLM Planner's latency in seconds unless otherwise noted (ex. given in TFLOPs).

\subsection{Model References}
\label{app:models}

Throughout the paper we refer to models by short colloquial names; Table~\ref{tab:app:checkpoints}
maps each to its reference and its model id (Hugging Face repo, GitHub project, or API name)
for reproducibility.
\begin{table}[H]
\centering
\begin{minipage}{\linewidth}
    \centering
    \caption{Models used in the paper: reference and model ID. Open-weight rows give the
    Hugging Face repo (or GitHub project); API rows are closed-weight and accessed via the listed
    endpoint.}
    \label{tab:app:checkpoints}
    \footnotesize
    \begin{tabular}{@{}>{\raggedright\arraybackslash}p{0.42\linewidth} l >{\raggedright\arraybackslash}p{0.48\linewidth}@{}}
    \toprule
    \textbf{Name in text} & \textbf{Ref.} & \textbf{Model id} \\
    \midrule
    \multicolumn{3}{@{}l}{\emph{Planner VLMs}} \\
    \addlinespace[1pt]
    Qwen3-VL-\{2B,8B\}-Instruct & \citep{bai2025qwen3vltechnicalreport} & \texttt{Qwen/Qwen3-VL-\{2B,8B\}-Instruct}  \\
    Qwen3-VL-\{4B,8B,33B,235B\}-FP8-Instruct & \citep{bai2025qwen3vltechnicalreport} & \texttt{Qwen/Qwen3-VL-\{4B,\dots,235B\}-Instruct-FP8}  \\
    Qwen3-VL-8B-Thinking & \citep{bai2025qwen3vltechnicalreport} & \texttt{Qwen/Qwen3-VL-8B-Thinking}  \\
    Qwen3.5-\{9B,27B\} & \citep{qwen35blog} & \texttt{Qwen/Qwen3.5-\{9B, 27B\}} \\
    GLM-4.1V-Thinking & \citep{hong2025glm} & \texttt{zai-org/GLM-4.1V-9B-Thinking} \\
    GLM-4.6V-Flash & \citep{hong2025glm} & \texttt{zai-org/GLM-4.6V-Flash} \\
    Cosmos-Reason2 & \citep{nvidia2025cosmosreason2} & \texttt{nvidia/Cosmos-Reason2-8B} \\
    RynnBrain-8B & \citep{dang2026rynnbrain} & \texttt{Alibaba-DAMO-Academy/RynnBrain-8B} \\
    RynnBrain-8B-Plan & \citep{dang2026rynnbrain} & \texttt{Alibaba-DAMO-Academy/RynnBrain-Plan-8B} \\
    Gemini-3-Flash\textsuperscript{*} & \citep{deepmind2025gemini3flash} & \texttt{gemini-3-flash-preview}  \\
    Gemini-Robotics-ER 1.6\textsuperscript{**} & \citep{deepmind2025geminiroboticser16} & \texttt{gemini-robotics-er-1.6-preview} \\
    \midrule
    \multicolumn{3}{@{}l}{\emph{Encoders (router embeddings)}} \\
    \addlinespace[1pt]
    SigLIP-2 (large) & \citep{tschannen2025siglip} & \texttt{google/siglip2-large-patch16-512} \\
    SigLIP-2 (so400m) & \citep{tschannen2025siglip} & \texttt{google/siglip2-so400m-patch14-384} \\
    SigLIP (large) & \citep{zhai2023siglip} & \texttt{google/siglip-large-patch16-384} \\
    DINOv2 (large) & \citep{oquab2024dinov2} & \texttt{facebook/dinov2-large} \\
    BGE-M3 (text) & \citep{chen-etal-2024-m3} & \texttt{BAAI/bge-m3} \\
    $\pi_0$ (VLA policy embedding) & \citep{black2025pi0} & \texttt{lerobot/pi0\_base} \\
    \midrule
    \multicolumn{3}{@{}l}{\emph{Low-level VLA policy}} \\
    \addlinespace[1pt]
    $\pi_{0.5}$ (DROID-finetuned) & \citep{black2025pi05} & \texttt{Physical-Intelligence/openpi} (\texttt{openpi05}) \\
    \bottomrule
    \end{tabular}
    
    \vspace{1.5ex}
    \raggedright
    \scriptsize
    \textsuperscript{*} Set thinking level to \texttt{MINIMAL} or use default. \\
    \textsuperscript{**} Set thinking budget to 0 (thinking off) vs default budget.
\end{minipage}
\end{table}
\section{Router Design Decisions}
\label{app:router:design}

We build on the VL-RouterBench~\citep{huang2025vlrouterbenchbenchmarkvisionlanguagemodel}
structure for training and evaluating a collection of diverse routers. This includes making use of their scaffolding for the router architectures and the classification-headed routers, as well as their cost/quality matrix and embedding-generation pipelines, and training/evaluation steps. We
introduce our own pipeline adapted to our input
dataset and scoring structure and modify the existing methods to fit our project's unique requirements. Beyond classification routing, we repurpose the existing
router architectures to perform \emph{regression} directly on cost and score, and
define a set of test-time decision rules (utility functions) that select the VLM planner
from the predicted score and cost. We also extended the vision embedding-generation pipeline
with newer vision encoders (i.e., two from the SigLIP-2 family).

\subsection{Embeddings and Fusion Strategies}
\label{app:router:embeddings}

The router input is an embedding of the task. We use a \textbf{text} embedding of the
language instruction (text encoder) and a \textbf{vision} embedding of the scene observation
(vision encoder); a \emph{fusion method} specifies how these two are combined into
the single vector the router consumes. We additionally consider a \textbf{VLA} (policy)
embedding from the $\pi_0$ VLA, which enters only through the policy-embedding strategy
below. We ablate:

\begin{itemize}
    \setlength{\itemsep}{0.15em}
    \item \texttt{concat}: sequence concatenation of the text and image embeddings.
    \item \texttt{normalize\_concat}: the same after $\ell_2$-normalizing the text and
    image embeddings.
    \item \texttt{average}: mean of the text and image embeddings.
    \item \texttt{weighted\_average}: text-weighted mean of the text and image embeddings
    (text weight $\in\{0.25,0.75\}$, denoted \texttt{w0.25}/\texttt{w0.75} or \texttt{weighted\_average\_ text025}/\texttt{weighted\_average\_text075}).
    \item \emph{text embedding}: the instruction (text) embedding alone.
    \item \emph{vision embedding}: the scene (image) embedding alone.
    \item \emph{policy embedding}: Following the feature extraction strategy of SAFE \cite{gu2026safe}, we use the $\pi_0$ VLA policy's prefix hidden states (image+text), masked-mean-pooled into a single vector.
\end{itemize}
Quantitative comparison appears in the embedding ablations in Appendix~\ref{app:ablations}. We additionally ablate both the fusion strategy and the vision encoder for every routing model pair; the top-performing results are also reported in Appendix~\ref{app:ablations}.

\subsection{Router Architectures}
\label{app:router:arch}

All base router architectures were already implemented by VL-RouterBench \cite{huang2025vlrouterbenchbenchmarkvisionlanguagemodel}; we adapt
their training signal and inference rule (Appendix~\ref{app:router:heads}). We summarize
each architecture below; the chosen hyperparameters are listed in
Table~\ref{tab:app:router_hparams}.
\begin{itemize}
    \setlength{\itemsep}{0.15em}
    \item \textbf{Linear}: a single linear layer (\texttt{nn.Linear}). Despite the
    ``soft-label'' naming, for classification it reduces to a hard-label objective, since our binarized
    scores yield exactly one correct planner.
    \item \textbf{MLP}: a small two-hidden-layer network.
    \item \textbf{KNN}: distance-weighted vote over the $k$ nearest training embeddings.
    \item \textbf{PR-KNN}  \cite{zheng2023efficient}: preference-ranking KNN, aggregating neighbors by
    Copeland-score voting; often identical to KNN in our settings.
    \item \textbf{K-means}: a nearest-centroid classifier with one prototype per planner, the mean embedding of the training samples that planner is the labeled winner, routing each query to the most cosine-similar prototype.
    \item \textbf{OVR} \cite{rifkin2004defense}: one-vs-rest, one logistic regressor per planner.

\end{itemize}

    The above describes base behavior, modifications to behavior for regression are described in \ref{app:router:preference}.

\subsection{Router Heads}
\label{app:router:heads}

We consider two router-head families. Let $\phi(x)$ denote the fused embedding of
input $x$, $\mathcal{M}$ the planner pool, $Q$ the per-variant quality matrix, and
$C$ the per-variant cost matrix.

\begin{description}
    \setlength{\itemsep}{0.25em}
    \item[\emph{Classification.}] Given $\phi(x)$, the head directly predicts a planner choice from $\mathcal{M}$ based on training labels of optimal planner choice per sample, as described below (\ref{app:router:preference}).

    \item[\emph{Regression.}] Given $\phi(x)$, regression heads predict cost and quality, trained on the cost and quality of individual samples from $Q$ and $C$. These predicted cost and quality values are then used to choose a planner from $\mathcal{M}$ based on the decision functions defined below (\ref{app:router:preference}). 
\end{description}

\subsection{Preference and Test-Time Decision Functions}
\label{app:router:preference}

Given predicted score and cost, the
chosen planner is determined by utility rule:

\begin{equation}
    k_i^\star \;=\; \arg\max_{k} \;\; U\!\left(q_{i,k},\, c_{i,k}\right)
    \label{eq:app-target}
\end{equation}

\paragraph{Label definition (classification).} Training labels are determined by the
highest-scoring planner, with ties broken by lowest latency (cost), this is the same as generating labels with the following utility  rule with $\alpha = \epsilon$:
\begin{equation}
U_{\text{class}}({q}_{i,k},\, {c}_{i,k}) \;=\; \frac{1}{{c}_{i,k}} \cdot \mathbf{1}[ \max_{k'}{q}_{i,k'} - {q}_{i,k} < \alpha\,]
\end{equation}

We then train the routers to directly predict a model based on these training labels.

\paragraph{Test-time selection (regression).}

The example utility function given in the main paper is:
\begin{equation}
U_{\text{reg}}(\hat{q}_{i,k},\, \hat{c}_{i,k}) \;=\; \frac{1}{\hat{c}_{i,k}} \cdot \mathbf{1}[ \max_{k'}\hat{q}_{i,k'} - \hat{q}_{i,k} < \alpha\,]
\end{equation},

which is used for the following \textbf{Base} and \textbf{Tie} implementations.

Here are example decision methods used:
\begin{itemize}
    \setlength{\itemsep}{0.15em}
    \item \textbf{Base} ($U_{\text{reg}}(\hat{q}_{i,k},\, \hat{c}_{i,k})$ with  $\alpha = \epsilon$): highest predicted score, ties broken by
    lowest latency.
    \item \textbf{Tie} ($U_{\text{reg}}(\hat{q}_{i,k},\, \hat{c}_{i,k})$ with $\alpha = 0.1 + \epsilon$): all predicted scores within $0.1$ of the
    maximum are treated as equal, ties broken by lowest latency.
    \item \textbf{Rank}: uses the planner that maximizes the VL-RouterBench-derived rank-score $S(\beta)$ \cite{huang2025vlrouterbenchbenchmarkvisionlanguagemodel}. This utility score is computed for each sample $i$ and planner $k$ using the predicted quality $\hat{q}_{ik}$ and cost $\hat{c}_{ik}$, with costs log-normalized against the minimum and maximum values observed in the training set. We use tradeoff parameter $\beta = 0.1$.


\begin{equation}
C^{\mathrm{norm}}_{ik} =
\frac{\log_2(c_{\max}+\varepsilon) - \log_2\!\big(\operatorname{clip}(\hat{c}_{i,k},\,c_{\min},\,c_{\max})+\varepsilon\big)}
     {\log_2(c_{\max}+\varepsilon) - \log_2(c_{\min}+\varepsilon) + \varepsilon}
\end{equation}

\begin{equation}
S_{i,k}(\beta) =
\frac{(1+\beta)\,\hat{q}_{i,k}\,C^{\mathrm{norm}}_{i,k}}
     {\beta\,\hat{q}_{i,k} + C^{\mathrm{norm}}_{i,k} + \varepsilon}
\end{equation}

\begin{equation}
k^{\star}_i = \operatorname*{arg\,max}_{k} \; S_{i,k}(\beta)
\end{equation}

\end{itemize}

\paragraph{Regression-based score and cost prediction.} In every regression variant, the router
predicts for each planner, a continuous quality score (regressed on the
ground-truth scores) and a cost (regressed on the ground-truth costs). In all
cases the multi-output predictors fit each planner independently (sharing only
the input features). The two predictors are:
\begin{itemize}
    \setlength{\itemsep}{0.15em}
    \item \textbf{OVR}: \emph{score} --- one \texttt{Ridge} regressor per planner,
    each fit to that planner's scores; \emph{cost} --- the same per-planner
    \texttt{Ridge}, here expressed as a \texttt{MultiOutputRegressor} over all
    planners' costs.
    \item \textbf{KNN}: \emph{score} --- a $k$-NN regressor (distance-weighted mean of
    the $k$ nearest training neighbors' scores, per planner); \emph{cost} --- a
    second $k$-NN regressor over the same neighbors, predicting each planner's cost.
    \item \textbf{PR-KNN}: \emph{score} --- distance-weighted mean of the $k$ neighbors'
   scores per planner; \emph{cost} --- distance-weighted mean of those same
    neighbors' costs.
    \item \textbf{K-means}: each prototype stores a per-planner \emph{score} lookup (the
    mean score of the samples assigned to that planner) and a per-planner \emph{cost}
    lookup (the mean cost of that group); a query's predicted score and cost are the
    similarity-weighted combinations of these lookups across prototypes.
    \item \textbf{Linear}: \emph{score} --- a linear model with one output per planner
    regressed on ground-truth scores (per-planner \texttt{Ridge} on CPU, or a linear layer with
    MSE loss on GPU); \emph{cost} --- a separate per-planner \texttt{Ridge} fit to costs.
    \item \textbf{MLP}: \emph{score} --- a multilayer perceptron with one output per
    planner trained as a regressor (MSE) on ground-truth scores; \emph{cost} --- a separate
    per-planner \texttt{Ridge} fit to costs.
\end{itemize}
Given the predicted per-planner score $\hat{q}_{i,k}$ and cost $\hat{c}_{i,k}$, the planner
is then chosen by the selection rule (base, tie, or rank).

\subsection{Routing Cost and Latency}
\label{app:router:cost}

A routing decision costs the embedding extraction plus the router's own forward pass, and
both are negligible relative to the planner (seconds to tens of seconds per query), so the
estimates here are deliberately coarse. Router inputs are formed from a text encoder
(BGE-M3)~\citep{chen-etal-2024-m3} over the instruction, a vision encoder over the observation
(we ablate SigLIP~\citep{zhai2023siglip}, two SigLIP-2~\citep{tschannen2025siglip}
variants, and DINOv2~\citep{oquab2024dinov2}); extraction dominates and is itself dominated by the vision
encoder, with text embeddings amortizing to a few milliseconds per sample and a single
vision embedding estimated on the order of $\sim$10--25\,ms if optimized
(observed $\sim$25--45\,ms amortized in our un-optimized batched runs; a $\pi_0$-based VLA encoder \cite{black2025pi0}, which we include only for ablation studies over fusion method, adds additional overhead). The trained routers
themselves are extremely light--- on the order of $0.0088$--$1.82$\,ms (amortized) per sample across architectures (derived from latency profiling for GLM 4.1V Thinking / GLM 4.6V Flash routing decisions over all router architectures, ablated over the $4$ vision encoders using the $\texttt{normalize\_concat}$ fusion method. The GPU was pre-warmed and a batch of size 16 was run 50 times per router/ablation). For fused text and vision embeddings, a routing decision totals roughly $\sim$20--50\,ms amortized per query, two to three orders
of magnitude below planner latency; embeddings could additionally be cached and reused across
repeated queries on the same observation for further optimization.

\subsection{Per-Experiment Configuration}
\label{app:configs}

Table~\ref{tab:app:exp_configs} records the vision encoder and fusion method used for
each experiment in the main paper and appendix.

\begin{table}[h]
\centering
\caption{Vision encoder and fusion method per experiment.}
\label{tab:app:exp_configs}
\footnotesize
\begin{tabular}{@{}lll@{}}
\toprule
\textbf{Experiment} & \textbf{Fusion} & \textbf{Vision encoder} \\
\midrule
Chain-of-Thought (main Table~1) & \texttt{normalize\_concat} & SigLIP-2 (large) \\
Qwen size scaling & \texttt{normalize\_concat} & SigLIP (large)\\
RoboMME (memory) & \texttt{concat} & SigLIP (large)\\
Hardware (Franka/DROID) & \texttt{normalize\_concat} & SigLIP (large)\\
\bottomrule
\end{tabular}
\end{table}

\section{Router Ablations: Architecture, Fusion, Vision Encoder, and Heads}
\label{app:ablations}

For each routing model pair we ablate the router design space along three axes --- the
router \emph{architecture} (linear, MLP, KNN, PR-KNN, K-means, OVR;
Appendix~\ref{app:router:arch}), the embedding \emph{fusion} method
(Appendix~\ref{app:router:embeddings}), and the \emph{vision encoder} --- and present two
tables per pair. The first is a \textbf{router-architecture ablation}: fixing the embedding
configuration to the deployed setting (\texttt{normalize\_concat} fusion with the SigLIP-2 (large)
encoder; Appendix~\ref{app:configs}), it reports the best router per architecture by $\eta$,
the routing-efficiency metric of the main paper (the preference-weighted harmonic mean of
scaled quality and cost, reported at $\beta=0.1$). The second is a combined \textbf{embedding
ablation} that sweeps (i) the fusion method (vision encoder fixed) and (ii) the vision
encoder (fusion fixed), reporting the best router per setting by average score and by $\eta$;
the decision \emph{head} (classification vs.\ the regression variants;
Appendix~\ref{app:router:heads}) and the architecture are encoded in each router's name
(convention below). All tables also list the fixed baselines and oracles
(Appendix~\ref{app:baselines}); average score and $\eta$ are multiplied by $100$ and cost is
in seconds. Router inference hyperparameters are in
Appendix~\ref{app:router:hyperparams}, and the complete per-router sweep for a
representative pair is in Table~\ref{tab:app:full_router_sweep}
(Appendix~\ref{app:full_sweep}).

\paragraph{Router naming convention.}
\label{app:router_naming}
Each trained router is specified by its configuration: \texttt{\{architecture\} \{fusion\} \{head\}}:

\begin{itemize}
    \item \textit{architecture}: \texttt{linear}, \texttt{mlp} (with hidden sizes, e.g., \texttt{h2048\_1024}), \texttt{knn} (with neighbor count, e.g., \texttt{k5}), \texttt{prknn}, \texttt{kmeans}, or \texttt{ovr} (Appendix B.2).
    \item \textit{fusion}: the embedding fusion strategy (Appendix B.1): \texttt{normalize\_concat}, \texttt{concat}, \texttt{average}, \texttt{weighted\_average\_w0.25\_w0.75}, \texttt{only\_text}, \texttt{only\_image}, or \texttt{only\_vla}. This duplicates the “Fusion Method” / “Vision Encoder” column of each table.
    \item \textit{head}: the router head / decision rule (Appendix B.3): no suffix denotes classification; \texttt{regression}, \texttt{rank\_regression}, and \texttt{tie\_regression} denote the regression variants; \texttt{soft\_$\lambda$0.1} denotes the soft-label linear router with $\lambda = 0.1$.
\end{itemize}

\subsection{Router Inference Hyperparameters}
\label{app:router:hyperparams}

During inference, standard classification routers employ conventional decision rules: $\arg\max$ over probabilities or logits for Linear, MLP, and OVR models, and majority voting for KNN. For alternative formulations, PR-KNN utilizes Copeland-score voting to aggregate pairwise preferences, while K-means assigns inputs to the nearest centroid to retrieve its learned planner preference. Note that regression variants (\texttt{-{}-regression}, \texttt{-{}-rank\_regression}, \texttt{-{}-tie\_regression}) share these base architectures but modify the inference rule to suit their training signals (detailed in Appendix~\ref{app:router:preference}). 

Table~\ref{tab:app:router_hparams} lists the core configuration and hyperparameters for each base router architecture.

\begin{table}[h]
\centering
\caption{Router architecture hyperparameters (base classification configuration).}
\label{tab:app:router_hparams}
\footnotesize
\begin{tabular}{@{}ll@{}}
\toprule
\textbf{Router} & \textbf{Hyperparameters} \\
\midrule
KNN & $k=5$, fusion \texttt{concat} \\
\addlinespace[2pt]
Linear & penalty \texttt{l2}, $C=1.0$, \texttt{train\_lambda}$=0.1$, seed $=42$ \\
\addlinespace[2pt]
MLP & hidden $=[2048, 1024]$, \texttt{relu}, $\alpha = 10^{-4}$ (L2), seed $=42$ \\
\addlinespace[2pt]
OVR & one \texttt{LogisticRegression} / planner (sklearn default: L2, $C=1.0$) \\
\addlinespace[2pt]
PR-KNN & $k=5$, metric \texttt{cosine}, weights \texttt{distance} \\
\addlinespace[2pt]
K-means & \texttt{normalize}=True (L2), \texttt{lambda\_cost}$=0.0$ \\
\bottomrule
\end{tabular}
\end{table}
\subsection{GLM-4.1V-Thinking vs.\ GLM-4.6V-Flash}
\label{app:ablations:GLM4_1V_Thinking_GLM4_6V_Flash}
\begin{table}[H]
\centering
\caption{Router-architecture ablation (embedding fusion fixed to \texttt{normalize\_concat}, vision encoder SigLIP-2 --- the deployed configuration, App.~\ref{app:configs}). Per architecture we report the best router by $\eta$, with the fixed baselines and oracles. avg\_score / $\eta$ $\times100$; avg\_cost in seconds; $\eta$ cells shaded \colorbox{red!25!white}{red} to \colorbox{green!25!white}{green}.}
\label{tab:arch_GLM4_1V_Thinking_GLM4_6V_Flash}
\footnotesize
\resizebox{\linewidth}{!}{%
\begin{tabular}{@{}llrrr l@{}}
\toprule
\textbf{Architecture} & \textbf{Selection Head} & \textbf{Avg Score} & \textbf{Avg Cost (s)} & \textbf{$\eta$} & \textbf{Category} \\
\midrule
Linear & soft-label ($\lambda$=0.1) & 51.79 & 31.82 & \cellcolor{green!75!red!25!white} 75.26 & Trained Router \\
MLP [2048,1024] & classification & 50.19 & 30.13 & \cellcolor{green!73!red!25!white} 73.12 & Trained Router \\
KNN ($k$=5) & classification & 51.27 & 31.99 & \cellcolor{green!74!red!25!white} 73.83 & Trained Router \\
PR-KNN ($k$=5) & classification & 51.27 & 31.99 & \cellcolor{green!74!red!25!white} 73.83 & Trained Router \\
K-means & classification & 48.51 & 26.20 & \cellcolor{green!71!red!25!white} 71.41 & Trained Router \\
OVR & tie regression & 49.57 & 31.61 & \cellcolor{green!70!red!25!white} 70.08 & Trained Router \\
\midrule
expensive global & --- & 54.02 & 45.10 & \cellcolor{green!0!red!25!white} 0.00 & Fixed Baseline \\
strongest global & --- & 54.02 & 45.10 & \cellcolor{green!0!red!25!white} 0.00 & Fixed Baseline \\
strongest per category & --- & 54.02 & 45.10 & \cellcolor{green!0!red!25!white} 0.00 & Fixed Baseline \\
random & --- & 40.02 & 22.39 & \cellcolor{green!46!red!25!white} 46.33 & Fixed Baseline \\
cheapest global & --- & 27.24 & 2.11 & \cellcolor{green!0!red!25!white} 0.00 & Fixed Baseline \\
ood baseline & --- & 49.83 & 39.60 & \cellcolor{green!55!red!25!white} 54.52 & OOD Baseline \\
\midrule
accuracy oracle & --- & 55.17 & 27.96 & \cellcolor{green!88!red!25!white} 87.94 & Oracle \\
$\eta$ oracle & --- & 55.17 & 33.32 & \cellcolor{green!81!red!25!white} 80.59 & Oracle \\
\bottomrule
\end{tabular}}
\end{table}

\begin{table}[H]
\centering
\caption{Embedding ablation: (i) embedding-fusion method (vision encoder fixed to SigLIP-2) and (ii) vision encoder (fusion fixed to \texttt{normalize\_concat}). Best router per setting by average score and by $\eta$. \emph{text} / \emph{vision} / \emph{policy embedding} are the single-modality fusions (instruction, scene, and $\pi_0$ VLA embeddings). avg\_score / $\eta$ $\times100$; avg\_cost in seconds; $\eta$ cells shaded \colorbox{red!25!white}{red} to \colorbox{green!25!white}{green}.}
\label{tab:embed_GLM4_1V_Thinking_GLM4_6V_Flash}
\footnotesize
\resizebox{\linewidth}{!}{%
\begin{tabular}{@{}llllrrr l@{}}
\toprule
\textbf{Fusion / Encoder} & \textbf{Selected by} & \textbf{Model Architecture} & \textbf{Selection Head} & \textbf{Avg Score} & \textbf{Avg Cost (s)} & \textbf{$\eta$} & \textbf{Category} \\
\midrule
\multicolumn{8}{@{}l}{\emph{(i) Embedding fusion --- vision encoder fixed (SigLIP-2)}} \\
\addlinespace[1pt]
average & best avg score & MLP [2048,1024] & regression & 53.47 & 38.52 & \cellcolor{green!64!red!25!white} 64.02 & Trained Router \\
average & best $\eta$ & Linear & tie regression & 49.51 & 25.83 & \cellcolor{green!74!red!25!white} 74.46 & Trained Router \\
concat & best avg score & MLP [2048,1024] & regression & 53.43 & 38.26 & \cellcolor{green!65!red!25!white} 64.91 & Trained Router \\
concat & best $\eta$ & Linear & soft-label ($\lambda$=0.1) & 51.79 & 31.82 & \cellcolor{green!75!red!25!white} 75.26 & Trained Router \\
normalize\_concat & best avg score & MLP [2048,1024] & regression & 53.63 & 38.80 & \cellcolor{green!63!red!25!white} 63.20 & Trained Router \\
normalize\_concat & best $\eta$ & Linear & soft-label ($\lambda$=0.1) & 51.79 & 31.82 & \cellcolor{green!75!red!25!white} 75.26 & Trained Router \\
\emph{vision embedding} & best avg score & MLP [2048,1024] & rank regression & 53.88 & 39.41 & \cellcolor{green!61!red!25!white} 61.02 & Trained Router \\
\emph{vision embedding} & best $\eta$ & MLP [2048,1024] & classification & 51.47 & 31.68 & \cellcolor{green!75!red!25!white} 74.68 & Trained Router \\
\emph{text embedding} & best avg score & MLP [2048,1024] & regression & 53.90 & 42.83 & \cellcolor{green!37!red!25!white} 37.41 & Trained Router \\
\emph{text embedding} & best $\eta$ & Linear & soft-label ($\lambda$=0.1) & 51.82 & 31.86 & \cellcolor{green!75!red!25!white} 75.29 & Trained Router \\
\emph{policy embedding} & best avg score & KNN ($k$=5) & rank regression & 53.74 & 43.41 & \cellcolor{green!31!red!25!white} 30.67 & Trained Router \\
\emph{policy embedding} & best $\eta$ & KNN ($k$=5) & classification & 50.94 & 32.34 & \cellcolor{green!73!red!25!white} 72.60 & Trained Router \\
weighted\_average\_text025 & best avg score & KNN ($k$=5) & rank regression & 53.51 & 42.93 & \cellcolor{green!36!red!25!white} 36.20 & Trained Router \\
weighted\_average\_text025 & best $\eta$ & Linear & soft-label ($\lambda$=0.1) & 52.17 & 33.23 & \cellcolor{green!74!red!25!white} 74.20 & Trained Router \\
weighted\_average\_text075 & best avg score & MLP [2048,1024] & regression & 53.63 & 43.12 & \cellcolor{green!34!red!25!white} 34.11 & Trained Router \\
weighted\_average\_text075 & best $\eta$ & Linear & soft-label ($\lambda$=0.1) & 51.69 & 32.07 & \cellcolor{green!75!red!25!white} 74.72 & Trained Router \\
\midrule
\multicolumn{8}{@{}l}{\emph{(ii) Vision encoder --- fusion fixed (\texttt{normalize\_concat})}} \\
\addlinespace[1pt]
dinov2-large & best avg score & MLP [2048,1024] & regression & 53.95 & 44.09 & \cellcolor{green!21!red!25!white} 20.76 & Trained Router \\
dinov2-large & best $\eta$ & Linear & tie regression & 49.91 & 27.20 & \cellcolor{green!75!red!25!white} 74.71 & Trained Router \\
siglip-large-patch16-384 & best avg score & MLP [2048,1024] & regression & 53.47 & 37.95 & \cellcolor{green!66!red!25!white} 66.02 & Trained Router \\
siglip-large-patch16-384 & best $\eta$ & Linear & tie regression & 49.92 & 26.96 & \cellcolor{green!75!red!25!white} 74.92 & Trained Router \\
siglip2-large-patch16-512 & best avg score & MLP [2048,1024] & regression & 53.63 & 38.80 & \cellcolor{green!63!red!25!white} 63.20 & Trained Router \\
siglip2-large-patch16-512 & best $\eta$ & Linear & soft-label ($\lambda$=0.1) & 51.79 & 31.82 & \cellcolor{green!75!red!25!white} 75.26 & Trained Router \\
siglip2-so400m-patch14-384 & best avg score & KNN ($k$=5) & regression & 53.56 & 38.40 & \cellcolor{green!65!red!25!white} 64.59 & Trained Router \\
siglip2-so400m-patch14-384 & best $\eta$ & Linear & tie regression & 49.77 & 26.81 & \cellcolor{green!75!red!25!white} 74.59 & Trained Router \\
\midrule
--- & --- & strongest per category & --- & 54.02 & 45.10 & \cellcolor{green!0!red!25!white} 0.00 & Fixed Baseline \\
--- & --- & expensive global & --- & 54.02 & 45.10 & \cellcolor{green!0!red!25!white} 0.00 & Fixed Baseline \\
--- & --- & strongest global & --- & 54.02 & 45.10 & \cellcolor{green!0!red!25!white} 0.00 & Fixed Baseline \\
--- & --- & random & --- & 40.02 & 22.39 & \cellcolor{green!46!red!25!white} 46.33 & Fixed Baseline \\
--- & --- & cheapest global & --- & 27.24 & 2.11 & \cellcolor{green!0!red!25!white} 0.00 & Fixed Baseline \\
--- & --- & ood baseline (average) & --- & 54.02 & 45.10 & \cellcolor{green!0!red!25!white} 0.00 & OOD Baseline \\
--- & --- & ood baseline (concat) & --- & 49.83 & 39.60 & \cellcolor{green!55!red!25!white} 54.52 & OOD Baseline \\
--- & --- & ood baseline (normalize\_concat) & --- & 49.83 & 39.60 & \cellcolor{green!55!red!25!white} 54.52 & OOD Baseline \\
--- & --- & ood baseline (vision embedding) & --- & 28.63 & 3.36 & \cellcolor{green!5!red!25!white} 5.44 & OOD Baseline \\
--- & --- & ood baseline (text embedding) & --- & 37.28 & 22.40 & \cellcolor{green!37!red!25!white} 37.03 & OOD Baseline \\
--- & --- & ood baseline (policy embedding) & --- & 54.02 & 45.10 & \cellcolor{green!0!red!25!white} 0.00 & OOD Baseline \\
--- & --- & ood baseline (weighted\_average\_text025) & --- & 54.02 & 45.10 & \cellcolor{green!0!red!25!white} 0.00 & OOD Baseline \\
--- & --- & ood baseline (weighted\_average\_text075) & --- & 28.13 & 3.92 & \cellcolor{green!4!red!25!white} 3.50 & OOD Baseline \\
--- & --- & ood baseline (dinov2-large) & --- & 49.27 & 39.65 & \cellcolor{green!54!red!25!white} 53.52 & OOD Baseline \\
--- & --- & ood baseline (siglip-large-patch16-384) & --- & 49.11 & 38.96 & \cellcolor{green!56!red!25!white} 55.64 & OOD Baseline \\
--- & --- & ood baseline (siglip2-large-patch16-512) & --- & 49.83 & 39.60 & \cellcolor{green!55!red!25!white} 54.52 & OOD Baseline \\
--- & --- & ood baseline (siglip2-so400m-patch14-384) & --- & 49.74 & 39.46 & \cellcolor{green!55!red!25!white} 54.92 & OOD Baseline \\
\midrule
--- & --- & accuracy oracle & --- & 55.17 & 27.96 & \cellcolor{green!88!red!25!white} 87.94 & Oracle \\
--- & --- & $\eta$ oracle & --- & 55.17 & 33.32 & \cellcolor{green!81!red!25!white} 80.59 & Oracle \\
\bottomrule
\end{tabular}}
\end{table}

\subsection{GLM-4.1V-Thinking vs.\ RynnBrain-8B-Plan}
\label{app:ablations:GLM4_1V_Thinking_RynnBrain_8B_Plan}
\begin{table}[H]
\centering
\caption{Router-architecture ablation (embedding fusion fixed to \texttt{normalize\_concat}, vision encoder SigLIP-2 --- the deployed configuration, App.~\ref{app:configs}). Per architecture we report the best router by $\eta$, with the fixed baselines and oracles. avg\_score / $\eta$ $\times100$; avg\_cost in seconds; $\eta$ cells shaded \colorbox{red!25!white}{red} to \colorbox{green!25!white}{green}.}
\label{tab:arch_GLM4_1V_Thinking_RynnBrain_8B_Plan}
\footnotesize
\resizebox{\linewidth}{!}{%
\begin{tabular}{@{}llrrr l@{}}
\toprule
\textbf{Architecture} & \textbf{Selection Head} & \textbf{Avg Score} & \textbf{Avg Cost (s)} & \textbf{$\eta$} & \textbf{Category} \\
\midrule
Linear & regression & 54.33 & 27.50 & \cellcolor{green!64!red!25!white} 63.94 & Trained Router \\
MLP [2048,1024] & regression & 54.11 & 26.20 & \cellcolor{green!63!red!25!white} 63.18 & Trained Router \\
KNN ($k$=5) & regression & 53.23 & 27.16 & \cellcolor{green!57!red!25!white} 57.33 & Trained Router \\
PR-KNN ($k$=5) & regression & 53.23 & 27.16 & \cellcolor{green!57!red!25!white} 57.33 & Trained Router \\
K-means & classification & 52.23 & 27.91 & \cellcolor{green!51!red!25!white} 50.66 & Trained Router \\
OVR & regression & 53.21 & 29.06 & \cellcolor{green!56!red!25!white} 56.31 & Trained Router \\
\midrule
strongest per category & --- & 54.29 & 42.16 & \cellcolor{green!37!red!25!white} 37.19 & Fixed Baseline \\
expensive global & --- & 54.02 & 45.10 & \cellcolor{green!0!red!25!white} 0.00 & Fixed Baseline \\
strongest global & --- & 54.02 & 45.10 & \cellcolor{green!0!red!25!white} 0.00 & Fixed Baseline \\
random & --- & 49.44 & 22.17 & \cellcolor{green!33!red!25!white} 32.51 & Fixed Baseline \\
cheapest global & --- & 45.25 & 1.67 & \cellcolor{green!0!red!25!white} 0.00 & Fixed Baseline \\
ood baseline & --- & 52.56 & 32.72 & \cellcolor{green!50!red!25!white} 50.42 & OOD Baseline \\
\midrule
accuracy oracle & --- & 58.64 & 17.65 & \cellcolor{green!95!red!25!white} 94.98 & Oracle \\
$\eta$ oracle & --- & 58.60 & 22.90 & \cellcolor{green!92!red!25!white} 91.79 & Oracle \\
\bottomrule
\end{tabular}}
\end{table}

\begin{table}[H]
\centering
\caption{Embedding ablation: (i) embedding-fusion method (vision encoder fixed to SigLIP-2) and (ii) vision encoder (fusion fixed to \texttt{normalize\_concat}). Best router per setting by average score and by $\eta$. \emph{text} / \emph{vision} / \emph{policy embedding} are the single-modality fusions (instruction, scene, and $\pi_0$ VLA embeddings). avg\_score / $\eta$ $\times100$; avg\_cost in seconds; $\eta$ cells shaded \colorbox{red!25!white}{red} to \colorbox{green!25!white}{green}.}
\label{tab:embed_GLM4_1V_Thinking_RynnBrain_8B_Plan}
\footnotesize
\resizebox{\linewidth}{!}{%
\begin{tabular}{@{}llllrrr l@{}}
\toprule
\textbf{Fusion / Encoder} & \textbf{Selected by} & \textbf{Model Architecture} & \textbf{Selection Head} & \textbf{Avg Score} & \textbf{Avg Cost (s)} & \textbf{$\eta$} & \textbf{Category} \\
\midrule
\multicolumn{8}{@{}l}{\emph{(i) Embedding fusion --- vision encoder fixed (SigLIP-2)}} \\
\addlinespace[1pt]
average & best avg score & Linear & regression & 53.93 & 27.98 & \cellcolor{green!61!red!25!white} 61.25 & Trained Router \\
average & best $\eta$ & MLP [2048,1024] & regression & 53.70 & 24.47 & \cellcolor{green!61!red!25!white} 61.30 & Trained Router \\
concat & best avg score & Linear & regression & 54.33 & 27.49 & \cellcolor{green!64!red!25!white} 63.94 & Trained Router \\
concat & best $\eta$ & Linear & regression & 54.33 & 27.49 & \cellcolor{green!64!red!25!white} 63.94 & Trained Router \\
normalize\_concat & best avg score & Linear & regression & 54.33 & 27.50 & \cellcolor{green!64!red!25!white} 63.94 & Trained Router \\
normalize\_concat & best $\eta$ & Linear & regression & 54.33 & 27.50 & \cellcolor{green!64!red!25!white} 63.94 & Trained Router \\
\emph{vision embedding} & best avg score & MLP [2048,1024] & regression & 53.71 & 39.50 & \cellcolor{green!47!red!25!white} 46.66 & Trained Router \\
\emph{vision embedding} & best $\eta$ & Linear & regression & 53.24 & 26.84 & \cellcolor{green!57!red!25!white} 57.49 & Trained Router \\
\emph{text embedding} & best avg score & MLP [2048,1024] & regression & 54.32 & 34.49 & \cellcolor{green!58!red!25!white} 58.37 & Trained Router \\
\emph{text embedding} & best $\eta$ & Linear & regression & 53.68 & 27.45 & \cellcolor{green!60!red!25!white} 59.98 & Trained Router \\
\emph{policy embedding} & best avg score & Linear & regression & 53.57 & 32.57 & \cellcolor{green!56!red!25!white} 56.27 & Trained Router \\
\emph{policy embedding} & best $\eta$ & KNN ($k$=5) & regression & 53.08 & 25.78 & \cellcolor{green!57!red!25!white} 56.88 & Trained Router \\
weighted\_average\_text025 & best avg score & MLP [2048,1024] & regression & 54.32 & 33.22 & \cellcolor{green!60!red!25!white} 59.73 & Trained Router \\
weighted\_average\_text025 & best $\eta$ & Linear & regression & 53.93 & 28.29 & \cellcolor{green!61!red!25!white} 61.09 & Trained Router \\
weighted\_average\_text075 & best avg score & Linear & regression & 53.83 & 27.81 & \cellcolor{green!61!red!25!white} 60.72 & Trained Router \\
weighted\_average\_text075 & best $\eta$ & Linear & regression & 53.83 & 27.81 & \cellcolor{green!61!red!25!white} 60.72 & Trained Router \\
\midrule
\multicolumn{8}{@{}l}{\emph{(ii) Vision encoder --- fusion fixed (\texttt{normalize\_concat})}} \\
\addlinespace[1pt]
dinov2-large & best avg score & MLP [2048,1024] & regression & 54.04 & 30.74 & \cellcolor{green!60!red!25!white} 60.26 & Trained Router \\
dinov2-large & best $\eta$ & Linear & regression & 53.94 & 28.24 & \cellcolor{green!61!red!25!white} 61.16 & Trained Router \\
siglip-large-patch16-384 & best avg score & Linear & regression & 54.10 & 27.38 & \cellcolor{green!63!red!25!white} 62.61 & Trained Router \\
siglip-large-patch16-384 & best $\eta$ & Linear & regression & 54.10 & 27.38 & \cellcolor{green!63!red!25!white} 62.61 & Trained Router \\
siglip2-large-patch16-512 & best avg score & Linear & regression & 54.33 & 27.50 & \cellcolor{green!64!red!25!white} 63.94 & Trained Router \\
siglip2-large-patch16-512 & best $\eta$ & Linear & regression & 54.33 & 27.50 & \cellcolor{green!64!red!25!white} 63.94 & Trained Router \\
siglip2-so400m-patch14-384 & best avg score & Linear & regression & 54.34 & 27.35 & \cellcolor{green!64!red!25!white} 64.04 & Trained Router \\
siglip2-so400m-patch14-384 & best $\eta$ & Linear & regression & 54.34 & 27.35 & \cellcolor{green!64!red!25!white} 64.04 & Trained Router \\
\midrule
--- & --- & strongest per category & --- & 54.29 & 42.16 & \cellcolor{green!37!red!25!white} 37.19 & Fixed Baseline \\
--- & --- & expensive global & --- & 54.02 & 45.10 & \cellcolor{green!0!red!25!white} 0.00 & Fixed Baseline \\
--- & --- & strongest global & --- & 54.02 & 45.10 & \cellcolor{green!0!red!25!white} 0.00 & Fixed Baseline \\
--- & --- & random & --- & 49.44 & 22.17 & \cellcolor{green!33!red!25!white} 32.51 & Fixed Baseline \\
--- & --- & cheapest global & --- & 45.25 & 1.67 & \cellcolor{green!0!red!25!white} 0.00 & Fixed Baseline \\
--- & --- & ood baseline (average) & --- & 45.25 & 1.67 & \cellcolor{green!0!red!25!white} 0.00 & OOD Baseline \\
--- & --- & ood baseline (concat) & --- & 52.56 & 32.72 & \cellcolor{green!50!red!25!white} 50.42 & OOD Baseline \\
--- & --- & ood baseline (normalize\_concat) & --- & 52.56 & 32.72 & \cellcolor{green!50!red!25!white} 50.42 & OOD Baseline \\
--- & --- & ood baseline (vision embedding) & --- & 45.25 & 1.67 & \cellcolor{green!0!red!25!white} 0.00 & OOD Baseline \\
--- & --- & ood baseline (text embedding) & --- & 45.25 & 1.67 & \cellcolor{green!0!red!25!white} 0.00 & OOD Baseline \\
--- & --- & ood baseline (policy embedding) & --- & 54.02 & 45.10 & \cellcolor{green!0!red!25!white} 0.00 & OOD Baseline \\
--- & --- & ood baseline (weighted\_average\_text025) & --- & 45.25 & 1.67 & \cellcolor{green!0!red!25!white} 0.00 & OOD Baseline \\
--- & --- & ood baseline (weighted\_average\_text075) & --- & 45.25 & 1.67 & \cellcolor{green!0!red!25!white} 0.00 & OOD Baseline \\
--- & --- & ood baseline (dinov2-large) & --- & 47.82 & 11.29 & \cellcolor{green!21!red!25!white} 20.63 & OOD Baseline \\
--- & --- & ood baseline (siglip-large-patch16-384) & --- & 52.48 & 31.60 & \cellcolor{green!51!red!25!white} 50.64 & OOD Baseline \\
--- & --- & ood baseline (siglip2-large-patch16-512) & --- & 52.56 & 32.72 & \cellcolor{green!50!red!25!white} 50.42 & OOD Baseline \\
--- & --- & ood baseline (siglip2-so400m-patch14-384) & --- & 52.48 & 31.99 & \cellcolor{green!50!red!25!white} 50.41 & OOD Baseline \\
\midrule
--- & --- & accuracy oracle & --- & 58.64 & 17.65 & \cellcolor{green!95!red!25!white} 94.98 & Oracle \\
--- & --- & $\eta$ oracle & --- & 58.60 & 22.90 & \cellcolor{green!92!red!25!white} 91.79 & Oracle \\
\bottomrule
\end{tabular}}
\end{table}

\subsection{GLM-4.6V-Flash-Thinking vs.\ GLM-4.6V-Flash}
\label{app:ablations:GLM4_6V_Flash_Thinking_GLM4_6V_Flash}
\begin{table}[H]
\centering
\caption{Router-architecture ablation (embedding fusion fixed to \texttt{normalize\_concat}, vision encoder SigLIP-2 --- the deployed configuration, App.~\ref{app:configs}). Per architecture we report the best router by $\eta$, with the fixed baselines and oracles. avg\_score / $\eta$ $\times100$; avg\_cost in seconds; $\eta$ cells shaded \colorbox{red!25!white}{red} to \colorbox{green!25!white}{green}.}
\label{tab:arch_GLM4_6V_Flash_Thinking_GLM4_6V_Flash}
\footnotesize
\resizebox{\linewidth}{!}{%
\begin{tabular}{@{}llrrr l@{}}
\toprule
\textbf{Architecture} & \textbf{Selection Head} & \textbf{Avg Score} & \textbf{Avg Cost (s)} & \textbf{$\eta$} & \textbf{Category} \\
\midrule
Linear & regression & 44.69 & 16.97 & \cellcolor{green!75!red!25!white} 74.92 & Trained Router \\
MLP [2048,1024] & tie regression & 43.59 & 15.89 & \cellcolor{green!73!red!25!white} 73.08 & Trained Router \\
KNN ($k$=5) & tie regression & 44.15 & 16.44 & \cellcolor{green!74!red!25!white} 74.14 & Trained Router \\
PR-KNN ($k$=5) & tie regression & 44.15 & 16.44 & \cellcolor{green!74!red!25!white} 74.14 & Trained Router \\
K-means & classification & 43.53 & 16.00 & \cellcolor{green!73!red!25!white} 72.68 & Trained Router \\
OVR & tie regression & 42.22 & 14.84 & \cellcolor{green!69!red!25!white} 69.27 & Trained Router \\
\midrule
expensive global & --- & 46.09 & 24.08 & \cellcolor{green!0!red!25!white} 0.00 & Fixed Baseline \\
strongest per category & --- & 46.09 & 24.08 & \cellcolor{green!0!red!25!white} 0.00 & Fixed Baseline \\
strongest global & --- & 46.09 & 24.08 & \cellcolor{green!0!red!25!white} 0.00 & Fixed Baseline \\
random & --- & 35.97 & 13.24 & \cellcolor{green!44!red!25!white} 43.68 & Fixed Baseline \\
cheapest global & --- & 27.24 & 2.11 & \cellcolor{green!0!red!25!white} 0.00 & Fixed Baseline \\
ood baseline & --- & 42.73 & 20.85 & \cellcolor{green!55!red!25!white} 55.34 & OOD Baseline \\
\midrule
accuracy oracle & --- & 47.47 & 13.37 & \cellcolor{green!91!red!25!white} 91.28 & Oracle \\
$\eta$ oracle & --- & 47.29 & 16.00 & \cellcolor{green!85!red!25!white} 84.94 & Oracle \\
\bottomrule
\end{tabular}}
\end{table}

\begin{table}[H]
\centering
\caption{Embedding ablation: (i) embedding-fusion method (vision encoder fixed to SigLIP-2) and (ii) vision encoder (fusion fixed to \texttt{normalize\_concat}). Best router per setting by average score and by $\eta$. \emph{text} / \emph{vision} / \emph{policy embedding} are the single-modality fusions (instruction, scene, and $\pi_0$ VLA embeddings). avg\_score / $\eta$ $\times100$; avg\_cost in seconds; $\eta$ cells shaded \colorbox{red!25!white}{red} to \colorbox{green!25!white}{green}.}
\label{tab:embed_GLM4_6V_Flash_Thinking_GLM4_6V_Flash}
\footnotesize
\resizebox{\linewidth}{!}{%
\begin{tabular}{@{}llllrrr l@{}}
\toprule
\textbf{Fusion / Encoder} & \textbf{Selected by} & \textbf{Model Architecture} & \textbf{Selection Head} & \textbf{Avg Score} & \textbf{Avg Cost (s)} & \textbf{$\eta$} & \textbf{Category} \\
\midrule
\multicolumn{8}{@{}l}{\emph{(i) Embedding fusion --- vision encoder fixed (SigLIP-2)}} \\
\addlinespace[1pt]
average & best avg score & KNN ($k$=5) & regression & 45.65 & 19.44 & \cellcolor{green!70!red!25!white} 69.94 & Trained Router \\
average & best $\eta$ & KNN ($k$=5) & tie regression & 43.83 & 15.60 & \cellcolor{green!74!red!25!white} 74.39 & Trained Router \\
concat & best avg score & KNN ($k$=5) & rank regression & 45.73 & 22.68 & \cellcolor{green!41!red!25!white} 41.31 & Trained Router \\
concat & best $\eta$ & Linear & regression & 44.69 & 16.97 & \cellcolor{green!75!red!25!white} 74.92 & Trained Router \\
normalize\_concat & best avg score & KNN ($k$=5) & rank regression & 45.73 & 22.68 & \cellcolor{green!41!red!25!white} 41.31 & Trained Router \\
normalize\_concat & best $\eta$ & Linear & regression & 44.69 & 16.97 & \cellcolor{green!75!red!25!white} 74.92 & Trained Router \\
\emph{vision embedding} & best avg score & MLP [2048,1024] & regression & 45.95 & 20.57 & \cellcolor{green!64!red!25!white} 64.44 & Trained Router \\
\emph{vision embedding} & best $\eta$ & KNN ($k$=5) & tie regression & 44.24 & 16.21 & \cellcolor{green!75!red!25!white} 74.88 & Trained Router \\
\emph{text embedding} & best avg score & MLP [2048,1024] & regression & 45.38 & 22.82 & \cellcolor{green!38!red!25!white} 38.45 & Trained Router \\
\emph{text embedding} & best $\eta$ & Linear & regression & 44.86 & 17.56 & \cellcolor{green!74!red!25!white} 74.05 & Trained Router \\
\emph{policy embedding} & best avg score & KNN ($k$=5) & regression & 45.92 & 19.85 & \cellcolor{green!69!red!25!white} 68.62 & Trained Router \\
\emph{policy embedding} & best $\eta$ & KNN ($k$=5) & tie regression & 43.83 & 16.03 & \cellcolor{green!74!red!25!white} 73.73 & Trained Router \\
weighted\_average\_text025 & best avg score & KNN ($k$=5) & rank regression & 45.74 & 22.66 & \cellcolor{green!42!red!25!white} 41.51 & Trained Router \\
weighted\_average\_text025 & best $\eta$ & KNN ($k$=5) & tie regression & 44.04 & 16.08 & \cellcolor{green!74!red!25!white} 74.37 & Trained Router \\
weighted\_average\_text075 & best avg score & KNN ($k$=5) & regression & 45.40 & 19.46 & \cellcolor{green!69!red!25!white} 69.19 & Trained Router \\
weighted\_average\_text075 & best $\eta$ & KNN ($k$=5) & tie regression & 43.88 & 15.63 & \cellcolor{green!75!red!25!white} 74.55 & Trained Router \\
\midrule
\multicolumn{8}{@{}l}{\emph{(ii) Vision encoder --- fusion fixed (\texttt{normalize\_concat})}} \\
\addlinespace[1pt]
dinov2-large & best avg score & KNN ($k$=5) & regression & 45.71 & 19.59 & \cellcolor{green!69!red!25!white} 69.41 & Trained Router \\
dinov2-large & best $\eta$ & Linear & regression & 45.02 & 17.31 & \cellcolor{green!75!red!25!white} 75.22 & Trained Router \\
siglip-large-patch16-384 & best avg score & KNN ($k$=5) & regression & 45.47 & 19.38 & \cellcolor{green!70!red!25!white} 69.72 & Trained Router \\
siglip-large-patch16-384 & best $\eta$ & Linear & regression & 44.91 & 16.93 & \cellcolor{green!76!red!25!white} 75.75 & Trained Router \\
siglip2-large-patch16-512 & best avg score & KNN ($k$=5) & rank regression & 45.73 & 22.68 & \cellcolor{green!41!red!25!white} 41.31 & Trained Router \\
siglip2-large-patch16-512 & best $\eta$ & Linear & regression & 44.69 & 16.97 & \cellcolor{green!75!red!25!white} 74.92 & Trained Router \\
siglip2-so400m-patch14-384 & best avg score & MLP [2048,1024] & regression & 46.10 & 23.13 & \cellcolor{green!32!red!25!white} 32.45 & Trained Router \\
siglip2-so400m-patch14-384 & best $\eta$ & Linear & regression & 44.69 & 16.97 & \cellcolor{green!75!red!25!white} 74.93 & Trained Router \\
\midrule
--- & --- & strongest global & --- & 46.09 & 24.08 & \cellcolor{green!0!red!25!white} 0.00 & Fixed Baseline \\
--- & --- & expensive global & --- & 46.09 & 24.08 & \cellcolor{green!0!red!25!white} 0.00 & Fixed Baseline \\
--- & --- & strongest per category & --- & 46.09 & 24.08 & \cellcolor{green!0!red!25!white} 0.00 & Fixed Baseline \\
--- & --- & random & --- & 35.97 & 13.24 & \cellcolor{green!44!red!25!white} 43.68 & Fixed Baseline \\
--- & --- & cheapest global & --- & 27.24 & 2.11 & \cellcolor{green!0!red!25!white} 0.00 & Fixed Baseline \\
--- & --- & ood baseline (average) & --- & 46.09 & 24.08 & \cellcolor{green!0!red!25!white} 0.00 & OOD Baseline \\
--- & --- & ood baseline (concat) & --- & 42.73 & 20.85 & \cellcolor{green!55!red!25!white} 55.34 & OOD Baseline \\
--- & --- & ood baseline (normalize\_concat) & --- & 42.73 & 20.85 & \cellcolor{green!55!red!25!white} 55.34 & OOD Baseline \\
--- & --- & ood baseline (vision embedding) & --- & 28.07 & 2.63 & \cellcolor{green!5!red!25!white} 4.51 & OOD Baseline \\
--- & --- & ood baseline (text embedding) & --- & 34.47 & 13.06 & \cellcolor{green!37!red!25!white} 36.70 & OOD Baseline \\
--- & --- & ood baseline (policy embedding) & --- & 46.09 & 24.08 & \cellcolor{green!0!red!25!white} 0.00 & OOD Baseline \\
--- & --- & ood baseline (weighted\_average\_text025) & --- & 46.09 & 24.08 & \cellcolor{green!0!red!25!white} 0.00 & OOD Baseline \\
--- & --- & ood baseline (weighted\_average\_text075) & --- & 28.00 & 3.16 & \cellcolor{green!4!red!25!white} 4.11 & OOD Baseline \\
--- & --- & ood baseline (dinov2-large) & --- & 42.76 & 21.39 & \cellcolor{green!52!red!25!white} 51.86 & OOD Baseline \\
--- & --- & ood baseline (siglip-large-patch16-384) & --- & 42.31 & 20.57 & \cellcolor{green!56!red!25!white} 55.89 & OOD Baseline \\
--- & --- & ood baseline (siglip2-large-patch16-512) & --- & 42.73 & 20.85 & \cellcolor{green!55!red!25!white} 55.34 & OOD Baseline \\
--- & --- & ood baseline (siglip2-so400m-patch14-384) & --- & 42.70 & 20.77 & \cellcolor{green!56!red!25!white} 55.73 & OOD Baseline \\
\midrule
--- & --- & accuracy oracle & --- & 47.47 & 13.37 & \cellcolor{green!91!red!25!white} 91.28 & Oracle \\
--- & --- & $\eta$ oracle & --- & 47.29 & 16.00 & \cellcolor{green!85!red!25!white} 84.94 & Oracle \\
\bottomrule
\end{tabular}}
\end{table}

\subsection{Gemini-3-Flash vs.\ Gemini-3-Flash-MinimalThinking}
\label{app:ablations:Gemini3Flash_Gemini3Flash_MinimalThinking}
\begin{table}[H]
\centering
\caption{Router-architecture ablation (embedding fusion fixed to \texttt{normalize\_concat}, vision encoder SigLIP-2 --- the deployed configuration, App.~\ref{app:configs}). Per architecture we report the best router by $\eta$, with the fixed baselines and oracles. avg\_score / $\eta$ $\times100$; avg\_cost in seconds; $\eta$ cells shaded \colorbox{red!25!white}{red} to \colorbox{green!25!white}{green}.}
\label{tab:arch_Gemini3Flash_Gemini3Flash_MinimalThinking}
\footnotesize
\resizebox{\linewidth}{!}{%
\begin{tabular}{@{}llrrr l@{}}
\toprule
\textbf{Architecture} & \textbf{Selection Head} & \textbf{Avg Score} & \textbf{Avg Cost (s)} & \textbf{$\eta$} & \textbf{Category} \\
\midrule
Linear & regression & 64.78 & 8.76 & \cellcolor{green!57!red!25!white} 57.04 & Trained Router \\
MLP [2048,1024] & regression & 64.83 & 8.47 & \cellcolor{green!58!red!25!white} 57.99 & Trained Router \\
KNN ($k$=5) & regression & 64.57 & 7.25 & \cellcolor{green!55!red!25!white} 54.86 & Trained Router \\
PR-KNN ($k$=5) & regression & 64.57 & 7.25 & \cellcolor{green!55!red!25!white} 54.86 & Trained Router \\
K-means & classification & 64.28 & 11.63 & \cellcolor{green!47!red!25!white} 46.82 & Trained Router \\
OVR & regression & 64.07 & 9.55 & \cellcolor{green!46!red!25!white} 46.10 & Trained Router \\
\midrule
cheapest global & --- & 61.22 & 1.88 & \cellcolor{green!0!red!25!white} 0.00 & Fixed Baseline \\
ood baseline & --- & 61.83 & 6.70 & \cellcolor{green!11!red!25!white} 10.81 & OOD Baseline \\
strongest per category & --- & 64.58 & 12.24 & \cellcolor{green!50!red!25!white} 49.51 & Fixed Baseline \\
expensive global & --- & 64.22 & 15.86 & \cellcolor{green!0!red!25!white} 0.00 & Fixed Baseline \\
strongest global & --- & 64.22 & 15.86 & \cellcolor{green!0!red!25!white} 0.00 & Fixed Baseline \\
random & --- & 62.11 & 8.82 & \cellcolor{green!16!red!25!white} 15.53 & Fixed Baseline \\
\midrule
accuracy oracle & --- & 67.39 & 4.49 & \cellcolor{green!98!red!25!white} 97.96 & Oracle \\
$\eta$ oracle & --- & 67.31 & 6.23 & \cellcolor{green!95!red!25!white} 94.97 & Oracle \\
\bottomrule
\end{tabular}}
\end{table}

\begin{table}[H]
\centering
\caption{Embedding ablation: (i) embedding-fusion method (vision encoder fixed to SigLIP-2) and (ii) vision encoder (fusion fixed to \texttt{normalize\_concat}). Best router per setting by average score and by $\eta$. \emph{text} / \emph{vision} / \emph{policy embedding} are the single-modality fusions (instruction, scene, and $\pi_0$ VLA embeddings). avg\_score / $\eta$ $\times100$; avg\_cost in seconds; $\eta$ cells shaded \colorbox{red!25!white}{red} to \colorbox{green!25!white}{green}.}
\label{tab:embed_Gemini3Flash_Gemini3Flash_MinimalThinking}
\footnotesize
\resizebox{\linewidth}{!}{%
\begin{tabular}{@{}llllrrr l@{}}
\toprule
\textbf{Fusion / Encoder} & \textbf{Selected by} & \textbf{Model Architecture} & \textbf{Selection Head} & \textbf{Avg Score} & \textbf{Avg Cost (s)} & \textbf{$\eta$} & \textbf{Category} \\
\midrule
\multicolumn{8}{@{}l}{\emph{(i) Embedding fusion --- vision encoder fixed (SigLIP-2)}} \\
\addlinespace[1pt]
average & best avg score & Linear & regression & 64.95 & 8.60 & \cellcolor{green!60!red!25!white} 59.60 & Trained Router \\
average & best $\eta$ & Linear & regression & 64.95 & 8.60 & \cellcolor{green!60!red!25!white} 59.60 & Trained Router \\
concat & best avg score & MLP [2048,1024] & regression & 64.96 & 9.04 & \cellcolor{green!59!red!25!white} 59.28 & Trained Router \\
concat & best $\eta$ & MLP [2048,1024] & regression & 64.96 & 9.04 & \cellcolor{green!59!red!25!white} 59.28 & Trained Router \\
normalize\_concat & best avg score & MLP [2048,1024] & regression & 64.83 & 8.47 & \cellcolor{green!58!red!25!white} 57.99 & Trained Router \\
normalize\_concat & best $\eta$ & MLP [2048,1024] & regression & 64.83 & 8.47 & \cellcolor{green!58!red!25!white} 57.99 & Trained Router \\
\emph{vision embedding} & best avg score & Linear & regression & 64.76 & 10.50 & \cellcolor{green!55!red!25!white} 54.82 & Trained Router \\
\emph{vision embedding} & best $\eta$ & Linear & regression & 64.76 & 10.50 & \cellcolor{green!55!red!25!white} 54.82 & Trained Router \\
\emph{text embedding} & best avg score & Linear & regression & 64.84 & 8.66 & \cellcolor{green!58!red!25!white} 57.92 & Trained Router \\
\emph{text embedding} & best $\eta$ & Linear & regression & 64.84 & 8.66 & \cellcolor{green!58!red!25!white} 57.92 & Trained Router \\
\emph{policy embedding} & best avg score & Linear & regression & 64.58 & 10.16 & \cellcolor{green!53!red!25!white} 52.83 & Trained Router \\
\emph{policy embedding} & best $\eta$ & Linear & regression & 64.58 & 10.16 & \cellcolor{green!53!red!25!white} 52.83 & Trained Router \\
weighted\_average\_text025 & best avg score & Linear & regression & 64.85 & 9.09 & \cellcolor{green!58!red!25!white} 57.71 & Trained Router \\
weighted\_average\_text025 & best $\eta$ & Linear & regression & 64.85 & 9.09 & \cellcolor{green!58!red!25!white} 57.71 & Trained Router \\
weighted\_average\_text075 & best avg score & Linear & regression & 64.91 & 8.75 & \cellcolor{green!59!red!25!white} 58.84 & Trained Router \\
weighted\_average\_text075 & best $\eta$ & Linear & regression & 64.91 & 8.75 & \cellcolor{green!59!red!25!white} 58.84 & Trained Router \\
\midrule
\multicolumn{8}{@{}l}{\emph{(ii) Vision encoder --- fusion fixed (\texttt{normalize\_concat})}} \\
\addlinespace[1pt]
dinov2-large & best avg score & Linear & regression & 64.76 & 8.42 & \cellcolor{green!57!red!25!white} 57.02 & Trained Router \\
dinov2-large & best $\eta$ & Linear & regression & 64.76 & 8.42 & \cellcolor{green!57!red!25!white} 57.02 & Trained Router \\
siglip-large-patch16-384 & best avg score & Linear & regression & 64.75 & 8.60 & \cellcolor{green!57!red!25!white} 56.69 & Trained Router \\
siglip-large-patch16-384 & best $\eta$ & KNN ($k$=5) & regression & 64.72 & 7.63 & \cellcolor{green!57!red!25!white} 56.85 & Trained Router \\
siglip2-large-patch16-512 & best avg score & MLP [2048,1024] & regression & 64.83 & 8.47 & \cellcolor{green!58!red!25!white} 57.99 & Trained Router \\
siglip2-large-patch16-512 & best $\eta$ & MLP [2048,1024] & regression & 64.83 & 8.47 & \cellcolor{green!58!red!25!white} 57.99 & Trained Router \\
siglip2-so400m-patch14-384 & best avg score & Linear & regression & 64.90 & 9.01 & \cellcolor{green!58!red!25!white} 58.40 & Trained Router \\
siglip2-so400m-patch14-384 & best $\eta$ & Linear & regression & 64.90 & 9.01 & \cellcolor{green!58!red!25!white} 58.40 & Trained Router \\
\midrule
--- & --- & cheapest global & --- & 61.22 & 1.88 & \cellcolor{green!0!red!25!white} 0.00 & Fixed Baseline \\
--- & --- & ood baseline (average) & --- & 61.22 & 1.88 & \cellcolor{green!0!red!25!white} 0.00 & OOD Baseline \\
--- & --- & ood baseline (concat) & --- & 61.83 & 6.70 & \cellcolor{green!11!red!25!white} 10.81 & OOD Baseline \\
--- & --- & ood baseline (normalize\_concat) & --- & 61.83 & 6.70 & \cellcolor{green!11!red!25!white} 10.81 & OOD Baseline \\
--- & --- & ood baseline (vision embedding) & --- & 61.22 & 1.88 & \cellcolor{green!0!red!25!white} 0.00 & OOD Baseline \\
--- & --- & ood baseline (text embedding) & --- & 61.22 & 1.88 & \cellcolor{green!0!red!25!white} 0.00 & OOD Baseline \\
--- & --- & ood baseline (policy embedding) & --- & 64.22 & 15.86 & \cellcolor{green!0!red!25!white} 0.00 & OOD Baseline \\
--- & --- & ood baseline (weighted\_average\_text025) & --- & 61.22 & 1.88 & \cellcolor{green!0!red!25!white} 0.00 & OOD Baseline \\
--- & --- & ood baseline (weighted\_average\_text075) & --- & 61.22 & 1.88 & \cellcolor{green!0!red!25!white} 0.00 & OOD Baseline \\
--- & --- & ood baseline (dinov2-large) & --- & 61.23 & 1.89 & \cellcolor{green!0!red!25!white} 0.20 & OOD Baseline \\
--- & --- & ood baseline (siglip-large-patch16-384) & --- & 61.44 & 6.13 & \cellcolor{green!4!red!25!white} 3.94 & OOD Baseline \\
--- & --- & ood baseline (siglip2-large-patch16-512) & --- & 61.83 & 6.70 & \cellcolor{green!11!red!25!white} 10.81 & OOD Baseline \\
--- & --- & ood baseline (siglip2-so400m-patch14-384) & --- & 61.58 & 6.34 & \cellcolor{green!6!red!25!white} 6.37 & OOD Baseline \\
--- & --- & strongest per category & --- & 64.58 & 12.24 & \cellcolor{green!50!red!25!white} 49.51 & Fixed Baseline \\
--- & --- & strongest global & --- & 64.22 & 15.86 & \cellcolor{green!0!red!25!white} 0.00 & Fixed Baseline \\
--- & --- & expensive global & --- & 64.22 & 15.86 & \cellcolor{green!0!red!25!white} 0.00 & Fixed Baseline \\
--- & --- & random & --- & 62.11 & 8.82 & \cellcolor{green!16!red!25!white} 15.53 & Fixed Baseline \\
\midrule
--- & --- & accuracy oracle & --- & 67.39 & 4.49 & \cellcolor{green!98!red!25!white} 97.96 & Oracle \\
--- & --- & $\eta$ oracle & --- & 67.31 & 6.23 & \cellcolor{green!95!red!25!white} 94.97 & Oracle \\
\bottomrule
\end{tabular}}
\end{table}

\subsection{Gemini-Robotics-ER vs.\ Gemini-Robotics-ER-ThinkingOff}
\label{app:ablations:GeminiRoboticsER_GeminiRoboticsER_ThinkingOff}
\begin{table}[H]
\centering
\caption{Router-architecture ablation (embedding fusion fixed to \texttt{normalize\_concat}, vision encoder SigLIP-2 --- the deployed configuration, App.~\ref{app:configs}). Per architecture we report the best router by $\eta$, with the fixed baselines and oracles. avg\_score / $\eta$ $\times100$; avg\_cost in seconds; $\eta$ cells shaded \colorbox{red!25!white}{red} to \colorbox{green!25!white}{green}.}
\label{tab:arch_GeminiRoboticsER_GeminiRoboticsER_ThinkingOff}
\footnotesize
\resizebox{\linewidth}{!}{%
\begin{tabular}{@{}llrrr l@{}}
\toprule
\textbf{Architecture} & \textbf{Selection Head} & \textbf{Avg Score} & \textbf{Avg Cost (s)} & \textbf{$\eta$} & \textbf{Category} \\
\midrule
Linear & regression & 63.92 & 4.72 & \cellcolor{green!37!red!25!white} 37.09 & Trained Router \\
MLP [2048,1024] & rank regression & 63.66 & 3.71 & \cellcolor{green!31!red!25!white} 31.40 & Trained Router \\
KNN ($k$=5) & rank regression & 63.58 & 4.13 & \cellcolor{green!29!red!25!white} 29.23 & Trained Router \\
PR-KNN ($k$=5) & rank regression & 63.58 & 4.13 & \cellcolor{green!29!red!25!white} 29.23 & Trained Router \\
K-means & classification & 63.03 & 4.90 & \cellcolor{green!15!red!25!white} 15.17 & Trained Router \\
OVR & regression & 63.01 & 4.51 & \cellcolor{green!15!red!25!white} 14.70 & Trained Router \\
\midrule
cheapest global & --- & 62.88 & 2.31 & \cellcolor{green!11!red!25!white} 11.39 & Fixed Baseline \\
ood baseline & --- & 63.19 & 3.13 & \cellcolor{green!20!red!25!white} 19.68 & OOD Baseline \\
strongest global & --- & 62.88 & 2.31 & \cellcolor{green!11!red!25!white} 11.39 & Fixed Baseline \\
strongest per category & --- & 63.58 & 5.82 & \cellcolor{green!28!red!25!white} 27.57 & Fixed Baseline \\
random & --- & 62.46 & 4.67 & \cellcolor{green!0!red!25!white} 0.00 & Fixed Baseline \\
expensive global & --- & 62.62 & 7.02 & \cellcolor{green!0!red!25!white} 0.00 & Fixed Baseline \\
\midrule
accuracy oracle & --- & 66.48 & 2.90 & \cellcolor{green!99!red!25!white} 98.72 & Oracle \\
$\eta$ oracle & --- & 66.42 & 3.45 & \cellcolor{green!97!red!25!white} 97.10 & Oracle \\
\bottomrule
\end{tabular}}
\end{table}

\begin{table}[H]
\centering
\caption{Embedding ablation: (i) embedding-fusion method (vision encoder fixed to SigLIP-2) and (ii) vision encoder (fusion fixed to \texttt{normalize\_concat}). Best router per setting by average score and by $\eta$. \emph{text} / \emph{vision} / \emph{policy embedding} are the single-modality fusions (instruction, scene, and $\pi_0$ VLA embeddings). avg\_score / $\eta$ $\times100$; avg\_cost in seconds; $\eta$ cells shaded \colorbox{red!25!white}{red} to \colorbox{green!25!white}{green}.}
\label{tab:embed_GeminiRoboticsER_GeminiRoboticsER_ThinkingOff}
\footnotesize
\resizebox{\linewidth}{!}{%
\begin{tabular}{@{}llllrrr l@{}}
\toprule
\textbf{Fusion / Encoder} & \textbf{Selected by} & \textbf{Model Architecture} & \textbf{Selection Head} & \textbf{Avg Score} & \textbf{Avg Cost (s)} & \textbf{$\eta$} & \textbf{Category} \\
\midrule
\multicolumn{8}{@{}l}{\emph{(i) Embedding fusion --- vision encoder fixed (SigLIP-2)}} \\
\addlinespace[1pt]
average & best avg score & MLP [2048,1024] & regression & 63.98 & 4.67 & \cellcolor{green!39!red!25!white} 38.60 & Trained Router \\
average & best $\eta$ & MLP [2048,1024] & regression & 63.98 & 4.67 & \cellcolor{green!39!red!25!white} 38.60 & Trained Router \\
concat & best avg score & Linear & regression & 63.92 & 4.72 & \cellcolor{green!37!red!25!white} 37.09 & Trained Router \\
concat & best $\eta$ & Linear & regression & 63.92 & 4.72 & \cellcolor{green!37!red!25!white} 37.09 & Trained Router \\
normalize\_concat & best avg score & Linear & regression & 63.92 & 4.72 & \cellcolor{green!37!red!25!white} 37.09 & Trained Router \\
normalize\_concat & best $\eta$ & Linear & regression & 63.92 & 4.72 & \cellcolor{green!37!red!25!white} 37.09 & Trained Router \\
\emph{vision embedding} & best avg score & MLP [2048,1024] & regression & 63.89 & 4.16 & \cellcolor{green!37!red!25!white} 36.91 & Trained Router \\
\emph{vision embedding} & best $\eta$ & MLP [2048,1024] & regression & 63.89 & 4.16 & \cellcolor{green!37!red!25!white} 36.91 & Trained Router \\
\emph{text embedding} & best avg score & Linear & regression & 64.15 & 4.56 & \cellcolor{green!43!red!25!white} 42.67 & Trained Router \\
\emph{text embedding} & best $\eta$ & Linear & regression & 64.15 & 4.56 & \cellcolor{green!43!red!25!white} 42.67 & Trained Router \\
\emph{policy embedding} & best avg score & Linear & rank regression & 63.74 & 4.59 & \cellcolor{green!33!red!25!white} 32.92 & Trained Router \\
\emph{policy embedding} & best $\eta$ & Linear & rank regression & 63.74 & 4.59 & \cellcolor{green!33!red!25!white} 32.92 & Trained Router \\
weighted\_average\_text025 & best avg score & Linear & regression & 63.91 & 4.69 & \cellcolor{green!37!red!25!white} 36.90 & Trained Router \\
weighted\_average\_text025 & best $\eta$ & Linear & regression & 63.91 & 4.69 & \cellcolor{green!37!red!25!white} 36.90 & Trained Router \\
weighted\_average\_text075 & best avg score & Linear & regression & 63.92 & 4.67 & \cellcolor{green!37!red!25!white} 37.30 & Trained Router \\
weighted\_average\_text075 & best $\eta$ & Linear & regression & 63.92 & 4.67 & \cellcolor{green!37!red!25!white} 37.30 & Trained Router \\
\midrule
\multicolumn{8}{@{}l}{\emph{(ii) Vision encoder --- fusion fixed (\texttt{normalize\_concat})}} \\
\addlinespace[1pt]
dinov2-large & best avg score & Linear & regression & 63.86 & 4.74 & \cellcolor{green!36!red!25!white} 35.77 & Trained Router \\
dinov2-large & best $\eta$ & Linear & regression & 63.86 & 4.74 & \cellcolor{green!36!red!25!white} 35.77 & Trained Router \\
siglip-large-patch16-384 & best avg score & Linear & regression & 63.88 & 4.62 & \cellcolor{green!36!red!25!white} 36.26 & Trained Router \\
siglip-large-patch16-384 & best $\eta$ & Linear & regression & 63.88 & 4.62 & \cellcolor{green!36!red!25!white} 36.26 & Trained Router \\
siglip2-large-patch16-512 & best avg score & Linear & regression & 63.92 & 4.72 & \cellcolor{green!37!red!25!white} 37.09 & Trained Router \\
siglip2-large-patch16-512 & best $\eta$ & Linear & regression & 63.92 & 4.72 & \cellcolor{green!37!red!25!white} 37.09 & Trained Router \\
siglip2-so400m-patch14-384 & best avg score & MLP [2048,1024] & regression & 64.05 & 4.37 & \cellcolor{green!41!red!25!white} 40.59 & Trained Router \\
siglip2-so400m-patch14-384 & best $\eta$ & MLP [2048,1024] & regression & 64.05 & 4.37 & \cellcolor{green!41!red!25!white} 40.59 & Trained Router \\
\midrule
--- & --- & strongest global & --- & 62.88 & 2.31 & \cellcolor{green!11!red!25!white} 11.39 & Fixed Baseline \\
--- & --- & cheapest global & --- & 62.88 & 2.31 & \cellcolor{green!11!red!25!white} 11.39 & Fixed Baseline \\
--- & --- & ood baseline (average) & --- & 62.88 & 2.31 & \cellcolor{green!11!red!25!white} 11.39 & OOD Baseline \\
--- & --- & ood baseline (concat) & --- & 63.19 & 3.13 & \cellcolor{green!20!red!25!white} 19.68 & OOD Baseline \\
--- & --- & ood baseline (normalize\_concat) & --- & 63.19 & 3.13 & \cellcolor{green!20!red!25!white} 19.68 & OOD Baseline \\
--- & --- & ood baseline (vision embedding) & --- & 62.88 & 2.31 & \cellcolor{green!11!red!25!white} 11.39 & OOD Baseline \\
--- & --- & ood baseline (text embedding) & --- & 62.88 & 2.31 & \cellcolor{green!11!red!25!white} 11.39 & OOD Baseline \\
--- & --- & ood baseline (policy embedding) & --- & 62.62 & 7.02 & \cellcolor{green!0!red!25!white} 0.00 & OOD Baseline \\
--- & --- & ood baseline (weighted\_average\_text025) & --- & 62.88 & 2.31 & \cellcolor{green!11!red!25!white} 11.39 & OOD Baseline \\
--- & --- & ood baseline (weighted\_average\_text075) & --- & 62.88 & 2.31 & \cellcolor{green!11!red!25!white} 11.39 & OOD Baseline \\
--- & --- & ood baseline (dinov2-large) & --- & 62.88 & 2.31 & \cellcolor{green!11!red!25!white} 11.39 & OOD Baseline \\
--- & --- & ood baseline (siglip-large-patch16-384) & --- & 63.12 & 3.32 & \cellcolor{green!18!red!25!white} 17.72 & OOD Baseline \\
--- & --- & ood baseline (siglip2-large-patch16-512) & --- & 63.19 & 3.13 & \cellcolor{green!20!red!25!white} 19.68 & OOD Baseline \\
--- & --- & ood baseline (siglip2-so400m-patch14-384) & --- & 63.17 & 3.04 & \cellcolor{green!19!red!25!white} 18.96 & OOD Baseline \\
--- & --- & strongest per category & --- & 63.58 & 5.82 & \cellcolor{green!28!red!25!white} 27.57 & Fixed Baseline \\
--- & --- & random & --- & 62.46 & 4.67 & \cellcolor{green!0!red!25!white} 0.00 & Fixed Baseline \\
--- & --- & expensive global & --- & 62.62 & 7.02 & \cellcolor{green!0!red!25!white} 0.00 & Fixed Baseline \\
\midrule
--- & --- & accuracy oracle & --- & 66.48 & 2.90 & \cellcolor{green!99!red!25!white} 98.72 & Oracle \\
--- & --- & $\eta$ oracle & --- & 66.42 & 3.45 & \cellcolor{green!97!red!25!white} 97.10 & Oracle \\
\bottomrule
\end{tabular}}
\end{table}

\subsection{Qwen3-VL-8B-Thinking vs.\ Cosmos-Reason2}
\label{app:ablations:Qwen3_VL_8B_Thinking_Cosmos_Reason2}
\begin{table}[H]
\centering
\caption{Router-architecture ablation (embedding fusion fixed to \texttt{normalize\_concat}, vision encoder SigLIP-2 --- the deployed configuration, App.~\ref{app:configs}). Per architecture we report the best router by $\eta$, with the fixed baselines and oracles. avg\_score / $\eta$ $\times100$; avg\_cost in seconds; $\eta$ cells shaded \colorbox{red!25!white}{red} to \colorbox{green!25!white}{green}.}
\label{tab:arch_Qwen3_VL_8B_Thinking_Cosmos_Reason2}
\footnotesize
\resizebox{\linewidth}{!}{%
\begin{tabular}{@{}llrrr l@{}}
\toprule
\textbf{Architecture} & \textbf{Selection Head} & \textbf{Avg Score} & \textbf{Avg Cost (s)} & \textbf{$\eta$} & \textbf{Category} \\
\midrule
Linear & rank regression & 46.49 & 85.04 & \cellcolor{green!70!red!25!white} 70.16 & Trained Router \\
MLP [2048,1024] & classification & 46.26 & 82.39 & \cellcolor{green!70!red!25!white} 70.44 & Trained Router \\
KNN ($k$=5) & tie regression & 46.98 & 86.36 & \cellcolor{green!71!red!25!white} 71.16 & Trained Router \\
PR-KNN ($k$=5) & tie regression & 46.98 & 86.36 & \cellcolor{green!71!red!25!white} 71.16 & Trained Router \\
K-means & tie regression & 48.28 & 97.62 & \cellcolor{green!67!red!25!white} 67.49 & Trained Router \\
OVR & classification & 45.01 & 72.34 & \cellcolor{green!69!red!25!white} 68.80 & Trained Router \\
\midrule
strongest per category & --- & 48.28 & 112.34 & \cellcolor{green!44!red!25!white} 43.53 & Fixed Baseline \\
expensive global & --- & 48.28 & 120.76 & \cellcolor{green!0!red!25!white} 0.00 & Fixed Baseline \\
strongest global & --- & 48.28 & 120.76 & \cellcolor{green!0!red!25!white} 0.00 & Fixed Baseline \\
random & --- & 38.93 & 64.49 & \cellcolor{green!45!red!25!white} 45.03 & Fixed Baseline \\
cheapest global & --- & 29.56 & 2.02 & \cellcolor{green!0!red!25!white} 0.00 & Fixed Baseline \\
ood baseline & --- & 37.32 & 49.37 & \cellcolor{green!38!red!25!white} 38.47 & OOD Baseline \\
\midrule
accuracy oracle & --- & 50.47 & 70.83 & \cellcolor{green!89!red!25!white} 88.87 & Oracle \\
$\eta$ oracle & --- & 48.32 & 55.90 & \cellcolor{green!89!red!25!white} 88.76 & Oracle \\
\bottomrule
\end{tabular}}
\end{table}

\begin{table}[H]
\centering
\caption{Embedding ablation: (i) embedding-fusion method (vision encoder fixed to SigLIP-2) and (ii) vision encoder (fusion fixed to \texttt{normalize\_concat}). Best router per setting by average score and by $\eta$. \emph{text} / \emph{vision} / \emph{policy embedding} are the single-modality fusions (instruction, scene, and $\pi_0$ VLA embeddings). avg\_score / $\eta$ $\times100$; avg\_cost in seconds; $\eta$ cells shaded \colorbox{red!25!white}{red} to \colorbox{green!25!white}{green}.}
\label{tab:embed_Qwen3_VL_8B_Thinking_Cosmos_Reason2}
\footnotesize
\resizebox{\linewidth}{!}{%
\begin{tabular}{@{}llllrrr l@{}}
\toprule
\textbf{Fusion / Encoder} & \textbf{Selected by} & \textbf{Model Architecture} & \textbf{Selection Head} & \textbf{Avg Score} & \textbf{Avg Cost (s)} & \textbf{$\eta$} & \textbf{Category} \\
\midrule
\multicolumn{8}{@{}l}{\emph{(i) Embedding fusion --- vision encoder fixed (SigLIP-2)}} \\
\addlinespace[1pt]
average & best avg score & KNN ($k$=5) & regression & 48.46 & 97.32 & \cellcolor{green!68!red!25!white} 68.19 & Trained Router \\
average & best $\eta$ & KNN ($k$=5) & tie regression & 47.74 & 85.89 & \cellcolor{green!74!red!25!white} 73.79 & Trained Router \\
concat & best avg score & KNN ($k$=5) & regression & 48.46 & 98.26 & \cellcolor{green!67!red!25!white} 67.31 & Trained Router \\
concat & best $\eta$ & KNN ($k$=5) & tie regression & 46.98 & 86.36 & \cellcolor{green!71!red!25!white} 71.16 & Trained Router \\
normalize\_concat & best avg score & KNN ($k$=5) & regression & 48.46 & 98.26 & \cellcolor{green!67!red!25!white} 67.31 & Trained Router \\
normalize\_concat & best $\eta$ & KNN ($k$=5) & tie regression & 46.98 & 86.36 & \cellcolor{green!71!red!25!white} 71.16 & Trained Router \\
\emph{vision embedding} & best avg score & K-means & regression & 48.28 & 120.76 & \cellcolor{green!0!red!25!white} 0.00 & Trained Router \\
\emph{vision embedding} & best $\eta$ & Linear & rank regression & 47.62 & 88.74 & \cellcolor{green!72!red!25!white} 71.95 & Trained Router \\
\emph{text embedding} & best avg score & KNN ($k$=5) & regression & 48.60 & 94.40 & \cellcolor{green!71!red!25!white} 71.02 & Trained Router \\
\emph{text embedding} & best $\eta$ & PR-KNN ($k$=5) & classification & 46.72 & 80.46 & \cellcolor{green!73!red!25!white} 72.70 & Trained Router \\
\emph{policy embedding} & best avg score & KNN ($k$=5) & regression & 48.70 & 101.90 & \cellcolor{green!64!red!25!white} 63.88 & Trained Router \\
\emph{policy embedding} & best $\eta$ & KNN ($k$=5) & classification & 45.53 & 76.75 & \cellcolor{green!70!red!25!white} 69.67 & Trained Router \\
weighted\_average\_text025 & best avg score & KNN ($k$=5) & regression & 48.46 & 98.26 & \cellcolor{green!67!red!25!white} 67.31 & Trained Router \\
weighted\_average\_text025 & best $\eta$ & KNN ($k$=5) & tie regression & 47.60 & 87.28 & \cellcolor{green!73!red!25!white} 72.67 & Trained Router \\
weighted\_average\_text075 & best avg score & KNN ($k$=5) & regression & 48.46 & 95.18 & \cellcolor{green!70!red!25!white} 70.03 & Trained Router \\
weighted\_average\_text075 & best $\eta$ & KNN ($k$=5) & classification & 46.44 & 78.00 & \cellcolor{green!73!red!25!white} 72.55 & Trained Router \\
\midrule
\multicolumn{8}{@{}l}{\emph{(ii) Vision encoder --- fusion fixed (\texttt{normalize\_concat})}} \\
\addlinespace[1pt]
dinov2-large & best avg score & KNN ($k$=5) & regression & 48.46 & 98.26 & \cellcolor{green!67!red!25!white} 67.31 & Trained Router \\
dinov2-large & best $\eta$ & MLP [2048,1024] & classification & 47.00 & 82.04 & \cellcolor{green!73!red!25!white} 73.05 & Trained Router \\
siglip-large-patch16-384 & best avg score & KNN ($k$=5) & regression & 48.46 & 97.94 & \cellcolor{green!68!red!25!white} 67.62 & Trained Router \\
siglip-large-patch16-384 & best $\eta$ & KNN ($k$=5) & tie regression & 47.88 & 88.84 & \cellcolor{green!73!red!25!white} 72.68 & Trained Router \\
siglip2-large-patch16-512 & best avg score & KNN ($k$=5) & regression & 48.46 & 98.26 & \cellcolor{green!67!red!25!white} 67.31 & Trained Router \\
siglip2-large-patch16-512 & best $\eta$ & KNN ($k$=5) & tie regression & 46.98 & 86.36 & \cellcolor{green!71!red!25!white} 71.16 & Trained Router \\
siglip2-so400m-patch14-384 & best avg score & KNN ($k$=5) & regression & 48.39 & 96.94 & \cellcolor{green!68!red!25!white} 68.36 & Trained Router \\
siglip2-so400m-patch14-384 & best $\eta$ & KNN ($k$=5) & rank regression & 48.16 & 92.17 & \cellcolor{green!71!red!25!white} 71.46 & Trained Router \\
\midrule
--- & --- & strongest per category & --- & 48.28 & 112.34 & \cellcolor{green!44!red!25!white} 43.53 & Fixed Baseline \\
--- & --- & strongest global & --- & 48.28 & 120.76 & \cellcolor{green!0!red!25!white} 0.00 & Fixed Baseline \\
--- & --- & expensive global & --- & 48.28 & 120.76 & \cellcolor{green!0!red!25!white} 0.00 & Fixed Baseline \\
--- & --- & random & --- & 38.93 & 64.49 & \cellcolor{green!45!red!25!white} 45.03 & Fixed Baseline \\
--- & --- & cheapest global & --- & 29.56 & 2.02 & \cellcolor{green!0!red!25!white} 0.00 & Fixed Baseline \\
--- & --- & ood baseline (average) & --- & 48.28 & 120.76 & \cellcolor{green!0!red!25!white} 0.00 & OOD Baseline \\
--- & --- & ood baseline (concat) & --- & 37.32 & 49.37 & \cellcolor{green!38!red!25!white} 38.47 & OOD Baseline \\
--- & --- & ood baseline (normalize\_concat) & --- & 37.32 & 49.37 & \cellcolor{green!38!red!25!white} 38.47 & OOD Baseline \\
--- & --- & ood baseline (vision embedding) & --- & 48.28 & 120.76 & \cellcolor{green!0!red!25!white} 0.00 & OOD Baseline \\
--- & --- & ood baseline (text embedding) & --- & 48.28 & 120.76 & \cellcolor{green!0!red!25!white} 0.00 & OOD Baseline \\
--- & --- & ood baseline (policy embedding) & --- & 48.28 & 120.76 & \cellcolor{green!0!red!25!white} 0.00 & OOD Baseline \\
--- & --- & ood baseline (weighted\_average\_text025) & --- & 48.28 & 120.76 & \cellcolor{green!0!red!25!white} 0.00 & OOD Baseline \\
--- & --- & ood baseline (weighted\_average\_text075) & --- & 48.28 & 120.76 & \cellcolor{green!0!red!25!white} 0.00 & OOD Baseline \\
--- & --- & ood baseline (dinov2-large) & --- & 48.28 & 120.76 & \cellcolor{green!0!red!25!white} 0.00 & OOD Baseline \\
--- & --- & ood baseline (siglip-large-patch16-384) & --- & 36.91 & 42.57 & \cellcolor{green!37!red!25!white} 36.69 & OOD Baseline \\
--- & --- & ood baseline (siglip2-large-patch16-512) & --- & 37.32 & 49.37 & \cellcolor{green!38!red!25!white} 38.47 & OOD Baseline \\
--- & --- & ood baseline (siglip2-so400m-patch14-384) & --- & 36.91 & 46.48 & \cellcolor{green!37!red!25!white} 36.60 & OOD Baseline \\
\midrule
--- & --- & accuracy oracle & --- & 50.47 & 70.83 & \cellcolor{green!89!red!25!white} 88.87 & Oracle \\
--- & --- & $\eta$ oracle & --- & 48.32 & 55.90 & \cellcolor{green!89!red!25!white} 88.76 & Oracle \\
\bottomrule
\end{tabular}}
\end{table}

\subsection{Qwen3-VL-8B-Thinking vs.\ Qwen3-VL-8B-Instruct}
\label{app:ablations:Qwen3_VL_8B_Thinking_Qwen3_VL_8B_Instruct}
\begin{table}[H]
\centering
\caption{Router-architecture ablation (embedding fusion fixed to \texttt{normalize\_concat}, vision encoder SigLIP-2 --- the deployed configuration, App.~\ref{app:configs}). Per architecture we report the best router by $\eta$, with the fixed baselines and oracles. avg\_score / $\eta$ $\times100$; avg\_cost in seconds; $\eta$ cells shaded \colorbox{red!25!white}{red} to \colorbox{green!25!white}{green}.}
\label{tab:arch_Qwen3_VL_8B_Thinking_Qwen3_VL_8B_Instruct}
\footnotesize
\resizebox{\linewidth}{!}{%
\begin{tabular}{@{}llrrr l@{}}
\toprule
\textbf{Architecture} & \textbf{Selection Head} & \textbf{Avg Score} & \textbf{Avg Cost (s)} & \textbf{$\eta$} & \textbf{Category} \\
\midrule
Linear & tie regression & 44.77 & 61.66 & \cellcolor{green!68!red!25!white} 68.36 & Trained Router \\
MLP [2048,1024] & tie regression & 47.48 & 81.72 & \cellcolor{green!75!red!25!white} 74.78 & Trained Router \\
KNN ($k$=5) & rank regression & 47.56 & 90.90 & \cellcolor{green!71!red!25!white} 70.55 & Trained Router \\
PR-KNN ($k$=5) & rank regression & 47.56 & 90.90 & \cellcolor{green!71!red!25!white} 70.55 & Trained Router \\
K-means & tie regression & 47.11 & 94.06 & \cellcolor{green!67!red!25!white} 67.08 & Trained Router \\
OVR & classification & 44.67 & 67.93 & \cellcolor{green!67!red!25!white} 66.94 & Trained Router \\
\midrule
strongest per category & --- & 48.28 & 112.33 & \cellcolor{green!44!red!25!white} 43.62 & Fixed Baseline \\
expensive global & --- & 48.28 & 120.76 & \cellcolor{green!0!red!25!white} 0.00 & Fixed Baseline \\
strongest global & --- & 48.28 & 120.76 & \cellcolor{green!0!red!25!white} 0.00 & Fixed Baseline \\
random & --- & 40.62 & 64.40 & \cellcolor{green!48!red!25!white} 48.48 & Fixed Baseline \\
cheapest global & --- & 31.64 & 1.85 & \cellcolor{green!0!red!25!white} 0.00 & Fixed Baseline \\
ood baseline & --- & 46.89 & 106.22 & \cellcolor{green!54!red!25!white} 54.20 & OOD Baseline \\
\midrule
accuracy oracle & --- & 50.12 & 62.72 & \cellcolor{green!91!red!25!white} 91.30 & Oracle \\
$\eta$ oracle & --- & 48.46 & 50.06 & \cellcolor{green!91!red!25!white} 91.05 & Oracle \\
\bottomrule
\end{tabular}}
\end{table}

\begin{table}[H]
\centering
\caption{Embedding ablation: (i) embedding-fusion method (vision encoder fixed to SigLIP-2) and (ii) vision encoder (fusion fixed to \texttt{normalize\_concat}). Best router per setting by average score and by $\eta$. \emph{text} / \emph{vision} / \emph{policy embedding} are the single-modality fusions (instruction, scene, and $\pi_0$ VLA embeddings). avg\_score / $\eta$ $\times100$; avg\_cost in seconds; $\eta$ cells shaded \colorbox{red!25!white}{red} to \colorbox{green!25!white}{green}.}
\label{tab:embed_Qwen3_VL_8B_Thinking_Qwen3_VL_8B_Instruct}
\footnotesize
\resizebox{\linewidth}{!}{%
\begin{tabular}{@{}llllrrr l@{}}
\toprule
\textbf{Fusion / Encoder} & \textbf{Selected by} & \textbf{Model Architecture} & \textbf{Selection Head} & \textbf{Avg Score} & \textbf{Avg Cost (s)} & \textbf{$\eta$} & \textbf{Category} \\
\midrule
\multicolumn{8}{@{}l}{\emph{(i) Embedding fusion --- vision encoder fixed (SigLIP-2)}} \\
\addlinespace[1pt]
average & best avg score & K-means & regression & 48.28 & 120.76 & \cellcolor{green!0!red!25!white} 0.00 & Trained Router \\
average & best $\eta$ & KNN ($k$=5) & classification & 45.75 & 71.73 & \cellcolor{green!71!red!25!white} 70.88 & Trained Router \\
concat & best avg score & K-means & regression & 48.28 & 120.76 & \cellcolor{green!0!red!25!white} 0.00 & Trained Router \\
concat & best $\eta$ & KNN ($k$=5) & rank regression & 47.56 & 90.90 & \cellcolor{green!71!red!25!white} 70.55 & Trained Router \\
normalize\_concat & best avg score & K-means & regression & 48.28 & 120.76 & \cellcolor{green!0!red!25!white} 0.00 & Trained Router \\
normalize\_concat & best $\eta$ & MLP [2048,1024] & tie regression & 47.48 & 81.72 & \cellcolor{green!75!red!25!white} 74.78 & Trained Router \\
\emph{vision embedding} & best avg score & K-means & regression & 48.28 & 120.76 & \cellcolor{green!0!red!25!white} 0.00 & Trained Router \\
\emph{vision embedding} & best $\eta$ & Linear & rank regression & 45.58 & 81.53 & \cellcolor{green!68!red!25!white} 67.54 & Trained Router \\
\emph{text embedding} & best avg score & K-means & regression & 48.28 & 120.76 & \cellcolor{green!0!red!25!white} 0.00 & Trained Router \\
\emph{text embedding} & best $\eta$ & KNN ($k$=5) & tie regression & 47.40 & 82.75 & \cellcolor{green!74!red!25!white} 74.04 & Trained Router \\
\emph{policy embedding} & best avg score & MLP [2048,1024] & regression & 48.31 & 120.22 & \cellcolor{green!5!red!25!white} 4.75 & Trained Router \\
\emph{policy embedding} & best $\eta$ & KNN ($k$=5) & rank regression & 46.63 & 91.77 & \cellcolor{green!67!red!25!white} 66.95 & Trained Router \\
weighted\_average\_text025 & best avg score & K-means & regression & 48.28 & 120.76 & \cellcolor{green!0!red!25!white} 0.00 & Trained Router \\
weighted\_average\_text025 & best $\eta$ & OVR & classification & 46.00 & 67.85 & \cellcolor{green!73!red!25!white} 72.76 & Trained Router \\
weighted\_average\_text075 & best avg score & K-means & regression & 48.28 & 120.76 & \cellcolor{green!0!red!25!white} 0.00 & Trained Router \\
weighted\_average\_text075 & best $\eta$ & KNN ($k$=5) & classification & 46.45 & 70.54 & \cellcolor{green!74!red!25!white} 74.08 & Trained Router \\
\midrule
\multicolumn{8}{@{}l}{\emph{(ii) Vision encoder --- fusion fixed (\texttt{normalize\_concat})}} \\
\addlinespace[1pt]
dinov2-large & best avg score & MLP [2048,1024] & regression & 48.28 & 119.56 & \cellcolor{green!10!red!25!white} 10.01 & Trained Router \\
dinov2-large & best $\eta$ & KNN ($k$=5) & rank regression & 48.16 & 91.41 & \cellcolor{green!72!red!25!white} 72.19 & Trained Router \\
siglip-large-patch16-384 & best avg score & K-means & regression & 48.28 & 120.76 & \cellcolor{green!0!red!25!white} 0.00 & Trained Router \\
siglip-large-patch16-384 & best $\eta$ & KNN ($k$=5) & tie regression & 47.33 & 84.88 & \cellcolor{green!73!red!25!white} 72.87 & Trained Router \\
siglip2-large-patch16-512 & best avg score & K-means & regression & 48.28 & 120.76 & \cellcolor{green!0!red!25!white} 0.00 & Trained Router \\
siglip2-large-patch16-512 & best $\eta$ & MLP [2048,1024] & tie regression & 47.48 & 81.72 & \cellcolor{green!75!red!25!white} 74.78 & Trained Router \\
siglip2-so400m-patch14-384 & best avg score & K-means & regression & 48.28 & 120.76 & \cellcolor{green!0!red!25!white} 0.00 & Trained Router \\
siglip2-so400m-patch14-384 & best $\eta$ & KNN ($k$=5) & tie regression & 47.12 & 85.98 & \cellcolor{green!72!red!25!white} 71.62 & Trained Router \\
\midrule
--- & --- & strongest per category & --- & 48.28 & 112.33 & \cellcolor{green!44!red!25!white} 43.62 & Fixed Baseline \\
--- & --- & strongest global & --- & 48.28 & 120.76 & \cellcolor{green!0!red!25!white} 0.00 & Fixed Baseline \\
--- & --- & expensive global & --- & 48.28 & 120.76 & \cellcolor{green!0!red!25!white} 0.00 & Fixed Baseline \\
--- & --- & random & --- & 40.62 & 64.40 & \cellcolor{green!48!red!25!white} 48.48 & Fixed Baseline \\
--- & --- & cheapest global & --- & 31.64 & 1.85 & \cellcolor{green!0!red!25!white} 0.00 & Fixed Baseline \\
--- & --- & ood baseline (average) & --- & 48.28 & 120.76 & \cellcolor{green!0!red!25!white} 0.00 & OOD Baseline \\
--- & --- & ood baseline (concat) & --- & 46.89 & 106.22 & \cellcolor{green!54!red!25!white} 54.20 & OOD Baseline \\
--- & --- & ood baseline (normalize\_concat) & --- & 46.89 & 106.22 & \cellcolor{green!54!red!25!white} 54.20 & OOD Baseline \\
--- & --- & ood baseline (vision embedding) & --- & 48.28 & 120.76 & \cellcolor{green!0!red!25!white} 0.00 & OOD Baseline \\
--- & --- & ood baseline (text embedding) & --- & 33.52 & 31.63 & \cellcolor{green!11!red!25!white} 11.06 & OOD Baseline \\
--- & --- & ood baseline (policy embedding) & --- & 48.28 & 120.76 & \cellcolor{green!0!red!25!white} 0.00 & OOD Baseline \\
--- & --- & ood baseline (weighted\_average\_text025) & --- & 48.28 & 120.76 & \cellcolor{green!0!red!25!white} 0.00 & OOD Baseline \\
--- & --- & ood baseline (weighted\_average\_text075) & --- & 48.28 & 120.76 & \cellcolor{green!0!red!25!white} 0.00 & OOD Baseline \\
--- & --- & ood baseline (dinov2-large) & --- & 46.20 & 107.06 & \cellcolor{green!51!red!25!white} 51.46 & OOD Baseline \\
--- & --- & ood baseline (siglip-large-patch16-384) & --- & 46.20 & 102.54 & \cellcolor{green!57!red!25!white} 57.24 & OOD Baseline \\
--- & --- & ood baseline (siglip2-large-patch16-512) & --- & 46.89 & 106.22 & \cellcolor{green!54!red!25!white} 54.20 & OOD Baseline \\
--- & --- & ood baseline (siglip2-so400m-patch14-384) & --- & 46.62 & 103.46 & \cellcolor{green!57!red!25!white} 57.25 & OOD Baseline \\
\midrule
--- & --- & accuracy oracle & --- & 50.12 & 62.72 & \cellcolor{green!91!red!25!white} 91.30 & Oracle \\
--- & --- & $\eta$ oracle & --- & 48.46 & 50.06 & \cellcolor{green!91!red!25!white} 91.05 & Oracle \\
\bottomrule
\end{tabular}}
\end{table}

\subsection{Comparing Router Architectures}
\label{app:classifiers}

Across pairs, the linear router with a soft-label or tie-regression head is most often
selected on $\eta$, while the \texttt{mlp} regression head tends to win on raw average
score at higher cost; hyperparameters for each architecture are in
Appendix~\ref{app:router:hyperparams}.

\section{Additional Results}
\label{app:results}

\subsection{Expanded Main-Paper Result Figures}
\label{app:results:expanded}

\begin{figure}[H]
\centering
\includegraphics[width=0.9\linewidth]{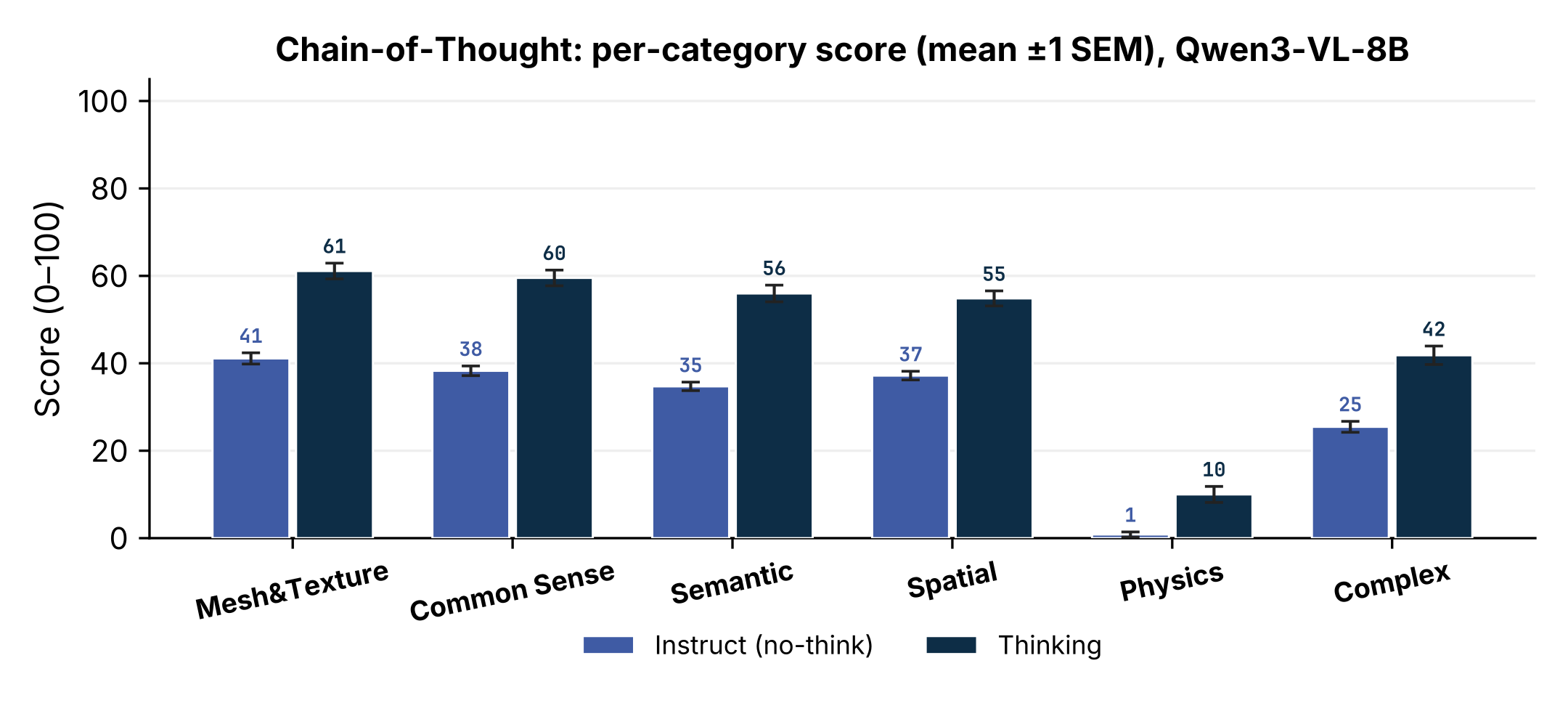}

\vspace{8pt}
\includegraphics[width=0.9\linewidth]{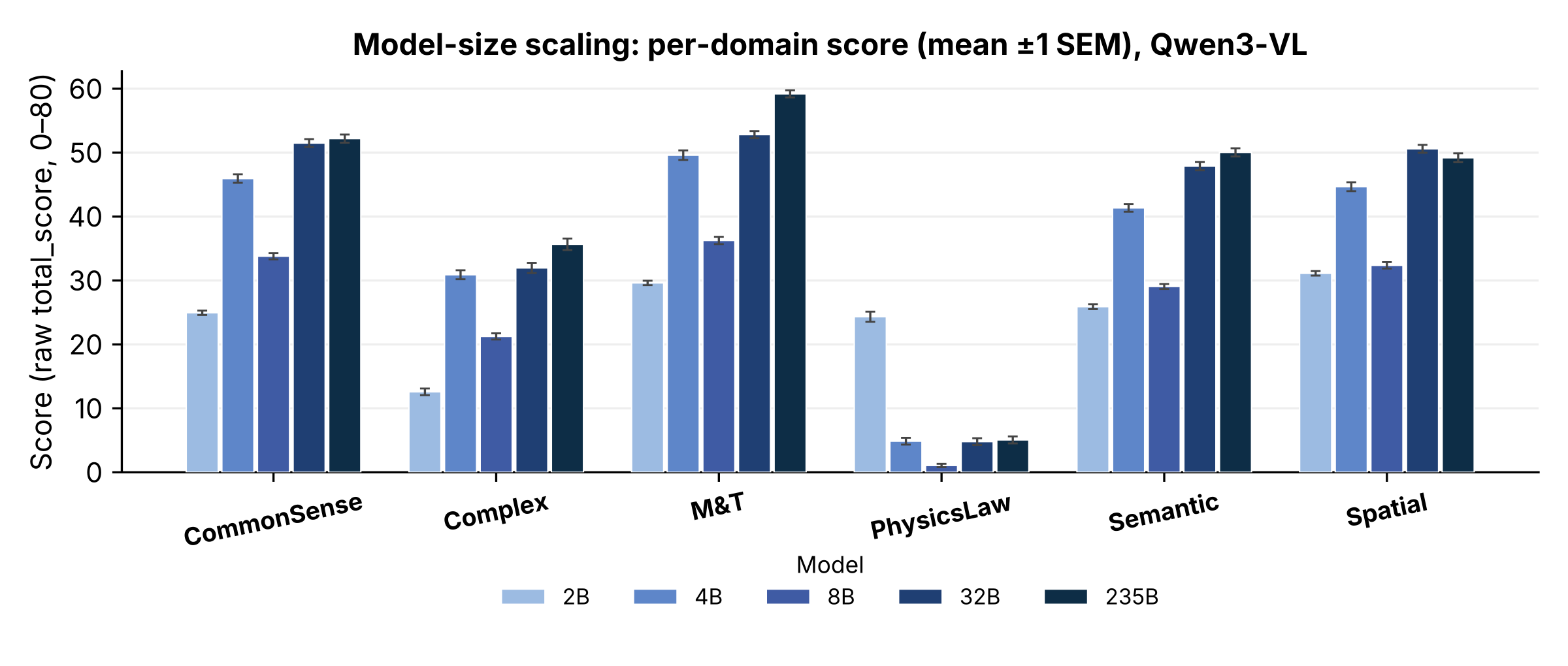}

\vspace{8pt}
\includegraphics[width=0.9\linewidth]{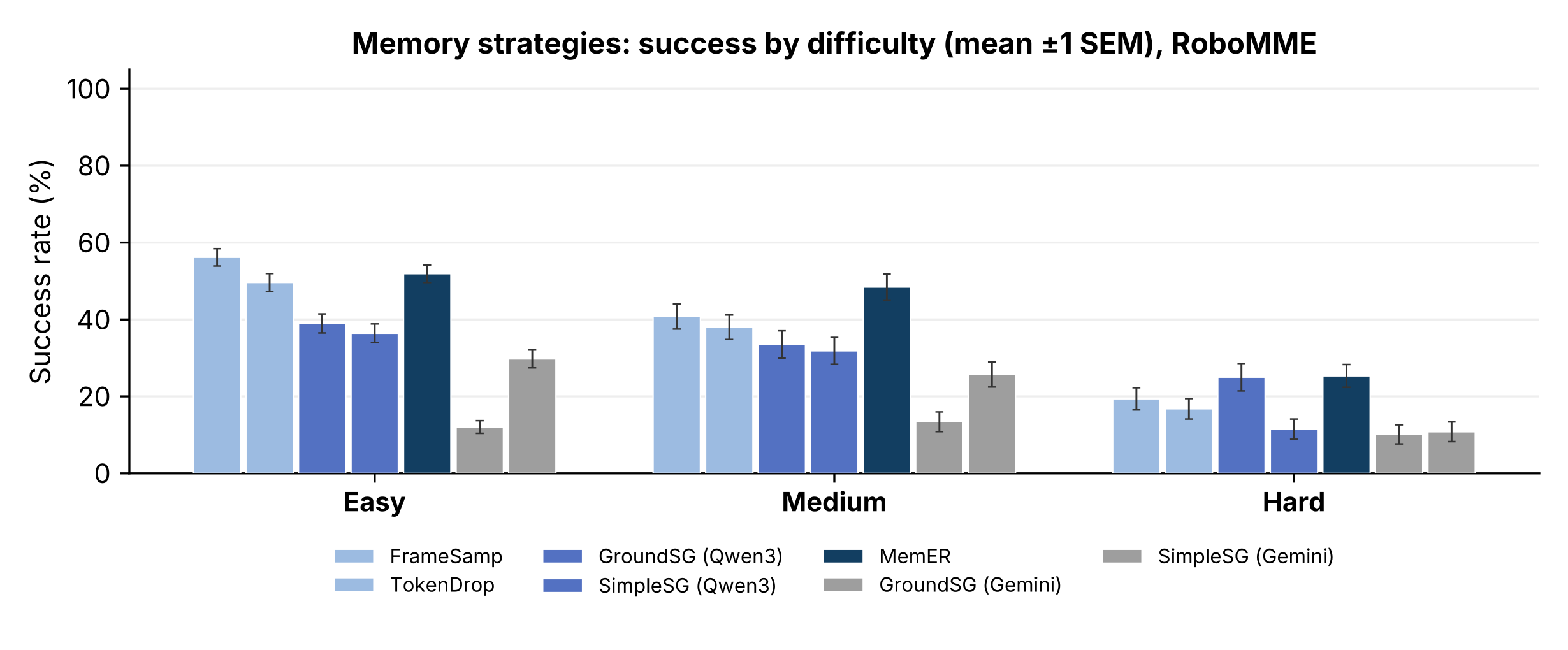}
\caption{\textbf{Main-paper results with $\pm 1$ SEM error bars}, by test-time-compute axis.
\textbf{Top --- intelligence / thinking:} score for No-Think vs.\ Think planners.
\textbf{Middle --- skill command / model size:} score for different-sized planners from the same model family.
\textbf{Bottom --- memory:} success for various memory-augmented VLA. Error bars
denote $\pm 1$ SEM.}
\label{fig:app:sem}
\end{figure}

For completeness, this section further details main-paper result figures
with $\pm 1$ standard-error-of-the-mean (SEM) error bars and other notable details.
Fig.~\ref{fig:app:sem} gives the per-axis results with SEM error bars for the
intelligence/thinking, skill/model-size, and memory axes.

\begin{figure}[H]
\centering
\includegraphics[width=\linewidth]{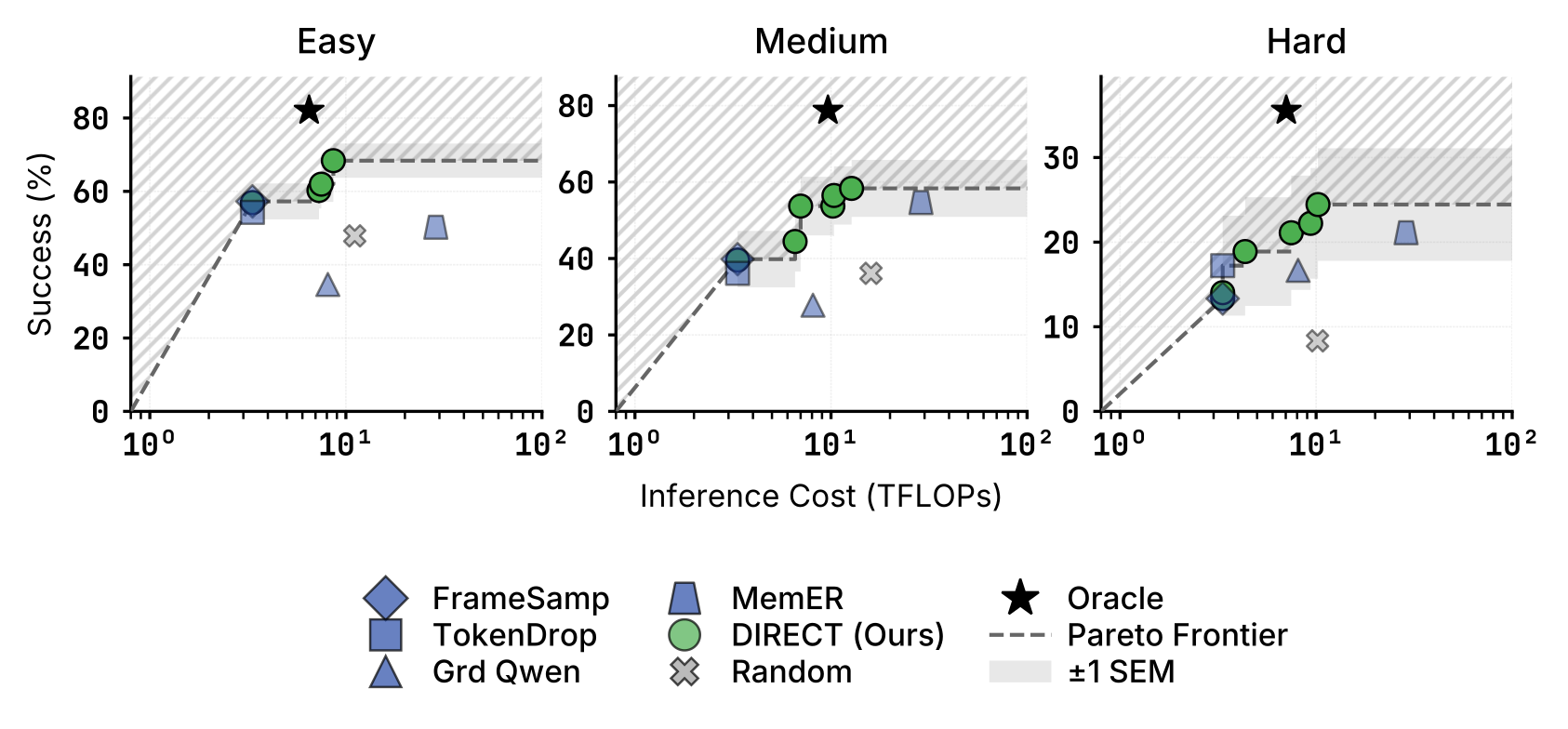}

\vspace{8pt}
\includegraphics[width=\linewidth]{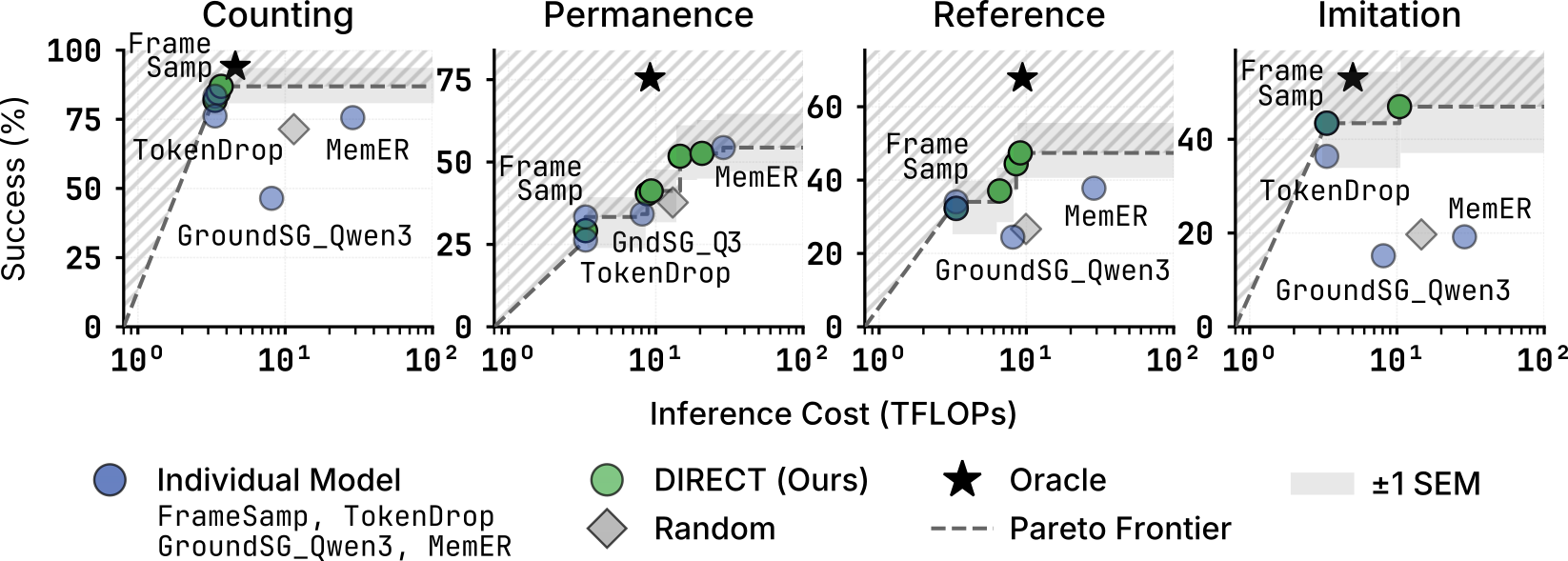}
\caption{\textbf{Enlarged RoboMME memory-routing Pareto frontiers}, success (\%) vs.\
inference cost (TFLOPs, log scale), labeled by memory architecture. Points are the individual
architectures (\texttt{FrameSamp}, \texttt{TokenDrop}, \texttt{GroundSG}-Qwen3,
\texttt{MemER}), \algname's routed frontier (dashed), and the random and oracle references.
\textbf{Top:} split by difficulty tier (easy / medium / hard), with $\pm1$ SEM error bars.
\textbf{Bottom:} split by task suite (counting, permanence, reference, imitation).}
\label{fig:app:mem_pareto}
\end{figure}

Fig.~\ref{fig:app:mem_pareto} gives the enlarged RoboMME memory-routing Pareto frontiers,
with each memory architecture labeled: the top row breaks the routed success--cost frontier
out by difficulty tier and the bottom row by task suite. Across both breakdowns \algname's
routed frontier tracks the oracle and sits at or above the best individual memory
architecture, at a fraction of \texttt{MemER}'s cost.

\subsection{Individual-Planner Simulation Performance}
\label{app:results:permodel}

We report the individual-planner performance underlying the routing pools, before any
routing. Table~\ref{tab:app:vlabench_per_planner} gives the VLABench results broken out by
task category and planner latency; Table~\ref{tab:app:robomme_per_planner} gives the RoboMME
memory-planner results broken out both by difficulty tier and by task suite, against the
$2NT$ inference-cost estimate. All entries are mean success ($\pm$ SEM), and the per-planner
\emph{overall} column is shaded red (low) to green (high) so the capability ordering is
visible at a glance. Latencies for closed-weight (API) planners are full round-trip and
noisier than the locally-measured open-weight ones (Appendix~\ref{app:setup}).

\begin{table}[H]\centering
\caption{\textbf{Per-planner performance on VLABench, by task category.} Cells are mean
success (\%) $\pm$ SEM. Unmarked rows use the full $100$ episodes/task ($n\approx4500$);
\textsuperscript{$\ddagger$} marks the $20$-episode/task ($20\%$) subsample ($n\approx900$) used
for the pairwise routing comparisons in the main paper (a model may appear as both).
\emph{Think} marks chain-of-thought planners. Color runs \colorbox{red!25!white}{red} to
\colorbox{green!25!white}{green}: \emph{Overall} (mean across categories) is higher\,$=$\,green,
while the resource columns \emph{Size} and \emph{Lat.}\ are inverted (smaller / faster\,$=$\,green,
log-scaled); API planners have round-trip latencies (Appendix~\ref{app:setup}) and a grey \emph{Size} cell.}
\label{tab:app:vlabench_per_planner}
\footnotesize
\resizebox{\linewidth}{!}{%
\begin{tabular}{@{}ll l c rrrrrr r r@{}}
\toprule
\textbf{Model} & \textbf{Size} & \textbf{Think} & \textbf{Common} & \textbf{Complex} & \textbf{M\&T} & \textbf{Physics} & \textbf{Semantic} & \textbf{Spatial} & \textbf{Lat.\,(s)} & \textbf{Overall} \\
\midrule
Gemini 3 Flash & \cellcolor{black!10} API & \ding{51} & \pms{77.06}{0.78} & \pms{56.43}{1.21} & \pms{85.78}{0.58} & \pms{1.83}{0.38} & \pms{75.92}{0.77} & \pms{76.25}{0.67} & \cellcolor{green!45!red!25!white} \pms{15.57}{0.22} & \cellcolor{green!66!red!25!white} \pms{66.11}{0.5} \\
Gemini Robotics-ER & \cellcolor{black!10} API & \ding{51} & \pms{74.44}{0.8} & \pms{45.45}{1.2} & \pms{84.08}{0.66} & \pms{3.25}{0.5} & \pms{72.44}{0.8} & \pms{75.16}{0.65} & \cellcolor{green!63!red!25!white} \pms{6.99}{0.03} & \cellcolor{green!63!red!25!white} \pms{63.38}{0.5} \\
Gemini 3 Flash (min-think) & \cellcolor{black!10} API & -- & \pms{73.22}{0.77} & \pms{56.31}{1.22} & \pms{80.92}{0.67} & \pms{0.0}{0.0} & \pms{69.64}{0.79} & \pms{75.81}{0.65} & \cellcolor{green!93!red!25!white} \pms{1.87}{0.01} & \cellcolor{green!63!red!25!white} \pms{62.86}{0.49} \\
Gemini Robotics-ER (think-off) & \cellcolor{black!10} API & -- & \pms{73.03}{0.79} & \pms{46.5}{1.23} & \pms{83.81}{0.63} & \pms{1.25}{0.32} & \pms{69.67}{0.83} & \pms{77.47}{0.63} & \cellcolor{green!88!red!25!white} \pms{2.31}{0.01} & \cellcolor{green!63!red!25!white} \pms{62.78}{0.5} \\
Qwen3-VL-235B-Instruct (FP8) & \cellcolor{green!0!red!25!white} 235B & -- & \pms{65.25}{0.8} & \pms{44.58}{1.13} & \pms{74.0}{0.7} & \pms{6.33}{0.68} & \pms{62.56}{0.81} & \pms{61.5}{0.87} & \cellcolor{green!51!red!25!white} \pms{12.19}{0.08} & \cellcolor{green!56!red!25!white} \pms{55.64}{0.46} \\
GLM-4.1V-Thinking & \cellcolor{green!68!red!25!white} 9B & \ding{51} & \pms{66.28}{0.84} & \pms{37.88}{1.04} & \pms{70.61}{0.82} & \pms{1.08}{0.3} & \pms{61.42}{0.82} & \pms{65.81}{0.82} & \cellcolor{green!22!red!25!white} \pms{44.93}{0.53} & \cellcolor{green!54!red!25!white} \pms{54.24}{0.48} \\
Qwen3-VL-32B-Instruct (FP8) & \cellcolor{green!42!red!25!white} 32B & -- & \pms{64.38}{0.77} & \pms{39.96}{1.0} & \pms{66.03}{0.69} & \pms{6.0}{0.66} & \pms{59.86}{0.8} & \pms{63.25}{0.77} & \cellcolor{green!16!red!25!white} \pms{58.64}{0.88} & \cellcolor{green!53!red!25!white} \pms{53.11}{0.44} \\
Qwen3-VL-8B-Thinking\textsuperscript{$\ddagger$} & \cellcolor{green!71!red!25!white} 8B & \ding{51} & \pms{59.53}{1.8} & \pms{41.83}{2.11} & \pms{61.11}{1.8} & \pms{10.0}{1.83} & \pms{55.97}{1.91} & \pms{54.84}{1.72} & \cellcolor{green!0!red!25!white} \pms{117.7}{2.03} & \cellcolor{green!50!red!25!white} \pms{49.73}{0.94} \\
Qwen3-VL-4B-Instruct (FP8) & \cellcolor{green!85!red!25!white} 4B & -- & \pms{57.44}{0.84} & \pms{38.63}{0.88} & \pms{62.0}{0.94} & \pms{6.08}{0.67} & \pms{51.69}{0.75} & \pms{55.84}{0.87} & \cellcolor{green!73!red!25!white} \pms{4.53}{0.21} & \cellcolor{green!48!red!25!white} \pms{47.98}{0.44} \\
GLM-4.6V-Flash-Thinking & \cellcolor{green!68!red!25!white} 9B & \ding{51} & \pms{55.06}{0.87} & \pms{32.37}{0.98} & \pms{64.25}{0.81} & \pms{0.25}{0.14} & \pms{52.03}{0.81} & \pms{60.12}{0.82} & \cellcolor{green!36!red!25!white} \pms{23.77}{0.2} & \cellcolor{green!47!red!25!white} \pms{47.36}{0.45} \\
RynnBrain-8B-Plan & \cellcolor{green!71!red!25!white} 8B & -- & \pms{56.91}{0.91} & \pms{43.18}{1.02} & \pms{64.81}{0.86} & \pms{0.0}{0.0} & \pms{47.47}{0.9} & \pms{55.34}{0.91} & \cellcolor{green!95!red!25!white} \pms{1.68}{0.01} & \cellcolor{green!47!red!25!white} \pms{47.21}{0.46} \\
Qwen3-VL-8B-Instruct (FP8) & \cellcolor{green!71!red!25!white} 8B & -- & \pms{42.25}{0.61} & \pms{26.57}{0.61} & \pms{45.33}{0.72} & \pms{1.33}{0.33} & \pms{36.33}{0.49} & \pms{40.47}{0.63} & \cellcolor{green!76!red!25!white} \pms{3.89}{0.08} & \cellcolor{green!34!red!25!white} \pms{34.17}{0.32} \\
Qwen3-VL-2B-Instruct & \cellcolor{green!100!red!25!white} 2B & -- & \pms{31.19}{0.43} & \pms{15.73}{0.66} & \pms{37.03}{0.43} & \pms{30.42}{1.0} & \pms{32.39}{0.49} & \pms{38.91}{0.44} & \cellcolor{green!66!red!25!white} \pms{6.21}{0.24} & \cellcolor{green!32!red!25!white} \pms{32.15}{0.25} \\
Qwen3-VL-8B-Instruct\textsuperscript{$\ddagger$} & \cellcolor{green!71!red!25!white} 8B & -- & \pms{38.3}{1.1} & \pms{25.5}{1.3} & \pms{41.1}{1.3} & \pms{0.8}{0.6} & \pms{34.7}{1.0} & \pms{37.2}{1.0} & \cellcolor{green!93!red!25!white} \pms{1.85}{0.02} & \cellcolor{green!32!red!25!white} \pms{31.5}{0.6} \\
Qwen3-VL-8B-Instruct & \cellcolor{green!71!red!25!white} 8B & -- & \pms{38.06}{0.47} & \pms{24.82}{0.55} & \pms{39.69}{0.48} & \pms{0.75}{0.25} & \pms{34.81}{0.43} & \pms{37.59}{0.48} & \cellcolor{green!93!red!25!white} \pms{1.89}{0.02} & \cellcolor{green!31!red!25!white} \pms{31.21}{0.27} \\
Cosmos-Reason2 & \cellcolor{green!71!red!25!white} 8B & -- & \pms{31.62}{0.39} & \pms{27.48}{0.63} & \pms{35.39}{0.41} & \pms{5.25}{0.63} & \pms{32.08}{0.39} & \pms{36.97}{0.44} & \cellcolor{green!91!red!25!white} \pms{2.07}{0.01} & \cellcolor{green!29!red!25!white} \pms{29.44}{0.24} \\
Cosmos-Reason2\textsuperscript{$\ddagger$} & \cellcolor{green!71!red!25!white} 8B & -- & \pms{30.9}{0.9} & \pms{26.0}{1.4} & \pms{34.7}{0.9} & \pms{5.8}{1.5} & \pms{30.8}{0.8} & \pms{38.0}{1.0} & \cellcolor{green!91!red!25!white} \pms{2.05}{0.03} & \cellcolor{green!29!red!25!white} \pms{29.0}{0.5} \\
GLM-4.6V-Flash & \cellcolor{green!68!red!25!white} 9B & -- & \pms{31.38}{0.52} & \pms{17.02}{0.58} & \pms{38.22}{0.58} & \pms{0.0}{0.0} & \pms{28.42}{0.48} & \pms{37.22}{0.56} & \cellcolor{green!89!red!25!white} \pms{2.18}{0.03} & \cellcolor{green!27!red!25!white} \pms{27.41}{0.28} \\
RynnBrain-8B & \cellcolor{green!71!red!25!white} 8B & -- & \pms{26.25}{0.52} & \pms{7.06}{0.54} & \pms{31.28}{0.55} & \pms{0.0}{0.0} & \pms{16.53}{0.6} & \pms{33.47}{0.51} & \cellcolor{green!100!red!25!white} \pms{1.36}{0.02} & \cellcolor{green!21!red!25!white} \pms{20.96}{0.28} \\
\bottomrule
\end{tabular}}
\end{table}

\begin{table}[H]\centering
\caption{\textbf{Per-planner RoboMME success (\%) $\pm$ SEM}, broken out by difficulty tier and by
task suite. Color scale \colorbox{red!25!white}{red} to \colorbox{green!25!white}{green}:
\emph{Overall} is higher\,$=$\,green, while \emph{Cost} (TFLOPs, Appendix~\ref{app:bench:robomme})
is \emph{inverted} (lower\,$=$\,green, log-scaled). The Gemini-predictor variants have unknown
parameter counts, so their FLOPs are not computed (grey, ---). \emph{Overall} is the same under
either breakdown; rows sorted by overall score.}
\label{tab:app:robomme_per_planner}
\footnotesize
\resizebox{\linewidth}{!}{%
\begin{tabular}{@{}l l rrr rrrr r@{}}
\toprule
 & & \multicolumn{3}{c}{\textbf{By difficulty}} & \multicolumn{4}{c}{\textbf{By task suite}} & \\
\cmidrule(lr){3-5}\cmidrule(lr){6-9}
\textbf{Planner} & \textbf{Cost (TFLOPs)} & \textbf{Easy} & \textbf{Medium} & \textbf{Hard} & \textbf{Counting} & \textbf{Permanence} & \textbf{Reference} & \textbf{Imitation} & \textbf{Overall} \\
\midrule
MemER & \cellcolor{green!0!red!25!white} 28.761 & \pms{51.9}{2.28} & \pms{48.42}{3.36} & \pms{25.34}{2.96} & \pms{71.83}{3.25} & \pms{53.47}{3.27} & \pms{37.86}{2.92} & \pms{29.5}{3.0} & \cellcolor{green!46!red!25!white} \pms{45.55}{1.66} \\
FrameSamp & \cellcolor{green!100!red!25!white} 3.338 & \pms{56.15}{2.26} & \pms{40.78}{3.27} & \pms{19.37}{2.88} & \pms{79.06}{3.28} & \pms{25.55}{2.71} & \pms{36.68}{3.01} & \pms{50.83}{3.23} & \cellcolor{green!45!red!25!white} \pms{44.72}{1.68} \\
TokenDrop & \cellcolor{green!100!red!25!white} 3.338 & \pms{49.62}{2.33} & \pms{37.99}{3.19} & \pms{16.78}{2.66} & \pms{73.0}{3.51} & \pms{26.4}{2.79} & \pms{35.18}{3.0} & \pms{38.0}{3.19} & \cellcolor{green!40!red!25!white} \pms{39.93}{1.66} \\
GroundSG (Qwen3) & \cellcolor{green!59!red!25!white} 8.091 & \pms{38.97}{2.47} & \pms{33.52}{3.54} & \pms{25.0}{3.57} & \pms{55.37}{4.54} & \pms{39.09}{3.49} & \pms{31.66}{3.31} & \pms{21.0}{2.89} & \cellcolor{green!35!red!25!white} \pms{34.73}{1.78} \\
SimpleSG (Qwen3) & \cellcolor{green!59!red!25!white} 8.091 & \pms{36.41}{2.44} & \pms{31.84}{3.49} & \pms{11.49}{2.63} & \pms{61.16}{4.45} & \pms{17.77}{2.73} & \pms{24.62}{3.06} & \pms{29.0}{3.22} & \cellcolor{green!30!red!25!white} \pms{30.13}{1.71} \\
SimpleSG (Gemini) & \cellcolor{black!10} API & \pms{29.74}{2.32} & \pms{25.7}{3.25} & \pms{10.81}{2.56} & \pms{56.2}{4.53} & \pms{14.21}{2.47} & \pms{23.12}{3.0} & \pms{18.0}{2.72} & \cellcolor{green!25!red!25!white} \pms{24.83}{1.61} \\
GroundSG (Gemini) & \cellcolor{black!10} API & \pms{12.05}{1.65} & \pms{13.41}{2.55} & \pms{10.14}{2.49} & \pms{17.36}{3.46} & \pms{16.24}{2.63} & \pms{11.56}{2.27} & \pms{5.0}{1.54} & \cellcolor{green!12!red!25!white} \pms{11.99}{1.21} \\
\bottomrule
\end{tabular}}
\end{table}

\subsection{Full Hardware Results}
\label{app:results:hardware}

The full physical Franka/DROID results are given in three parts: per-task
individual-planner success (Fig.~\ref{fig:app:hw_pertask}), the test-set Pareto frontiers
used to select the routers (Fig.~\ref{fig:app:hw_pareto}), and the trained routers run on
the robot (Tables~\ref{tab:app:physical_routers_intelligence}--\ref{tab:app:physical_routers_memory}).

\subsection{Individual-Planner Hardware Performance}
\label{app:results:hardware:pertask}

\begin{figure}[H]
\centering
\includegraphics[width=0.92\linewidth]{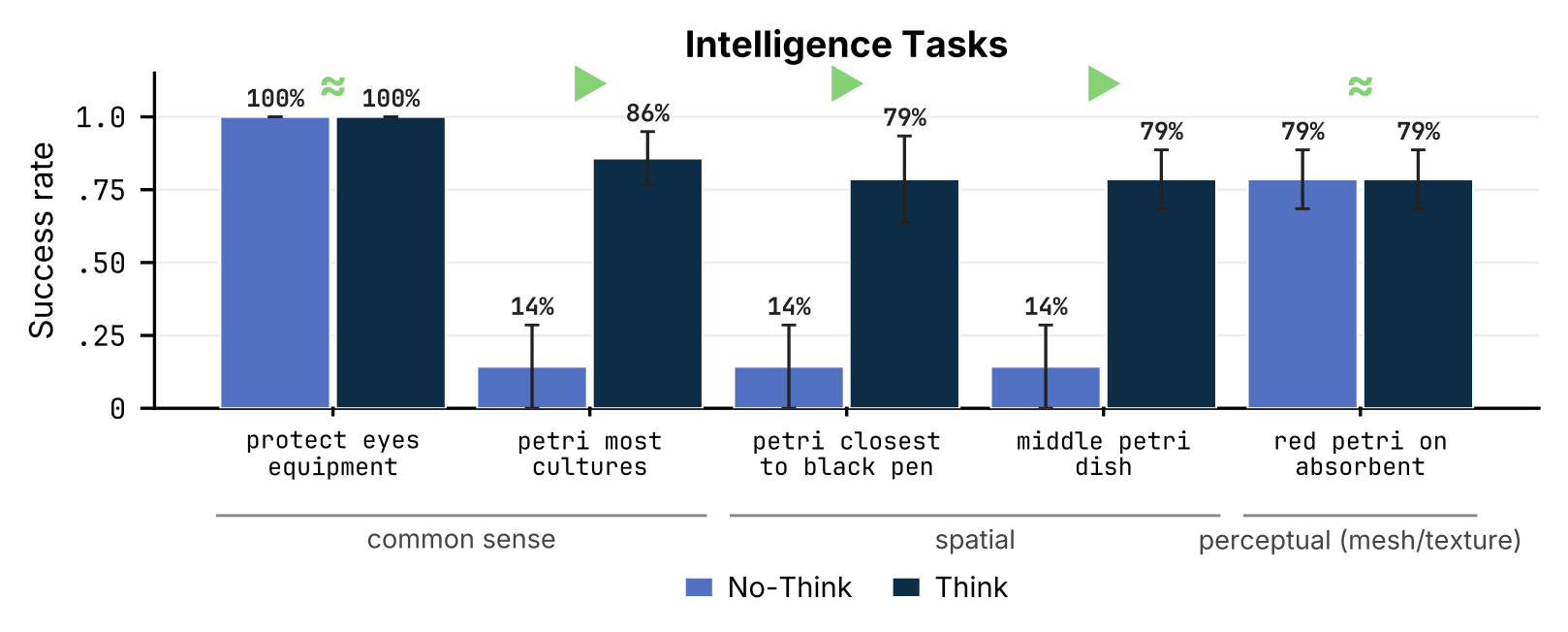}

\vspace{6pt}
\includegraphics[width=0.92\linewidth]{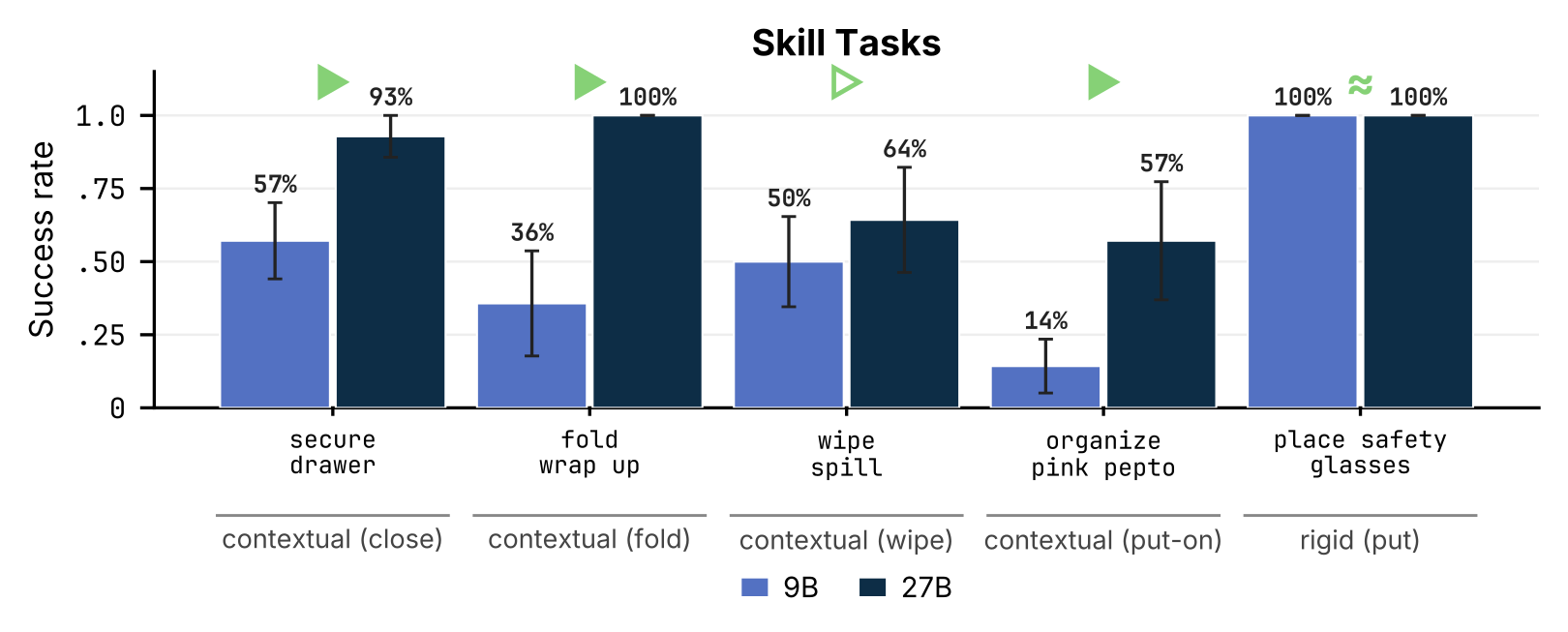}

\vspace{6pt}
\includegraphics[width=0.60\linewidth]{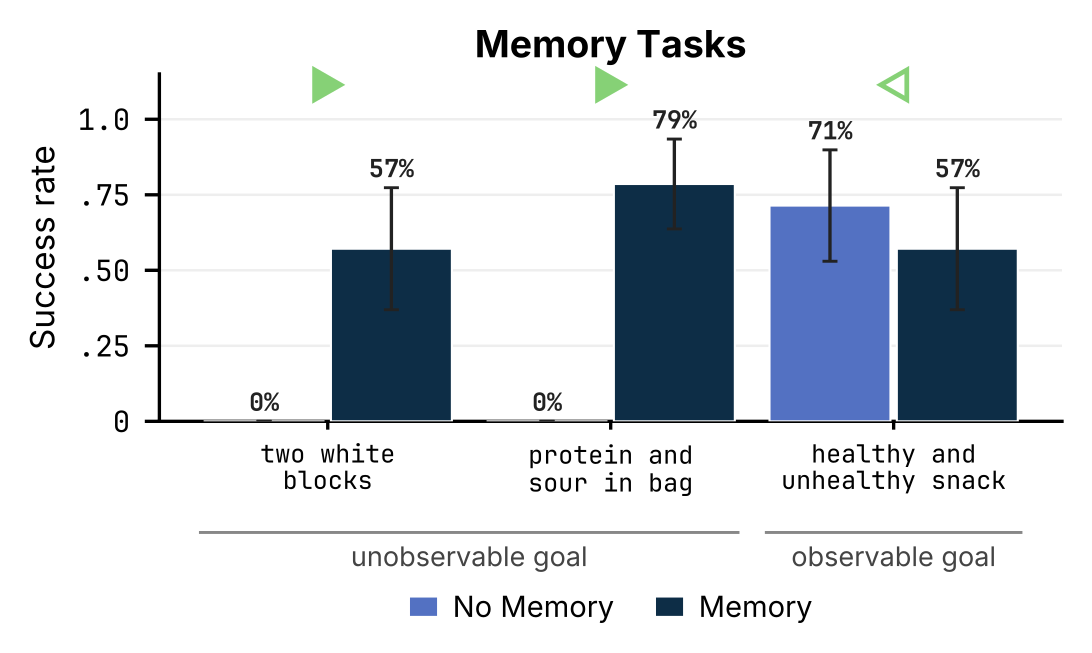}
\caption{\textbf{Per-task success of individual planners on the physical Franka/DROID
suite}, by test-time-compute axis. Each trial is scored $1$ (success), $0$ (failure), or
$0.5$ when the high-level planner emitted a plausible subtask but the low-level policy failed
to carry it out for reasons outside the planner's control (\eg an incorrect grasp); bars are
the mean over trials. Symbols above
each task mark the gap between the two planners in units of the standard error of the
mean (SEM): $\approx$ denotes a gap $\leq 0.5$~SEM (effectively tied); $\triangleleft\,/\,\triangleright$
a gap of $0.5$--$1$~SEM; and $\blacktriangleleft\,/\,\blacktriangleright$ a gap $>1$~SEM,
with the triangle pointing toward the stronger planner.}
\label{fig:app:hw_pertask}
\end{figure}

Fig.~\ref{fig:app:hw_pertask} reports each individual planner's per-task success on the
physical suite (Appendix~\ref{app:bench:droid}), grouped by within-axis category
(Table~\ref{tab:app:droid_tasks}). These planner-level numbers underlie the aggregated
results in the main paper and show which tasks drive each axis's cheap-vs-expensive gap.

\subsubsection{Test-Set Pareto Frontiers}
\label{app:results:hardware:pareto}

\begin{figure}[H]
\centering
\includegraphics[width=\linewidth]{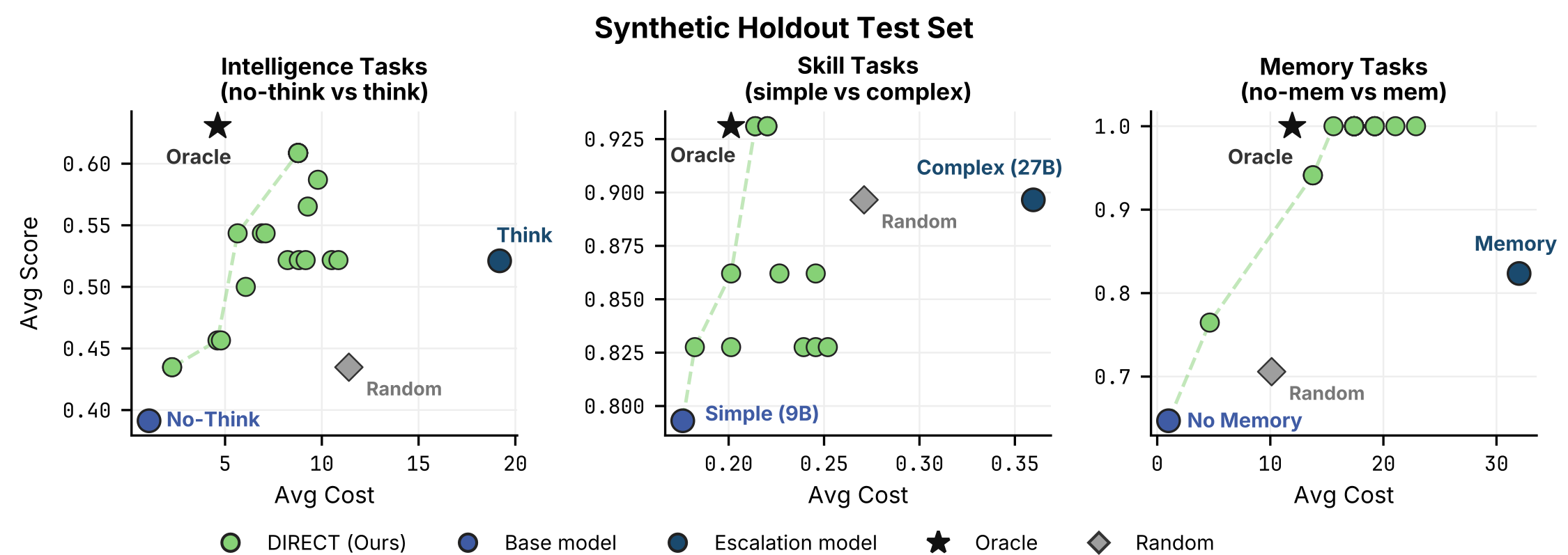}
\caption{\textbf{Held-out test-set Pareto frontiers.} Selected routers, fixed
cheap/expensive planners, and the oracle ceiling are marked.}
\label{fig:app:hw_pareto}
\end{figure}

Since physical rollouts are expensive, routers are selected on the held-out test set of the
synthetic routing data (Appendix~\ref{app:bench:synthetic}) before robot deployment.
Fig.~\ref{fig:app:hw_pareto} shows this test-set cost--quality frontier, marking the
selected routers alongside the fixed cheap/expensive planners and the oracle ceiling.

\subsubsection{Trained Routers on the Physical Suite}
\label{app:results:hardware:routers}

We evaluate the selected routers on the \emph{held-out test split of our synthetic data}
for the physical task suite, reporting each axis separately in the same format as the
simulation ablations (Appendix~\ref{app:ablations}). \textbf{These tables do not report
real-robot results}: the score column is success rate on held-out synthetic episodes, and
there is a single fixed input configuration per axis (Table~\ref{tab:app:exp_configs}).
Results from deploying the routers on the physical hardware are reported separately in
Figures~\ref{fig:...6}, \ref{fig:...8}, and~\ref{fig:...10} of the main text.

\paragraph{Intelligence / thinking axis.} Table~\ref{tab:app:physical_routers_intelligence}
routes between the No-Think and Think planners. The best router by $\eta$ (KNN/PR-KNN,
regression) reaches $60.9\%$ success, exceeding the Think planner ($52.2\%$) and approaching
the oracle ($63.0\%$) while cutting average latency more than half ($8.8$\,s vs.\ $19.2$\,s);
cheaper configurations (\eg K-means classification) trade success for lower cost, tracing the
cost--quality trade-off between the two planners.

\begin{table}[H]\centering
\caption{\textbf{Physical routing results: intelligence / thinking axis} (No-Think vs.\ Think planner). Best trained router per architecture (by routing efficiency $\eta$), with the fixed cheap/expensive planners, the random and OOD baselines, and the success oracle. $Q$ is physical task success rate (\%); $\eta$ is routing efficiency (\%, $\beta=0.1$); cost is planner latency (s). $\eta$ cells are shaded \colorbox{red!25!white}{red} (low) to \colorbox{green!25!white}{green} (high).}
\label{tab:app:physical_routers_intelligence}
\footnotesize
\begin{tabular}{@{}lll rrr@{}}
\toprule
\textbf{Method} & \textbf{Architecture} & \textbf{Head} & \textbf{$Q$ (\%)} & \textbf{Lat (s)} & \textbf{$\eta$ (\%)} \\
\midrule
Trained Router & KNN / PR-KNN ($k$=5) & regression & 60.87 & 8.78 & \cellcolor{green!86!red!25!white} 86.34 \\
Trained Router & Linear & regression & 58.70 & 9.80 & \cellcolor{green!78!red!25!white} 77.73 \\
Trained Router & OVR & rank regression & 54.35 & 6.90 & \cellcolor{green!64!red!25!white} 64.00 \\
Trained Router & MLP [2048,1024] & tie regression & 52.17 & 8.23 & \cellcolor{green!55!red!25!white} 55.04 \\
Trained Router & K-means & classification & 52.17 & 10.86 & \cellcolor{green!54!red!25!white} 53.64 \\
\midrule
Fixed Baseline & Think planner & --- & 52.17 & 19.19 & \cellcolor{green!0!red!25!white} 0.00 \\
Fixed Baseline & random & --- & 43.48 & 11.40 & \cellcolor{green!19!red!25!white} 19.19 \\
Fixed Baseline & No-Think planner & --- & 39.13 & 1.07 & \cellcolor{green!0!red!25!white} 0.00 \\
\midrule
OOD Baseline & OOD baseline & --- & 50.00 & 17.65 & \cellcolor{green!33!red!25!white} 32.57 \\
\midrule
Oracle & accuracy oracle & --- & 63.04 & 4.62 & \cellcolor{green!98!red!25!white} 97.84 \\
\bottomrule
\end{tabular}
\end{table}

\paragraph{Skill command / model-size axis.} Table~\ref{tab:app:physical_routers_skill}
routes between the $9$B and $27$B planners. The top router by $\eta$ (KNN/PR-KNN, tie
regression) reaches $93.1\%$ success---matching the oracle and exceeding the $27$B planner
($89.7\%$)---at near-cheap latency ($0.21$\,s vs.\ the $27$B planner's $0.36$\,s), while the
remaining routers fall between the $9$B and $27$B planners.

\begin{table}[H]\centering
\caption{\textbf{Physical routing results: skill command / model-size axis} ($9$B vs.\ $27$B planner). Best trained router per architecture (by routing efficiency $\eta$), with the fixed cheap/expensive planners, the random and OOD baselines, and the success oracle. $Q$ is physical task success rate (\%); $\eta$ is routing efficiency (\%, $\beta=0.1$); cost is planner latency (s). $\eta$ cells are shaded \colorbox{red!25!white}{red} (low) to \colorbox{green!25!white}{green} (high).}
\label{tab:app:physical_routers_skill}
\footnotesize
\begin{tabular}{@{}lll rrr@{}}
\toprule
\textbf{Method} & \textbf{Architecture} & \textbf{Head} & \textbf{$Q$ (\%)} & \textbf{Lat (s)} & \textbf{$\eta$ (\%)} \\
\midrule
Trained Router & KNN / PR-KNN ($k$=5) & tie regression & 93.10 & 0.21 & \cellcolor{green!98!red!25!white} 97.68 \\
Trained Router & Linear & tie regression & 86.21 & 0.20 & \cellcolor{green!52!red!25!white} 51.98 \\
Trained Router & K-means & classification & 86.21 & 0.23 & \cellcolor{green!51!red!25!white} 51.45 \\
Trained Router & MLP [2048,1024] & tie regression & 82.76 & 0.20 & \cellcolor{green!27!red!25!white} 26.72 \\
Trained Router & OVR & tie regression & 82.76 & 0.24 & \cellcolor{green!26!red!25!white} 26.49 \\
\midrule
Fixed Baseline & $27$B planner & --- & 89.66 & 0.36 & \cellcolor{green!0!red!25!white} 0.00 \\
Fixed Baseline & random & --- & 89.66 & 0.27 & \cellcolor{green!71!red!25!white} 71.41 \\
Fixed Baseline & $9$B planner & --- & 79.31 & 0.18 & \cellcolor{green!0!red!25!white} 0.00 \\
\midrule
OOD Baseline & OOD baseline & --- & 82.76 & 0.18 & \cellcolor{green!27!red!25!white} 26.81 \\
\midrule
Oracle & accuracy oracle & --- & 93.10 & 0.20 & \cellcolor{green!99!red!25!white} 98.57 \\
\bottomrule
\end{tabular}
\end{table}

\paragraph{Memory axis.} Table~\ref{tab:app:physical_routers_memory} routes between the
memory-free and memory planners (here cost is an input-frame proxy). The strongest router
(Linear, soft-label) attains $100\%$ success---matching the oracle and exceeding the
always-memory planner ($82.4\%$)---while processing roughly half its frames ($15.6$ vs.\
$32$); cheaper routers cut frames further at lower success.

\begin{table}[H]\centering
\caption{\textbf{Physical routing results: memory axis} (memory-free vs.\ memory planner). Best trained router per architecture (by routing efficiency $\eta$), with the fixed cheap/expensive planners, the random and OOD baselines, and the success oracle. $Q$ is physical task success rate (\%); $\eta$ is routing efficiency (\%, $\beta=0.1$); cost is an input-frame proxy. $\eta$ cells are shaded \colorbox{red!25!white}{red} (low) to \colorbox{green!25!white}{green} (high).}
\label{tab:app:physical_routers_memory}
\footnotesize
\begin{tabular}{@{}lll rrr@{}}
\toprule
\textbf{Method} & \textbf{Architecture} & \textbf{Head} & \textbf{$Q$ (\%)} & \textbf{Frames} & \textbf{$\eta$ (\%)} \\
\midrule
Trained Router & Linear & soft-label ($\lambda$=0.1) & 100.00 & 15.6 & \cellcolor{green!93!red!25!white} 92.52 \\
Trained Router & KNN / PR-KNN ($k$=5) & classification & 100.00 & 17.4 & \cellcolor{green!91!red!25!white} 90.72 \\
Trained Router & MLP [2048,1024] & tie regression & 100.00 & 17.4 & \cellcolor{green!91!red!25!white} 90.72 \\
Trained Router & OVR & tie regression & 100.00 & 17.4 & \cellcolor{green!91!red!25!white} 90.72 \\
Trained Router & K-means & classification & 100.00 & 19.2 & \cellcolor{green!89!red!25!white} 88.51 \\
\midrule
Fixed Baseline & memory planner & --- & 82.35 & 32.0 & \cellcolor{green!0!red!25!white} 0.00 \\
Fixed Baseline & random & --- & 70.59 & 10.1 & \cellcolor{green!30!red!25!white} 30.21 \\
Fixed Baseline & memory-free planner & --- & 64.71 & 1.0 & \cellcolor{green!15!red!25!white} 15.49 \\
\midrule
OOD Baseline & OOD baseline & --- & 58.82 & 8.3 & \cellcolor{green!0!red!25!white} 0.00 \\
\midrule
Oracle & accuracy oracle & --- & 100.00 & 11.9 & \cellcolor{green!95!red!25!white} 95.28 \\
\bottomrule
\end{tabular}
\end{table}

\subsection{Cross-Axis Routing}
\label{app:cross_axis}

We test whether \algname extends to pools that vary along multiple axes at once. The pool
contains three Qwen3-VL planners: 8B Instruct (FP8) as the cheapest planner, 8B Thinking
(adds chain-of-thought), and 32B Instruct (FP8) (increases model size). The router must
therefore decide not only whether to spend more compute on a task but also which kind.
Routers are trained and evaluated as in the per-axis experiments; baselines and $\eta$ follow
App.~\ref{app:baselines}. The table below shows that KNN achieves the
highest $\eta$ ($59.08$), outperforming the best fixed planner,
and recovers $96\%$ of the Thinking planner's score at $41\%$ of its cost.

\begin{table}[h]
\centering
\small
\caption{\textbf{Cross-axis routing} on VLABench. The pool varies chain-of-thought and
model size simultaneously. Best value per column in bold.}
\label{tab:app:cross_axis}
\begin{tabular}{lrrr}
\toprule
\textbf{Model / Router} & \textbf{Avg Score} $\uparrow$ & \textbf{Avg Cost (s)} $\downarrow$ & $\eta$ $\uparrow$ \\
\midrule
KNN ($k=5$), rank regression   & 38.29 & 150.78 & \cellcolor{green!35}\textbf{59.08} \\
Linear, rank regression        & 38.08 & 204.08 & \cellcolor{green!26}56.70 \\
Random                         & 34.07 & 144.49 & \cellcolor{yellow!22}43.11 \\
\midrule
Qwen3-VL-8B Thinking           & \textbf{39.81} & 369.16 & \cellcolor{red!18}0.00 \\
Qwen3-VL-8B Instruct (FP8)     & 23.99 & \textbf{4.46} & \cellcolor{red!17}0.87 \\
Qwen3-VL-32B Instruct (FP8)    & 37.00 & 85.71 & \cellcolor{green!23}55.30 \\
\bottomrule
\end{tabular}
\end{table}
\section{Experimental and Hardware Setup}
\label{app:setup}

\subsection{Evaluation Hardware}
\label{app:setup:hardware}

Open-weight planners were evaluated locally, on hardware that differs by experiment due to
model-memory requirements:
\begin{itemize}
    \setlength{\itemsep}{0.15em}
    \item \emph{Chain-of-thought} (think vs.\ no-think) experiments: a single
    NVIDIA RTX~4090 GPU.
    \item \emph{Model-size scaling} experiments (Qwen3-VL Instruct 2B--235B): NVIDIA H100
    GPU(s).
\end{itemize}
Because absolute latency is hardware-dependent, latencies are comparable within an
experiment but not directly across the RTX~4090 and H100 settings. RoboMME results are
taken from the public leaderboard (Appendix~\ref{app:bench:robomme}) rather than re-run
locally, and physical DROID planner latency is measured from cloud API endpoints
(Appendix~\ref{app:bench:droid}).

Latency for each planner is reported as \texttt{latency\_s}, the wall-clock time of
the \texttt{evaluate()} call, which includes both per-model preprocessing and
generation time (see Appendix~\ref{app:bench:vlabench}).

\paragraph{Closed-weight / API model latency.} Inference latency measurements for
closed-weight models are derived via cloud API endpoints. These values are subject to
significant variance compared to local open-weight deployment due to request
batching, dynamic server-side load balancing, and network-induced overhead, and the
reported figures therefore include the full round-trip HTTP request.\footnote{Denoted
$^\dagger$ in the result tables.}

\subsection{Inference Hyperparameters for the VLM Planners}
\label{app:setup:inference}

Decoding hyperparameters for each planner's \texttt{model.generate()} call were set
according to the corresponding model card recommendations when available.

\paragraph{Embedding / sequence concatenation.} Throughout, the fusion strategy we
report in the main paper (\texttt{concat} or \texttt{normalize\_concat}) refers to \emph{sequence}
concatenation of the per-modality embeddings (concatenating along the token/sequence
axis), as opposed to \emph{channel} concatenation (concatenating along the feature
axis). See Appendix~\ref{app:router:embeddings} for the full set of fusion strategies
we ablated.

\section{Baselines, Ceilings, and the AUC Metric}
\label{app:baselines}

We use several existing VL-RouterBench baselines (random, strongest-global,
cheapest-global, standard oracle) and contribute our own baseline / ceiling sweeps:
the out-of-distribution (OOD) baseline and a budgeted oracle (modified
from the original to trace a spread of cost--accuracy tradeoffs).

\subsection{Baselines and Ceilings}
\label{app:baselines:list}

\begin{description}
    \setlength{\itemsep}{0.3em}
    \item[Random.] Route each sample to a planner chosen uniformly at random.

    \item[OOD (calibration).] We detect out-of-distribution (OOD) embeddings via conformal prediction \cite{sinha2024realtime}. Given a success threshold $\gamma$: for each planner,
    all training-set samples with score $> \gamma$ form its calibration set. The
    per-planner percentile for pairwise calibration is $T = 5.0$. At inference, a
    test sample is routed to the first (cheapest) planner whose calibration set
    contains a neighbor with cosine similarity at least the $T$-percentile threshold; if no planner achieves this, we route to the most expensive planner. We sweep $\gamma$ from
    $0$ to $1$ in increments of $0.1$. Each $\gamma$ represents a different
    cost--accuracy tradeoff: a higher $\gamma$ shrinks the calibration set, making the
    router more likely to escalate to a more expensive planner in pursuit of a higher
    score.

    \item[Strongest per category.] Route to the strongest planner for each task category.

    \item[Strongest global.] Always route to the planner with the best average score.

    \item[Cheapest global.] Always route to the cheapest average planner.

    \item[Budgeted oracle.] Given a cost budget of $x\%$ of the dataset-wide maximum cost $c_{\max}$ (a single scalar representing the largest cost entry across all samples and planners), pick the
    most accurate planner within budget on the current sample, breaking exact ties by lower cost;  if no planner fits, fall back to the cheapest planner for that sample. We sweep
    $x \in \{0, 5, \dots, 95, 100\}$. As an oracle over the true scores, this traces an approximate theoretical upper bound on the routing cost--accuracy Pareto frontier.

    \item[Accuracy oracle.] Picks the most accurate planner, breaking exact ties by lower cost.

    \item[Rank-score oracle.] Picks the planner with the highest rank-score based on true score and cost, used as the representation of the cost-quality tradeoff on a sample level. Uses the per-sample rank-score equation taken from VL-RouterBench \cite{huang2025vlrouterbenchbenchmarkvisionlanguagemodel} (see section \ref{app:router:preference}), replacing predicted per-sample cost ($\hat{c}_{ik}$) and predicted per-sample score ($\hat{q}_{ik}$) with the true per-sample cost $c_{ik}$ and score $q_{ik}$.The minimum and maximum cost across planners and samples are taken over the test set for log normalization of cost.

\end{description}

\subsection{AUC Metric}
\label{app:baselines:auc}

We summarize each router / baseline / oracle by points of (scaled average quality $q$, scaled average cost $c$), which together form Pareto frontiers. To obtain a single AUC per method, we anchor each frontier with the individual-model points (cheapest / strongest) and compute the area under a piecewise-linear curve over $[\text{min\_cost}, \text{max\_cost}]$, or $[0, 1]$ in the scaled space. If a curve does not span the full domain, we extrapolate flat to the left (held at $y = 0$) and right (held at the last $y$-value).

We compute five specific curves, anchoring each Pareto frontier with the individual cheapest and strongest models:
\begin{enumerate}
    \setlength{\itemsep}{0.15em}
    \item \textbf{Router Pareto} --- Pareto frontier of all trained routers, plus the anchors.
    \item \textbf{OOD sweep} --- Pareto frontier of all OOD threshold points, plus the anchors.
    \item \textbf{Random baseline} --- Pareto frontier of the random baseline routing point, plus the anchors.
    \item \textbf{Strongest-per-category baseline} --- Pareto frontier of the strongest-per-category point, plus the anchors.
    \item \textbf{Oracle ceiling} --- Formed from the points of the budgeted oracle (described in \ref{app:baselines:list}). The budget-oracle curve is not a Pareto frontier: it is simply the budget-sweep
points connected in order of budget percentage, with no model anchors.
\end{enumerate}

\begin{table*}[t]
\centering 
\caption{\footnotesize \textbf{VLABench routing AUC metric results} ($N = 900$). Pair labels: I/T = Instruct/Thinking. $^\dagger$These models were evaluated on a 20\% subsample due to substantially higher inference time.}
\vspace{-5pt}
\label{tab:auc_metric}
\setlength{\tabcolsep}{4pt}
\renewcommand{\arraystretch}{1.15}
\setlength{\dashlinedash}{0.5pt}
\setlength{\dashlinegap}{1.5pt}
\resizebox{\textwidth}{!}{
\begin{tabular}{|l||>{\columncolor{lime!10}}c||c:c:c||>{\columncolor{black!6}\color{black!60}}c|}
\hline
& \cellcolor{lime!10}Ours & \multicolumn{3}{c||}{\textit{Baselines}} & \cellcolor{black!6}\textcolor{black!60}{\textit{Ceiling}} \\
\textbf{Pair (Cheap / Expensive)} & \cellcolor{lime!10}\textbf{\algname} & \textit{OOD} & \textit{Str./cat.} & \textit{Random} & \cellcolor{black!6}\textcolor{black!60}{\textit{Oracle}} \\ \hline
\rowcolor{black!15} \multicolumn{6}{|l|}{\textit{Homogeneous Open-Weight Pairs}} \\
GLM-4.6 I / 4.1 T                   & \textbf{0.5634} & 0.4207 & 0.4795 & 0.4822 & 0.7699 \\
GLM-4.6 I / T                       & \textbf{0.5547} & 0.4095 & 0.4660 & 0.4457 & 0.8177 \\
Qwen3-VL 8B I / T $^\dagger$        & \textbf{0.5957} & 0.4427 & 0.4822 & 0.4564 & 0.8192 \\
\rowcolor{black!15} \multicolumn{6}{|l|}{\textit{Heterogeneous Model Pairs}} \\
Cosmos-R2 / Qwen3-VL 8B T $^\dagger$ & \textbf{0.5732} & 0.4575 & 0.4794 & 0.4362 & 0.7997 \\
RynnBrain-Plan / GLM 4.1 T          & \textbf{0.4632} & 0.3279 & 0.3605 & 0.3295 & 0.8580 \\
\hline
\end{tabular}
}
\vspace{-15pt}
\end{table*}
\section{Benchmark Details}
\label{app:bench}

\subsection{VLABench}
\label{app:bench:vlabench}

VLABench~\citep{Zhang_2025_ICCV} is a language-conditioned manipulation benchmark. We use
its \emph{non-interactive VLM evaluation track} (the \texttt{VLABench/vlm\_evaluation\_v1.0}
dataset), which is designed specifically for evaluating VLMs as high-level planners. In this track the VLM is given the task instruction together with the segmented
scene images and must generate an \emph{action sequence} in the format of our skill
sequence. The generated sequence is then scored by matching its directed acyclic graph (DAG)
against the ground-truth skill DAG, yielding an overall score. This non-interactive,
skill-sequence formulation is exactly the high-level decomposition our router targets, which
is why we adopt it.

\paragraph{Score normalization.} The VLABench scoring utility
(\texttt{VLABench/evaluation/utils.py}, \texttt{get\_final\_score}) computes a weighted
average but includes only two of the sub-scores and weights each of them by $0.4$, so
the maximum attainable score is $80$ rather than $100$. For ease of comprehension and
consistency, we re-normalize all VLABench scores from the $[0, 80]$ range to
$[0, 100]$ (equivalently $[0, 1]$).

\paragraph{Additional metrics tracked.} We extended the evaluator to record, per model:
\begin{itemize}
    \setlength{\itemsep}{0.15em}
    \item \texttt{latency\_s} --- wall-clock seconds for the \texttt{evaluate()} call
    (preprocessing + generation). For API models this includes the full round-trip
    HTTP request and is therefore noisier.
    \item \texttt{power\_mean\_w} / \texttt{power\_peak\_w} --- mean and peak GPU power
    draw.
    \item \texttt{flops\_estimate} --- estimated as
    $2 \cdot \text{num\_params} \cdot (\text{input\_tokens} + \text{output\_tokens})$.
    \item \texttt{thinking\_tokens} --- best-effort count of thinking tokens in the
    output, accounting for per-model differences in thinking-token representation.
\end{itemize}
In this paper, we focus on VLABench's \texttt{latency\_s}, given its strong correlation with the other recorded metrics.

\paragraph{Other modifications.} We made the evaluator robust to per-model output
differences (generalizing the skill-sequence JSON parsing).

\subsection{Physical Setup: Franka / DROID}
\label{app:bench:droid}

\paragraph{Hardware and low-level policy.} We evaluate on a physical Franka arm in the
DROID setup. In the physical experiments the high-level planner is queried through
OpenRouter API calls, while the low-level $\pi_{0.5}$ VLA policy~\citep{black2025pi05}
(finetuned on DROID) executes the emitted subskills locally on
an RTX 5090.
Observations are captured from a wrist camera and a left third-person camera.

\paragraph{Planners.} The high-level planners vary per experiment: Qwen3.5-VL-9B with and
without CoT for the reasoning axis, Qwen3.5-VL-9B vs.\ 27B for the size axis, and
Gemini-Robotics-ER~1.6~\citep{deepmind2025geminiroboticser16} for the memory axis (a
single-frame memory-free planner vs.\ a memory planner given a 40-frame video history plus
a task-list summary).

\paragraph{Latency measurement.} Planner latency is measured from the OpenRouter
standard inference endpoint and is subject to small request-to-request variance.

\paragraph{Task suite.} Table~\ref{tab:app:droid_tasks} lists the physical task suite,
grouped by the test-time compute axis each task targets, and
Figure~\ref{fig:eval-tasks} shows the starting frame for every task. The grouping follows the
domain-bucket$\rightarrow$axis correspondence of the synthetic-data pipeline
(Appendix~\ref{app:bench:synthetic}).

\begingroup
\footnotesize
\setlength{\LTpre}{4pt}\setlength{\LTpost}{4pt}
\begin{longtable}{@{}>{\raggedright\arraybackslash}p{0.17\linewidth}l>{\raggedright\arraybackslash}p{0.35\linewidth}@{}}
\caption{Physical Franka/DROID task suite (instructions given to the planner), grouped by
test-time compute axis and by
within-axis category. For skill command, \emph{rigid} verbs (\eg \emph{put}) map directly
to an action, whereas \emph{contextual} verbs (\eg \emph{close}, \emph{fold}, \emph{wipe},
\emph{put-on}) require grounding the verb in the scene.}
\label{tab:app:droid_tasks}\\
\toprule
\textbf{Category} & \textbf{Task ID} & \textbf{Instruction} \\
\midrule
\endfirsthead
\toprule
\textbf{Category} & \textbf{Task ID} & \textbf{Instruction} \\
\midrule
\endhead
\midrule
\multicolumn{3}{r}{\emph{continued on next page}} \\
\endfoot
\bottomrule
\endlastfoot
\multicolumn{3}{@{}l}{\emph{(1) Intelligence form / thinking (CoT reasoning) --- science-themed tasks}} \\
\addlinespace[1pt]
common sense & \texttt{protect\_eyes\_equipment} & I need to protect my eyes during my next experiment. Place the corresponding equipment in the white box. \\
common sense & \texttt{petri\_most\_cultures} & Place the petri dish with the most cultures in the white box. \\
spatial & \texttt{petri\_closest\_to\_black\_pen} & Place the petri dish that is closest to the black pen into the white box. \\
spatial & \texttt{middle\_petri\_dish} & Place the middle petri dish into the white box. \\
perceptual (mesh/texture) & \texttt{red\_petri\_on\_absorbent} & Place the red petri dish on the absorbent item. \\
\midrule
\multicolumn{3}{@{}l}{\emph{(2) Skill command / model size --- household-themed tasks}} \\
\addlinespace[1pt]
contextual (close) & \texttt{secure\_drawer} & I'm done packing my drawer. Please secure the drawer. \\
contextual (fold) & \texttt{fold\_wrap\_up} & I am packing up my first aid supplies, wrap up. \\
contextual (wipe) & \texttt{wipe\_spill} & Clear away the spill with the towel. \\
contextual (put-on) & \texttt{organize\_pink\_pepto} & Organize the pink Pepto with the other drugs. \\
rigid (put) & \texttt{place\_safety\_glasses\_white\_box} & The safety glasses go in the white box. \\
\midrule
\multicolumn{3}{@{}l}{\emph{(3) Memory --- factory-themed tasks}} \\
\addlinespace[1pt]
unobservable goal & \texttt{two\_white\_blocks} & Place exactly two white blocks in the brown box. \\
observable goal & \texttt{healthy\_and\_unhealthy\_snack} & I want exactly one healthy snack and exactly one unhealthy snack on the plate. \\
unobservable goal & \texttt{protein\_and\_sour\_in\_bag} & Place exactly one high-protein snack and exactly one sour fruit in the bag. \\
\end{longtable}
\endgroup

\newcommand{\evaltaskid}[1]{\texttt{\renewcommand{\_}{\textunderscore\discretionary{}{}{}}#1}}
\newcommand{\evaltaskcard}[3]{%
  \begin{subfigure}[t]{0.19\linewidth}
    \centering
    \includegraphics[width=\linewidth]{figures/eval_tasks/#1}
    \caption{\evaltaskid{#2}: \emph{#3}}
  \end{subfigure}%
}

\begin{figure}[H]
  \centering
  \captionsetup[subfigure]{font=scriptsize,justification=raggedright,singlelinecheck=false}

  {\bfseries Intelligence}\par\smallskip
  \evaltaskcard{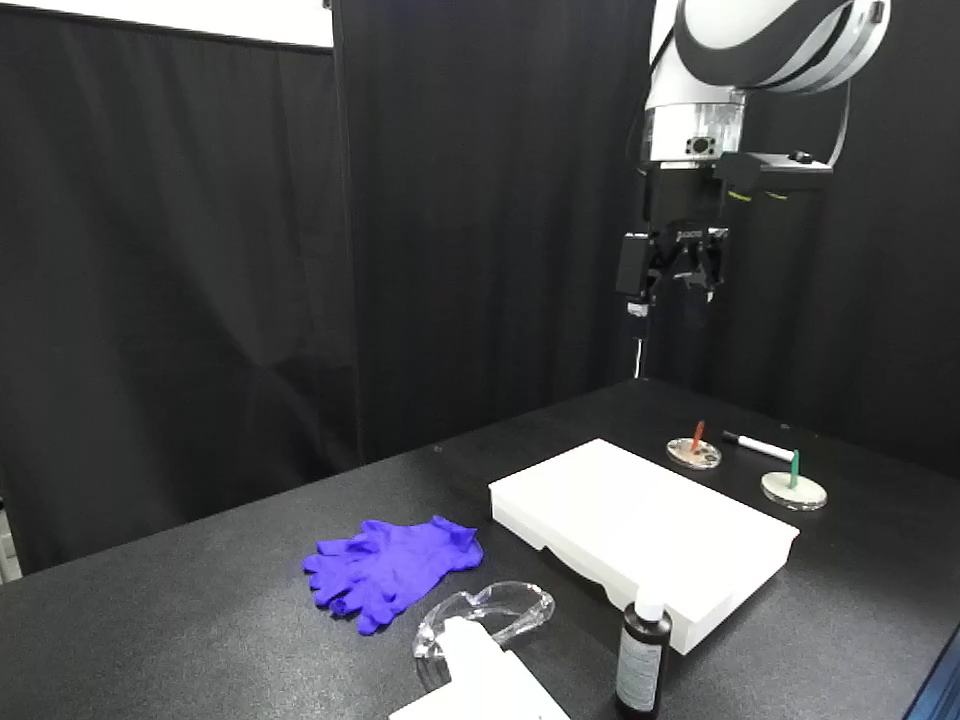}{protect\_eyes\_equipment}{I need to protect my eyes during my next experiment. Place the corresponding equipment in the white box.}\label{fig:eval:protect_eyes_equipment}\hfill
  \evaltaskcard{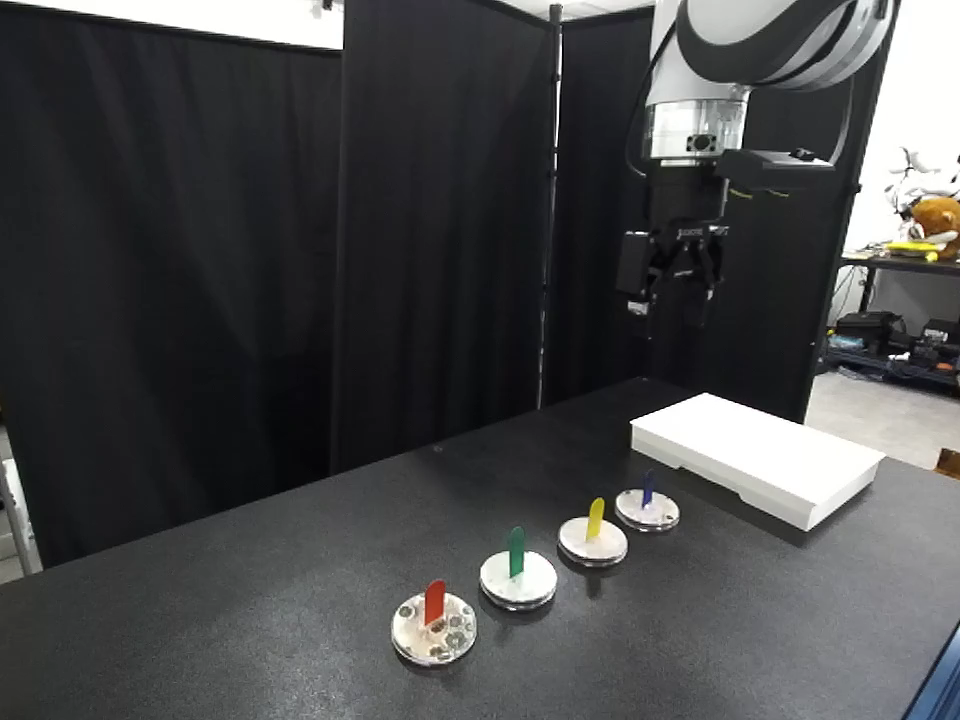}{petri\_most\_cultures}{Place the petri dish with the most cultures in the white box.}\label{fig:eval:petri_most_cultures}\hfill
  \evaltaskcard{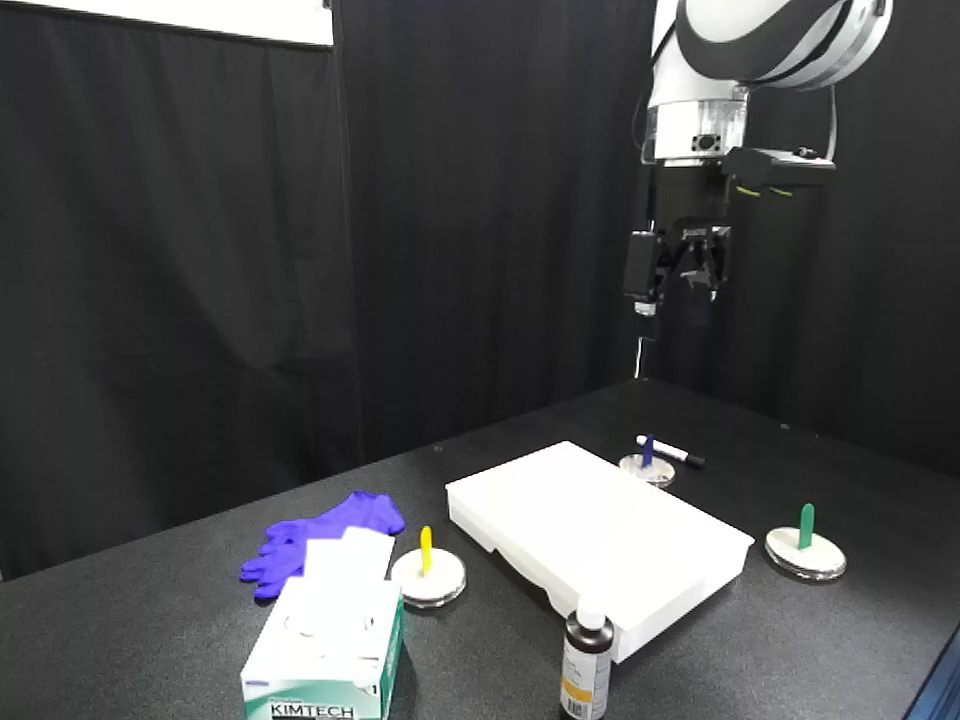}{petri\_closest\_to\_black\_pen}{Place the petri dish that is closest to the black pen into the white box.}\label{fig:eval:petri_closest_to_black_pen}\hfill
  \evaltaskcard{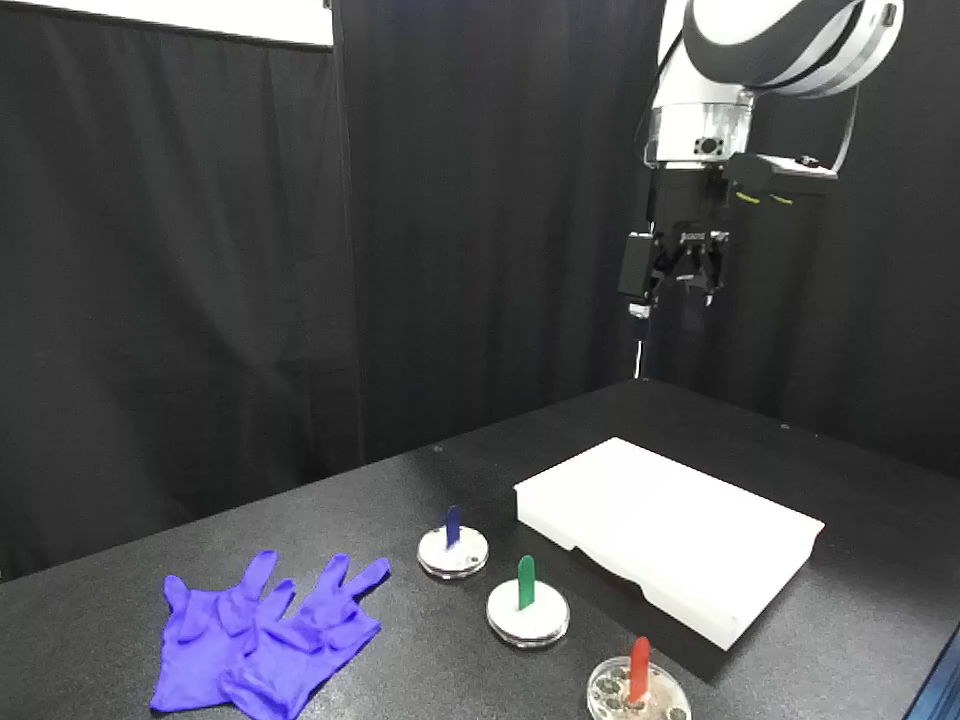}{middle\_petri\_dish}{Place the middle petri dish into the white box.}\label{fig:eval:middle_petri_dish}\hfill
  \evaltaskcard{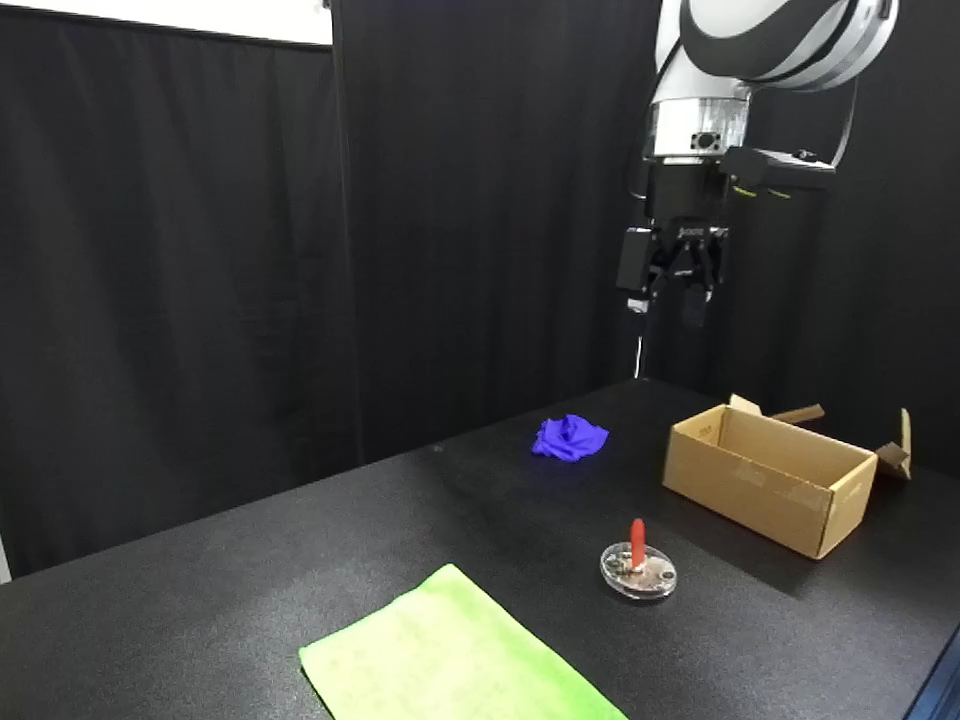}{red\_petri\_on\_absorbent}{Place the red petri dish on the absorbent item.}\label{fig:eval:red_petri_on_absorbent}

  \vspace{1.0em}

  {\bfseries Skill}\par\smallskip
  \evaltaskcard{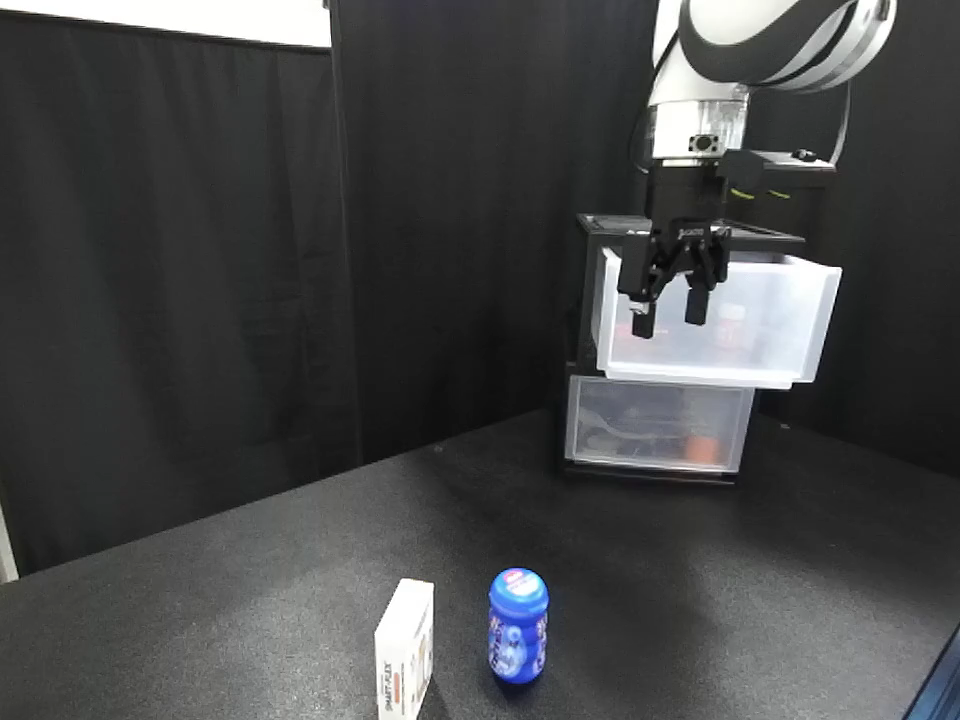}{secure\_drawer}{I'm done packing my drawer. Please secure the drawer.}\label{fig:eval:secure_drawer}\hfill
  \evaltaskcard{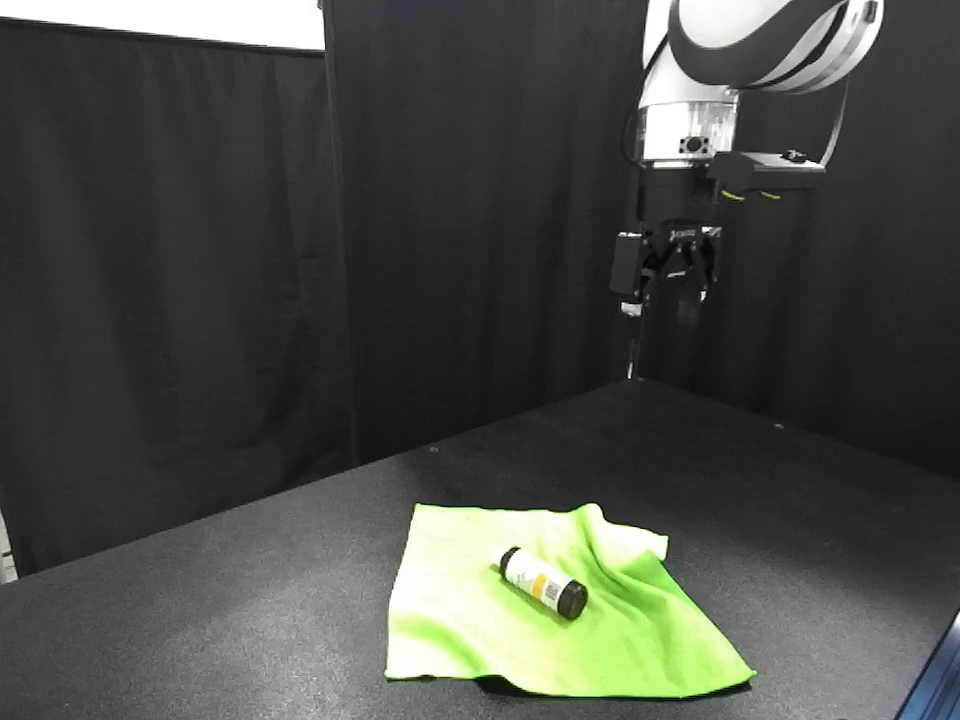}{fold\_compact\_towel}{I am packing up my first aid supplies, wrap up.}\label{fig:eval:fold_compact_towel}\hfill
  \evaltaskcard{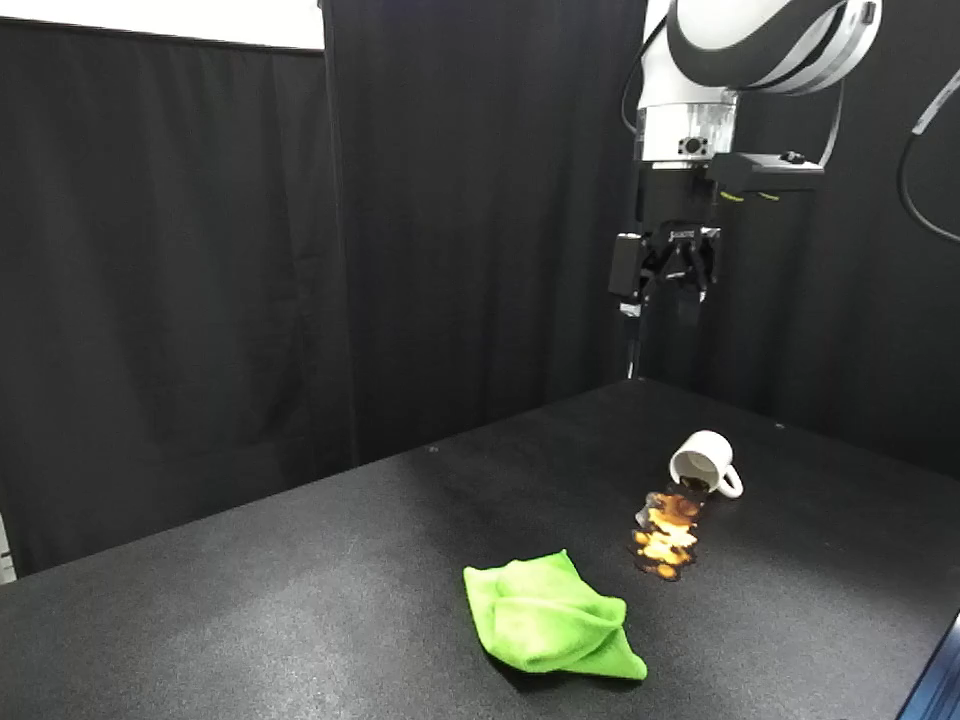}{wipe\_spill}{Clear away the spill with the towel.}\label{fig:eval:wipe_spill}\hfill
  \evaltaskcard{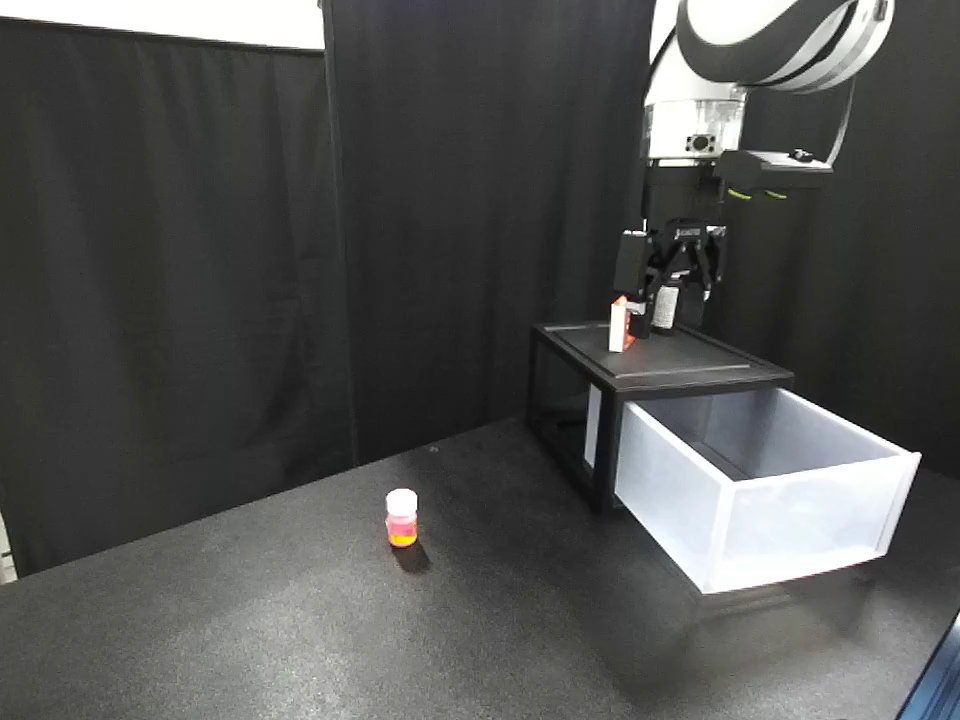}{organize\_pink\_pepto}{Organize the pink Pepto with the other drugs.}\label{fig:eval:organize_pink_pepto}\hfill
  \evaltaskcard{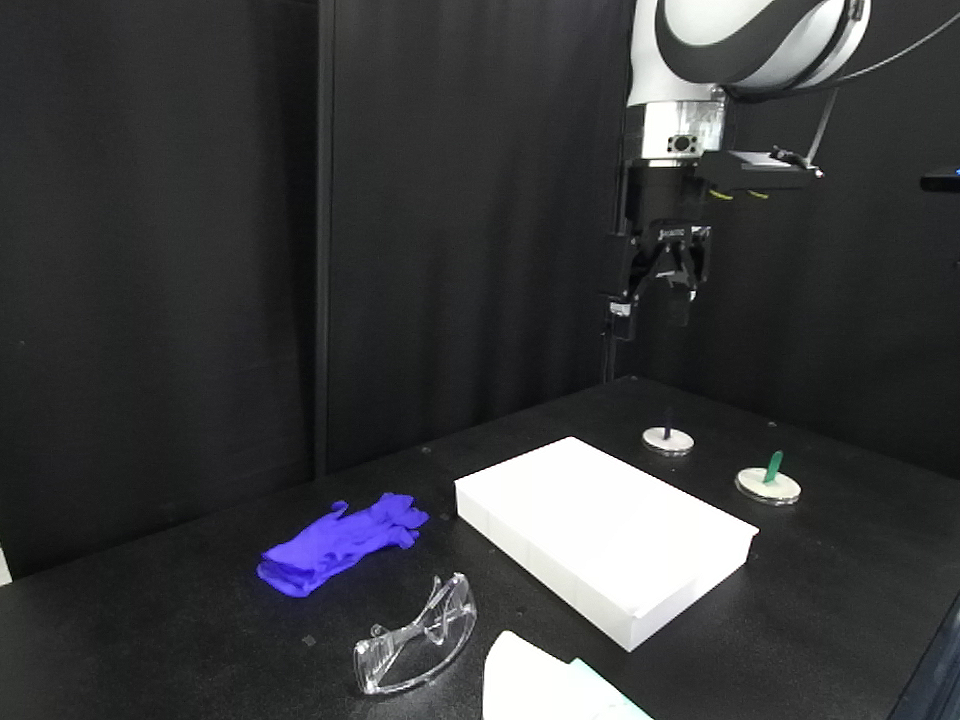}{lab\_goggles\_to\_box}{The safety glasses go in the white box.}\label{fig:eval:lab_goggles_to_box}

  \vspace{1.0em}

  {\bfseries Memory}\par\smallskip
  \begin{subfigure}[t]{0.19\linewidth}\end{subfigure}\hfill
  \evaltaskcard{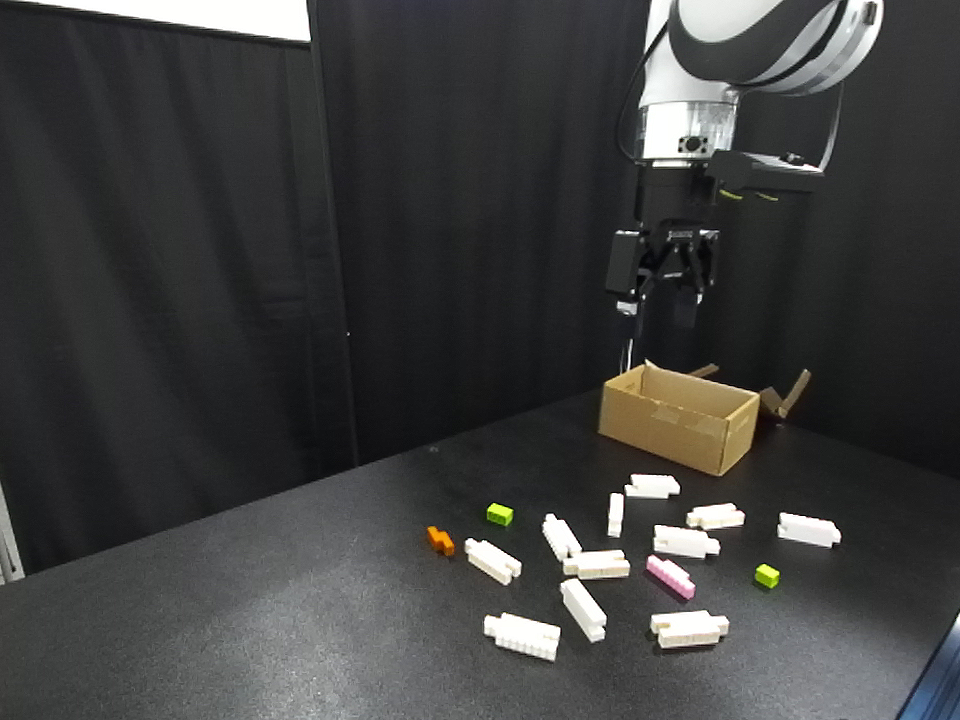}{two\_white\_blocks}{Place exactly two white blocks in the brown box.}\label{fig:eval:two_white_blocks}\hfill
  \evaltaskcard{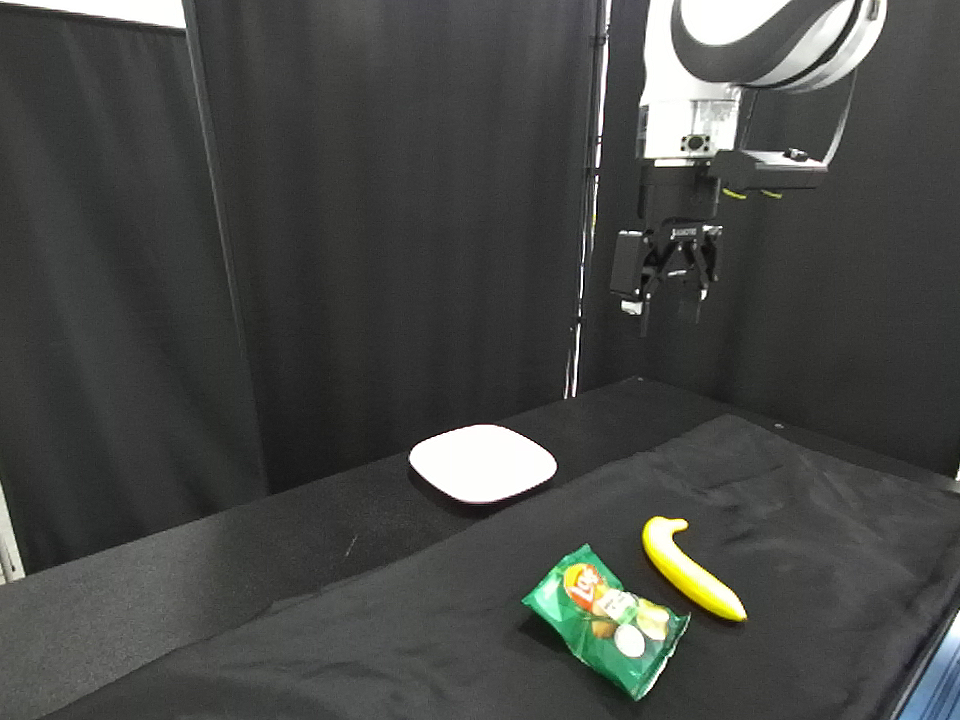}{healthy\_and\_unhealthy\_snack}{I want exactly one healthy snack and exactly one unhealthy snack on the plate.}\label{fig:eval:healthy_and_unhealthy_snack}\hfill
  \evaltaskcard{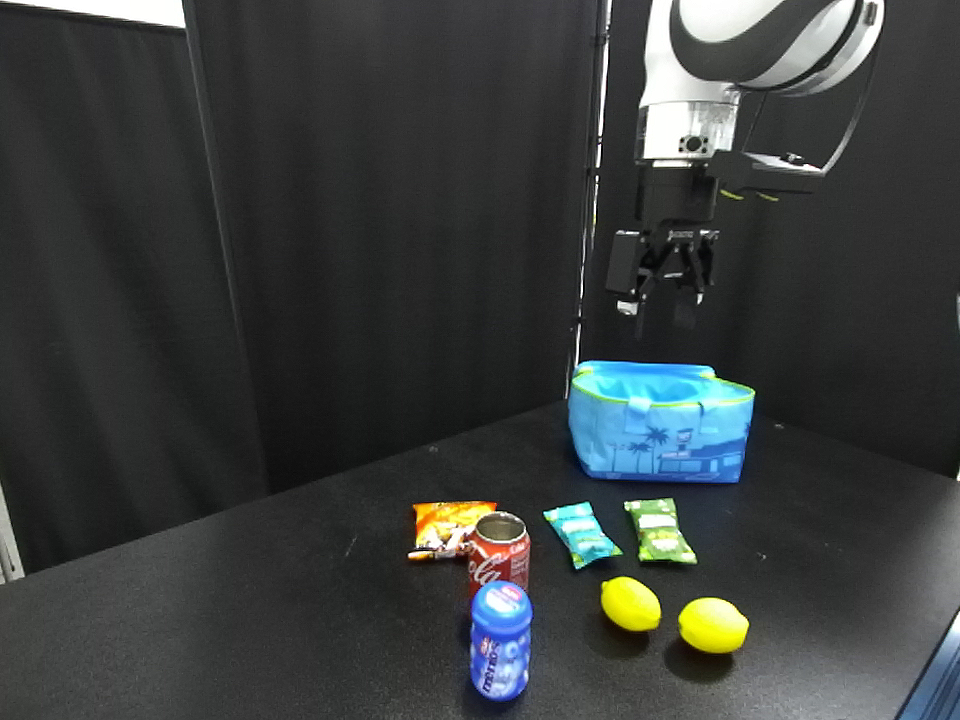}{protein\_and\_sour\_in\_bag}{Place exactly one high-protein snack and exactly one sour fruit in the bag.}\label{fig:eval:protein_and_sour_in_bag}\hfill
  \begin{subfigure}[t]{0.19\linewidth}\end{subfigure}

  \caption{Per-task starting frames for the physical evaluation suite, grouped by test-time-compute axis: \textbf{Intelligence} (top, $5$ tasks),
  \textbf{Skill} (middle, $5$ tasks), and \textbf{Memory} (bottom, $3$ tasks). Each card
  shows the starting-frame scene image, with the task ID and its instruction
  (\emph{italicized}) below; see Table~\ref{tab:app:droid_tasks} for the same tasks in
  tabular form.}
  \label{fig:eval-tasks}
\end{figure}

\subsection{Synthetic-Data Pipeline for the Hardware Router}
\label{app:bench:synthetic}

Physical rollouts are expensive, so we train the DROID router on synthetic routing data
rather than robot executions. At a high level, the pipeline (Fig.~\ref{fig:synthetic_pipeline}) follows these steps:

\begin{figure}[t]
\centering
\resizebox{\linewidth}{!}{%
\begin{tikzpicture}[
  font=\small,
  >={Stealth[length=2.4mm,width=1.8mm]},
  box/.style={draw, rounded corners=2pt, align=center, inner sep=5pt,
              fill=gray!6, minimum height=1.15cm},
  arr/.style={->, thick, black!55},
  alab/.style={midway, above, font=\scriptsize, text=black!70, align=center, inner sep=2pt},
  gcap/.style={font=\scriptsize, text=black!65, align=center},
  claudeburst/.pic={
    \fill[claudeorange] (0,0) circle (1.5pt);
    \foreach \a/\f in {4/1.0, 35/0.74, 64/1.06, 96/0.80, 123/1.0, 152/0.82,
                       185/1.05, 213/0.76, 245/1.03, 274/0.82, 306/1.04, 334/0.78}{
      \draw[claudeorange, line width=1.0pt, line cap=round] (0,0) -- (\a:{\f*8pt});
    }
  },
]
\def\gs{4.6mm} 

\node[box] (task) {canonical\\tasks\\[1pt]{\scriptsize scene + example}};

\coordinate (varc) at ([xshift=3.7cm]task.east);
\foreach \d in {2,1}{
  \draw[draw=black!45, fill=gray!12, rounded corners=2pt]
    ($(varc)+(-1.0cm,-0.575cm)+({\d*1.2mm},{\d*1.2mm})$) rectangle
    ($(varc)+( 1.0cm, 0.575cm)+({\d*1.2mm},{\d*1.2mm})$);
}
\node[box, minimum width=2.0cm, anchor=center] (var) at (varc)
     {feasible instruction\\variants\\[1pt]{\scriptsize balanced across}\\[-1pt]{\scriptsize routing classes}};

\coordinate (pgc) at ([xshift=3.05cm]var.east);
\foreach \r in {0,1,2}\foreach \c in {0,1,2}{
  \node[draw=black!55, fill=blue!6, minimum size=\gs, inner sep=0]
       (pg-\r-\c) at ($(pgc)+({\c*\gs-\gs},{\gs-\r*\gs})$) {};
}
\node[draw=none, fit=(pg-0-0)(pg-2-2)] (pg) {};
\node[gcap, below=1.5mm of pg] {plans\\{\scriptsize variant\,$\times$\,model}};

\coordinate (jgc) at ([xshift=2.55cm]pg.east);
\foreach \r in {0,1,2}\foreach \c in {0,1,2}{
  \pgfmathparse{\c>=\r ? "green!55!white" : "red!45!white"}\edef\cc{\pgfmathresult}
  \node[draw=black!55, fill=\cc, minimum size=\gs, inner sep=0]
       (jg-\r-\c) at ($(jgc)+({\c*\gs-\gs},{\gs-\r*\gs})$) {};
}
\node[draw=none, fit=(jg-0-0)(jg-2-2)] (jg) {};
\node[gcap, below=1.5mm of jg] {steerability\\{\scriptsize labels}};

\node[box, right=2.05cm of jg] (router) {router\\[1pt]{\scriptsize instr\,$+$\,scene}\\[-1pt]{\scriptsize $\to$ best planner}};

\draw[arr] (task) -- (var) node[alab](l1){(1) sample +\\generate instructions};
\draw[arr] (var) -- (pg.west) node[alab, pos=0.58](l2){(2) query\\$M$ models $\times$ modes};
\draw[arr] (pg) -- (jg) node[alab](l3){(3) LLM\\judge};
\pic at ([yshift=3.2mm]l1.north) {claudeburst};
\pic at ([yshift=3.2mm]l3.north) {claudeburst};
\draw[arr] (jg) -- (router) node[alab]{(4)\\train};
\end{tikzpicture}}
\caption{Synthetic-data pipeline for the hardware router (detailed below):
(1) generate feasible instruction variants per scene, (2) decompose each with every
candidate planner, (3) LLM-judge plan steerability (\textcolor{green!55!black}{green} $=$
steerable, \textcolor{red!70!black}{red} $=$ not), and (4) train the router on these labels.}
\label{fig:synthetic_pipeline}
\end{figure}
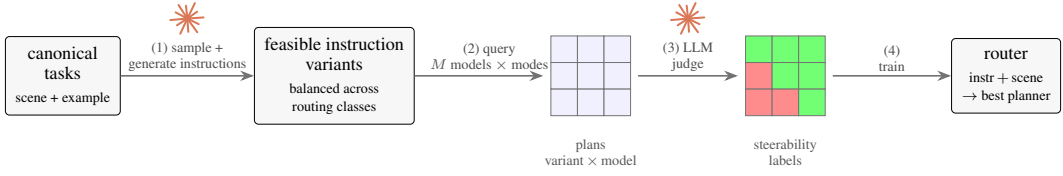

\setlength{\itemsep}{0.25em}
\quad \textbf{Generate variants.} Each canonical task is a representative scene paired with
one example instruction. For each scene we generate many additional feasible instructions
(the example serves only as a seed), indexed by routing axis (\texttt{INTELLIGENCE\_FORM} /
\texttt{SKILL\_COMMAND} / \texttt{MEMORY}) and balanced across classes. The classes are
chosen to \emph{carry} the routing signal: \texttt{MEMORY} spans the memory cells
(memory-needed / neither / -hurts), and \texttt{SKILL\_COMMAND} contrasts rigid verbs
(\emph{put}, \emph{pick}) with contextual ones (\emph{close}, \emph{fold}, \emph{wipe},
\emph{put-on}).

\quad \textbf{Query planners.} For each variant we query every candidate planner
(\eg 9B vs.\ 27B, think vs.\ no-think, memory vs.\ memory-free) for its subgoal decomposition
against the grounded frame. Nothing runs on the robot; the output is a per-(model, variant) grid of
plans.

\quad \textbf{Judge steerability.} With no execution, an LLM judge proxies success: a
decomposition is labeled a success iff its subgoals fall in the region of simple DROID tasks
the downstream policy reliably executes, and a failure if they are too complex, abstract, or
out-of-distribution. This yields per-(model, variant) steerability labels.

\quad \textbf{Train the router.} The steerability labels are the routing targets: from
instruction $+$ scene, the router selects the planner whose plan stays in the
high-steerability region --- \eg memory vs.\ memory-free, or a larger vs.\ smaller planner
for complex vs.\ simple skill commands.

\subsection{RoboMME}
\label{app:bench:robomme}

RoboMME~\citep{dai2026robomme} benchmarks memory-augmented robotic generalist policies:
a suite of long-horizon manipulation tasks evaluated over a family of memory-augmented
VLA (MME-VLA) variants built on a $\pi_{0.5}$ backbone~\citep{black2025pi05}. We use the
benchmark unmodified; all results are drawn from the RoboMME leaderboard test set (the
public evaluation results in
\texttt{Yinpei/selected\_robomme\_eval\_results}).

\paragraph{Difficulty suites.} In addition to memory type, tasks are stratified into \emph{easy}, \emph{medium}, and
\emph{hard} levels based on scene clutter, horizon length, and environmental
dynamics~\citep{dai2026robomme}; our per-suite Pareto analysis (Fig.~3 of the main paper)
routes within each level.

\paragraph{Routed memory variants.} From the RoboMME variant set we route over five
selected memory architectures: \texttt{FrameSamp} and \texttt{TokenDrop} (perceptual memory
that sub-samples / drops past visual tokens, adding no parameters), \texttt{SimpleSG} and
\texttt{GroundSG} (symbolic memory that summarizes history as language subgoals via an
auxiliary planner pass), and \texttt{MemER}~\citep{sridhar2025memer} (retrieval over
recalled keyframe history).

\paragraph{Cost metric (TFLOPs).} We report inference cost via the standard $2NT$
scaling-law approximation (two FLOPs per parameter per token),
$C = 2 \cdot N_{\text{params}} \cdot (T_{\text{in}} + T_{\text{out}})$. Each memory method
is costed as the $\pi_{0}$ planner \emph{base} cost plus the compute of its memory
mechanism:
\begin{equation}
C_{\text{method}} \;=\; \underbrace{C_{\text{base}}}_{\text{vision + VLM prefix + 10-step diffusion decode}} \;+\; \underbrace{2\, N_{\text{aux}}\, T_{\text{mem}}}_{\text{memory overhead}} .
\end{equation}
The constants are read from our stack: the $\pi_{0}$ backbone~\citep{black2025pi0} is a
\texttt{gemma\_2b} VLM ($N_{\text{vlm}}=2.114$B) plus a \texttt{gemma\_300m} action expert
($N_{\text{exp}}=0.302$B), and the base cost ($C_{\text{base}}\approx 3.34$ TFLOPs) is
vision $+$ the VLM prefix ($176$ tokens) $+$ $10$ denoising steps over the $51$-token
suffix. The symbolic and retrieval methods invoke an auxiliary Qwen3-VL-4B-Instruct planner
($N_{\text{aux}}\approx 4.0$B)~\citep{bai2025qwen3vltechnicalreport}; the perceptual methods
only re-weight existing visual tokens (sample / drop) and add no parameters, so their cost
$\approx C_{\text{base}}$. Per-method values then follow from the token counts
(Table~\ref{tab:app:robomme_flops}), giving an $\approx 8.6\times$ spread from
\texttt{FrameSamp} to \texttt{MemER}.

\begin{table}[H]
\centering
\caption{Memory-cost derivation for our reported figure. Every method is the $\pi_{0}$
planner base cost plus a memory overhead. TFLOPs reproduce via $2NT$.}
\label{tab:app:robomme_flops}
\footnotesize
\begin{tabular}{@{}llrr@{}}
\toprule
\textbf{Method} & \textbf{Accounting} & \textbf{$T_{\text{mem}}$} & \textbf{TFLOPs} \\
\midrule
\texttt{FrameSamp} / \texttt{TokenDrop} & base only (token re-weighting) & ---        & 3.338 \\
\texttt{SimpleSG} / \texttt{GroundSG}   & base $+$ Qwen3-VL-4B subgoal pass & $\sim$594  & 8.091 \\
\texttt{MemER}                          & base $+$ Qwen3-VL-4B keyframe retrieval & $\sim$3178 & 28.761 \\
\bottomrule
\end{tabular}
\end{table}

\noindent The token budgets are estimates: $\approx 594$ tokens for the symbolic pass (one
frame $+$ subgoal text) and $\approx 3178$ for \texttt{MemER}'s $\approx 12$ recalled keyframes.

\paragraph{Closed-weight variants and the physical cost proxy.} The Gemini-predictor
symbolic variants are omitted from the FLOPs accounting: their parameter count is unknown,
so $2NT$ cannot be evaluated. This is also why the physical memory experiment falls back to
\emph{input-frame count} as its cost proxy rather than FLOPs (main paper, Sec.~5.3).

\paragraph{Relation to RoboMME's reported cost.} Our reported FLOPs differ from
RoboMME~\citep[Fig.~4]{dai2026robomme} ($3.75$ / $6.26$ / $10.15$ / $15.37$) because of four
accounting choices. (i) \emph{Base policy}: we price the $\pi_{0}$ planner ($3.34$
TFLOPs) whereas RoboMME prices $\pi_{0.5}$ ($3.75$). (ii) \emph{Perceptual memory}: we
treat \texttt{FrameSamp}/\texttt{TokenDrop} as token re-weighting with no added parameters
(cost $\approx$ base, $3.34$), whereas RoboMME charges the modulation memory-token overhead
through the VLM ($+{\sim}2.5 \rightarrow 6.26$). (iii) \emph{Retrieval depth}: our \texttt{MemER} recalls a longer keyframe history ($\approx 12$ frames, $28.76$) than RoboMME's configuration ($\approx 5$ frames,
$15.37$). All use the $2NT$ estimate, so relative orderings hold even where magnitudes
differ, and the largest gap (\texttt{MemER}) is almost entirely choice (iii).

\section{Prompts}
\label{app:prompts}

\subsection{Synthetic-Pipeline Prompts (Task Generation and LLM-as-Judge)}
\label{app:prompts:synthetic}

The synthetic-data pipeline (Appendix~\ref{app:bench:synthetic}) uses two prompts: a
task-generation prompt that proposes the next subtask, and an LLM-as-Judge prompt that
labels/scores the candidate plans. We give the task-generation prompt first, followed by
the judge.

\subsubsection{Task-Generation Prompt (Synthetic Data and Hardware)}
\label{app:prompts:taskgen}

The prompt below drives the synthetic-data pipeline and the hardware task generation:
given the user request and the scene image, a large VLM proposes the immediate next subtask
in the plain imperative style the downstream VLA policy can execute.
\texttt{<vlm\_input>} is replaced with the user request at generation time.

\promptfile{prompts/subtask_generation.txt}

\subsubsection{Synthetic LLM-as-Judge Prompts}
\label{app:prompts:judge}

The synthetic-data pipeline (Appendix~\ref{app:bench:synthetic}) uses two LLM-as-Judge
prompts. The first labels whether a planner's predicted single-step subtask would drive the
\emph{same physical action} as the ground-truth primitive---i.e.\ whether it lands in the
steerable, groundable primitive distribution rather than merely string-matching. The second
labels whether a task requires \emph{memory} across steps or can instead be re-prompted
step-by-step from the scene alone. Both return strict JSON; we show condensed versions below
(the full synonym lists and few-shot anchors are omitted for space), with fields in
\texttt{\{\{...\}\}} interpolated per call.

\paragraph{Subtask-correctness judge.} Decides whether the predicted subtask grounds to the
same action as the ground truth, given the scene.

\promptfile{prompts/llm_judge_subtask_correctness.txt}

\paragraph{Memory-need / routing judge.} Decides whether the task needs cross-step memory or
is observable step-by-step from the scene (gripper state and source-object disappearance).

\promptfile{prompts/llm_judge_memory_routing.txt}

\subsection{VLABench Planning Prompt}
\label{app:prompts:vlabench}

The system prompt below is given to every VLABench planner. It defines nine skills
(\texttt{pick}, \texttt{place}, \texttt{press}, \texttt{open\_door}, \texttt{insert},
\texttt{pull}, \texttt{pour}, \texttt{push}, \texttt{lift}) with their JSON call formats,
describes the two-image input (unlabeled views $+$ numbered views), and asks the model to
emit a JSON skill-call sequence matching one of the allowed sub-skill patterns (e.g.\
\texttt{["pick","place"]}, \texttt{["pick","pour","place"]}). We use this English prompt; an
equivalent Chinese version (\texttt{eval\_vlm\_zh.txt}) is also available.

\promptfile{prompts/vlabench_planner.txt}

\section{Complete Router Sweep (Representative Pair)}
\label{app:full_sweep}

The per-pair ablation tables in Appendix~\ref{app:ablations} are summaries: for each
embedding-fusion method they list only the single best router under each selection criterion
(best average score and best $\eta$). For completeness,
Table~\ref{tab:app:full_router_sweep} gives the \emph{full} sweep for one representative pair
(GLM-4.1V-Thinking vs.\ GLM-4.6V-Flash) --- every trained router (architecture $\times$
fusion method $\times$ head), followed by the fixed baselines, oracles, the complete OOD sweep (every fusion method
$\times$ $\gamma$), and the full budget-oracle sweep --- so the reader can see the
complete option space behind the summarized tables. The same sweep is available for every
pair in our released results.

{\footnotesize
\setlength{\LTpre}{4pt}\setlength{\LTpost}{4pt}\setlength{\tabcolsep}{4pt}
\renewcommand{\arraystretch}{1.05}
\def\_{\textunderscore\allowbreak}
\begin{longtable}{@{}>{\raggedright\arraybackslash\ttfamily}p{0.18\linewidth}>{\raggedright\arraybackslash\ttfamily}p{0.34\linewidth} r r r@{}}
\caption{\textbf{Complete router sweep for a representative pair (GLM-4.1V-Thinking vs.\ GLM-4.6V-Flash), sorted by $\eta$.}
Every trained router, i.e.\ each (architecture $\times$ embedding-fusion method $\times$ head) combination, ordered
highest-to-lowest $\eta$. The \emph{Fusion Method} sets how the router-input embedding is formed: \texttt{concat} /
\texttt{normalize\_concat} / \texttt{average} / \texttt{weighted\_average} combine the \texttt{text} (instruction) and
\texttt{image} (scene; SigLIP-2) embeddings, while \texttt{only\_text} / \texttt{only\_image} use one of them alone and
\texttt{only\_vla} uses the $\pi_0$ VLA policy embedding alone (full definitions in App.~\ref{app:router:embeddings}).
Avg score and $\eta$ are $\times100$; avg cost in seconds; $\eta$ cells shaded
\colorbox{red!25!white}{red} (low) to \colorbox{green!25!white}{green} (high). Fixed baselines, oracles, the OOD sweep, and the budget-oracle frontier are listed in full at the foot of the table.}
\label{tab:app:full_router_sweep}\\
\toprule
\normalfont\textbf{Fusion Method} & \normalfont\textbf{Router (arch\_fusion\_head)} & \normalfont\textbf{Avg Score} & \normalfont\textbf{Avg Cost (s)} & \normalfont\textbf{$\eta$} \\
\midrule\endfirsthead
\toprule
\normalfont\textbf{Fusion Method} & \normalfont\textbf{Router (arch\_fusion\_head)} & \normalfont\textbf{Avg Score} & \normalfont\textbf{Avg Cost (s)} & \normalfont\textbf{$\eta$} \\
\midrule\endhead
\midrule\multicolumn{5}{r}{\normalfont\emph{continued on next page}}\\\endfoot
\bottomrule\endlastfoot
only\_text & linear\_only\_text\_soft$\lambda$0.1 & \normalfont 51.82 & \normalfont 31.86 & \cellcolor{green!75!red!25!white}\normalfont 75.29 \\
concat & linear\_concat\_soft$\lambda$0.1 & \normalfont 51.79 & \normalfont 31.82 & \cellcolor{green!75!red!25!white}\normalfont 75.26 \\
normalize\_concat & linear\_normalize\_concat\_soft$\lambda$0.1 & \normalfont 51.79 & \normalfont 31.82 & \cellcolor{green!75!red!25!white}\normalfont 75.26 \\
normalize\_concat & linear\_normalize\_concat\_tie\_regression & \normalfont 49.89 & \normalfont 26.52 & \cellcolor{green!75!red!25!white}\normalfont 75.12 \\
concat & linear\_concat\_tie\_regression & \normalfont 49.89 & \normalfont 26.57 & \cellcolor{green!75!red!25!white}\normalfont 75.08 \\
weighted\_average\_text075 & linear\_weighted\_average\_soft$\lambda$0.1 & \normalfont 51.69 & \normalfont 32.07 & \cellcolor{green!75!red!25!white}\normalfont 74.72 \\
only\_image & mlp\_only\_image\_h2048\_1024 & \normalfont 51.47 & \normalfont 31.68 & \cellcolor{green!75!red!25!white}\normalfont 74.68 \\
average & linear\_average\_tie\_regression & \normalfont 49.51 & \normalfont 25.83 & \cellcolor{green!74!red!25!white}\normalfont 74.46 \\
weighted\_average\_text075 & linear\_weighted\_average\_tie\_regression & \normalfont 49.48 & \normalfont 25.75 & \cellcolor{green!74!red!25!white}\normalfont 74.41 \\
only\_text & mlp\_only\_text\_h2048\_1024 & \normalfont 50.99 & \normalfont 30.94 & \cellcolor{green!74!red!25!white}\normalfont 74.36 \\
weighted\_average\_text075 & mlp\_weighted\_average\_h2048\_1024 & \normalfont 51.13 & \normalfont 31.29 & \cellcolor{green!74!red!25!white}\normalfont 74.32 \\
average & linear\_average\_soft$\lambda$0.1 & \normalfont 51.89 & \normalfont 32.79 & \cellcolor{green!74!red!25!white}\normalfont 74.22 \\
weighted\_average\_text025 & linear\_weighted\_average\_soft$\lambda$0.1 & \normalfont 52.17 & \normalfont 33.23 & \cellcolor{green!74!red!25!white}\normalfont 74.20 \\
only\_text & linear\_only\_text\_tie\_regression & \normalfont 49.54 & \normalfont 26.62 & \cellcolor{green!74!red!25!white}\normalfont 74.06 \\
weighted\_average\_text025 & linear\_weighted\_average\_tie\_regression & \normalfont 49.57 & \normalfont 26.78 & \cellcolor{green!74!red!25!white}\normalfont 74.05 \\
concat & mlp\_concat\_h2048\_1024 & \normalfont 50.80 & \normalfont 30.82 & \cellcolor{green!74!red!25!white}\normalfont 74.01 \\
weighted\_average\_text025 & mlp\_weighted\_average\_h2048\_1024\_tie\_regression & \normalfont 51.99 & \normalfont 33.16 & \cellcolor{green!74!red!25!white}\normalfont 73.90 \\
concat & knn\_k5\_concat & \normalfont 51.27 & \normalfont 31.99 & \cellcolor{green!74!red!25!white}\normalfont 73.83 \\
concat & prknn\_k5\_concat & \normalfont 51.27 & \normalfont 31.99 & \cellcolor{green!74!red!25!white}\normalfont 73.83 \\
normalize\_concat & knn\_k5\_normalize\_concat & \normalfont 51.27 & \normalfont 31.99 & \cellcolor{green!74!red!25!white}\normalfont 73.83 \\
normalize\_concat & prknn\_k5\_normalize\_concat & \normalfont 51.27 & \normalfont 31.99 & \cellcolor{green!74!red!25!white}\normalfont 73.83 \\
weighted\_average\_text025 & knn\_k5\_weighted\_average\_w0.25 & \normalfont 51.06 & \normalfont 31.55 & \cellcolor{green!74!red!25!white}\normalfont 73.83 \\
weighted\_average\_text025 & prknn\_k5\_weighted\_average & \normalfont 51.06 & \normalfont 31.55 & \cellcolor{green!74!red!25!white}\normalfont 73.83 \\
average & mlp\_average\_h2048\_1024 & \normalfont 50.37 & \normalfont 29.93 & \cellcolor{green!74!red!25!white}\normalfont 73.77 \\
only\_image & linear\_only\_image\_tie\_regression & \normalfont 49.61 & \normalfont 27.42 & \cellcolor{green!74!red!25!white}\normalfont 73.73 \\
average & mlp\_average\_h2048\_1024\_tie\_regression & \normalfont 52.62 & \normalfont 34.19 & \cellcolor{green!74!red!25!white}\normalfont 73.60 \\
average & knn\_k5\_average & \normalfont 51.15 & \normalfont 31.96 & \cellcolor{green!74!red!25!white}\normalfont 73.58 \\
average & prknn\_k5\_average & \normalfont 51.15 & \normalfont 31.96 & \cellcolor{green!74!red!25!white}\normalfont 73.58 \\
weighted\_average\_text025 & mlp\_weighted\_average\_h2048\_1024 & \normalfont 50.42 & \normalfont 30.31 & \cellcolor{green!74!red!25!white}\normalfont 73.55 \\
weighted\_average\_text075 & kmeans\_weighted\_average & \normalfont 49.80 & \normalfont 28.62 & \cellcolor{green!73!red!25!white}\normalfont 73.38 \\
only\_image & linear\_only\_image\_soft$\lambda$0.1 & \normalfont 51.37 & \normalfont 32.58 & \cellcolor{green!73!red!25!white}\normalfont 73.30 \\
average & kmeans\_average & \normalfont 49.53 & \normalfont 27.80 & \cellcolor{green!73!red!25!white}\normalfont 73.27 \\
weighted\_average\_text025 & linear\_weighted\_average\_regression & \normalfont 52.58 & \normalfont 34.35 & \cellcolor{green!73!red!25!white}\normalfont 73.23 \\
only\_text & ovr\_only\_text & \normalfont 49.71 & \normalfont 28.59 & \cellcolor{green!73!red!25!white}\normalfont 73.16 \\
normalize\_concat & mlp\_normalize\_concat\_h2048\_1024 & \normalfont 50.19 & \normalfont 30.13 & \cellcolor{green!73!red!25!white}\normalfont 73.12 \\
normalize\_concat & mlp\_normalize\_concat\_h2048\_1024\_tie\_regression & \normalfont 52.17 & \normalfont 33.90 & \cellcolor{green!73!red!25!white}\normalfont 73.12 \\
only\_text & ovr\_only\_text\_tie\_regression & \normalfont 50.66 & \normalfont 31.39 & \cellcolor{green!73!red!25!white}\normalfont 73.05 \\
weighted\_average\_text075 & ovr\_weighted\_average\_w0.75 & \normalfont 49.40 & \normalfont 27.63 & \cellcolor{green!73!red!25!white}\normalfont 73.03 \\
weighted\_average\_text075 & knn\_k5\_weighted\_average\_w0.75 & \normalfont 50.83 & \normalfont 31.77 & \cellcolor{green!73!red!25!white}\normalfont 73.01 \\
weighted\_average\_text075 & prknn\_k5\_weighted\_average & \normalfont 50.83 & \normalfont 31.77 & \cellcolor{green!73!red!25!white}\normalfont 73.01 \\
only\_image & linear\_only\_image\_regression & \normalfont 52.77 & \normalfont 34.69 & \cellcolor{green!73!red!25!white}\normalfont 73.00 \\
weighted\_average\_text075 & ovr\_weighted\_average\_w0.75\_tie\_regression & \normalfont 50.75 & \normalfont 31.62 & \cellcolor{green!73!red!25!white}\normalfont 73.00 \\
only\_text & kmeans\_only\_text & \normalfont 49.80 & \normalfont 29.14 & \cellcolor{green!73!red!25!white}\normalfont 72.98 \\
concat & mlp\_concat\_h2048\_1024\_tie\_regression & \normalfont 51.54 & \normalfont 33.12 & \cellcolor{green!73!red!25!white}\normalfont 72.93 \\
weighted\_average\_text075 & knn\_k5\_weighted\_average\_w0.75\_tie\_regression & \normalfont 52.00 & \normalfont 33.86 & \cellcolor{green!73!red!25!white}\normalfont 72.83 \\
weighted\_average\_text075 & prknn\_k5\_weighted\_average\_tie\_regression & \normalfont 52.00 & \normalfont 33.86 & \cellcolor{green!73!red!25!white}\normalfont 72.83 \\
concat & linear\_concat\_regression & \normalfont 52.38 & \normalfont 34.35 & \cellcolor{green!73!red!25!white}\normalfont 72.81 \\
normalize\_concat & linear\_normalize\_concat\_regression & \normalfont 52.38 & \normalfont 34.35 & \cellcolor{green!73!red!25!white}\normalfont 72.81 \\
concat & knn\_k5\_concat\_tie\_regression & \normalfont 52.20 & \normalfont 34.18 & \cellcolor{green!73!red!25!white}\normalfont 72.74 \\
concat & prknn\_k5\_concat\_tie\_regression & \normalfont 52.20 & \normalfont 34.18 & \cellcolor{green!73!red!25!white}\normalfont 72.74 \\
normalize\_concat & knn\_k5\_normalize\_concat\_tie\_regression & \normalfont 52.20 & \normalfont 34.18 & \cellcolor{green!73!red!25!white}\normalfont 72.74 \\
normalize\_concat & prknn\_k5\_normalize\_concat\_tie\_regression & \normalfont 52.20 & \normalfont 34.18 & \cellcolor{green!73!red!25!white}\normalfont 72.74 \\
weighted\_average\_text025 & ovr\_weighted\_average\_w0.25\_tie\_regression & \normalfont 51.13 & \normalfont 32.60 & \cellcolor{green!73!red!25!white}\normalfont 72.71 \\
weighted\_average\_text075 & mlp\_weighted\_average\_h2048\_1024\_tie\_regression & \normalfont 52.37 & \normalfont 34.41 & \cellcolor{green!73!red!25!white}\normalfont 72.68 \\
average & ovr\_average\_tie\_regression & \normalfont 51.36 & \normalfont 33.04 & \cellcolor{green!73!red!25!white}\normalfont 72.63 \\
average & knn\_k5\_average\_tie\_regression & \normalfont 52.13 & \normalfont 34.17 & \cellcolor{green!73!red!25!white}\normalfont 72.60 \\
average & prknn\_k5\_average\_tie\_regression & \normalfont 52.13 & \normalfont 34.17 & \cellcolor{green!73!red!25!white}\normalfont 72.60 \\
only\_image & ovr\_only\_image\_tie\_regression & \normalfont 52.37 & \normalfont 34.45 & \cellcolor{green!73!red!25!white}\normalfont 72.60 \\
only\_vla & knn\_k5\_only\_vla & \normalfont 50.94 & \normalfont 32.34 & \cellcolor{green!73!red!25!white}\normalfont 72.60 \\
only\_vla & prknn\_k5\_only\_vla & \normalfont 50.94 & \normalfont 32.34 & \cellcolor{green!73!red!25!white}\normalfont 72.60 \\
only\_text & mlp\_only\_text\_h2048\_1024\_tie\_regression & \normalfont 52.61 & \normalfont 34.73 & \cellcolor{green!73!red!25!white}\normalfont 72.59 \\
average & linear\_average\_regression & \normalfont 52.61 & \normalfont 34.74 & \cellcolor{green!73!red!25!white}\normalfont 72.57 \\
only\_text & knn\_k5\_only\_text & \normalfont 50.51 & \normalfont 31.49 & \cellcolor{green!73!red!25!white}\normalfont 72.55 \\
only\_text & prknn\_k5\_only\_text\_tie\_regression & \normalfont 52.05 & \normalfont 34.12 & \cellcolor{green!72!red!25!white}\normalfont 72.50 \\
only\_text & prknn\_k5\_only\_text & \normalfont 50.40 & \normalfont 31.29 & \cellcolor{green!72!red!25!white}\normalfont 72.49 \\
weighted\_average\_text025 & knn\_k5\_weighted\_average\_w0.25\_tie\_regression & \normalfont 52.24 & \normalfont 34.43 & \cellcolor{green!72!red!25!white}\normalfont 72.38 \\
weighted\_average\_text025 & prknn\_k5\_weighted\_average\_tie\_regression & \normalfont 52.24 & \normalfont 34.43 & \cellcolor{green!72!red!25!white}\normalfont 72.38 \\
only\_vla & mlp\_only\_vla\_h2048\_1024\_tie\_regression & \normalfont 50.79 & \normalfont 32.23 & \cellcolor{green!72!red!25!white}\normalfont 72.37 \\
only\_vla & mlp\_only\_vla\_h2048\_1024 & \normalfont 50.62 & \normalfont 32.05 & \cellcolor{green!72!red!25!white}\normalfont 72.19 \\
only\_text & knn\_k5\_only\_text\_tie\_regression & \normalfont 52.10 & \normalfont 34.41 & \cellcolor{green!72!red!25!white}\normalfont 72.11 \\
average & ovr\_average & \normalfont 48.88 & \normalfont 27.28 & \cellcolor{green!72!red!25!white}\normalfont 71.82 \\
only\_vla & knn\_k5\_only\_vla\_tie\_regression & \normalfont 52.45 & \normalfont 34.99 & \cellcolor{green!72!red!25!white}\normalfont 71.75 \\
only\_vla & prknn\_k5\_only\_vla\_tie\_regression & \normalfont 52.45 & \normalfont 34.99 & \cellcolor{green!72!red!25!white}\normalfont 71.75 \\
only\_image & knn\_k5\_only\_image & \normalfont 50.54 & \normalfont 32.40 & \cellcolor{green!72!red!25!white}\normalfont 71.56 \\
only\_image & prknn\_k5\_only\_image & \normalfont 50.54 & \normalfont 32.40 & \cellcolor{green!72!red!25!white}\normalfont 71.56 \\
weighted\_average\_text025 & ovr\_weighted\_average\_w0.25 & \normalfont 48.74 & \normalfont 27.12 & \cellcolor{green!72!red!25!white}\normalfont 71.52 \\
concat & kmeans\_concat & \normalfont 48.51 & \normalfont 26.20 & \cellcolor{green!71!red!25!white}\normalfont 71.41 \\
normalize\_concat & kmeans\_normalize\_concat & \normalfont 48.51 & \normalfont 26.20 & \cellcolor{green!71!red!25!white}\normalfont 71.41 \\
only\_image & knn\_k5\_only\_image\_tie\_regression & \normalfont 52.29 & \normalfont 35.11 & \cellcolor{green!71!red!25!white}\normalfont 71.19 \\
only\_image & prknn\_k5\_only\_image\_tie\_regression & \normalfont 52.29 & \normalfont 35.11 & \cellcolor{green!71!red!25!white}\normalfont 71.19 \\
only\_vla & ovr\_only\_vla\_tie\_regression & \normalfont 52.58 & \normalfont 35.42 & \cellcolor{green!71!red!25!white}\normalfont 71.15 \\
only\_vla & ovr\_only\_vla & \normalfont 48.82 & \normalfont 28.29 & \cellcolor{green!71!red!25!white}\normalfont 70.98 \\
weighted\_average\_text075 & linear\_weighted\_average\_regression & \normalfont 52.78 & \normalfont 35.73 & \cellcolor{green!71!red!25!white}\normalfont 70.86 \\
weighted\_average\_text025 & kmeans\_weighted\_average & \normalfont 48.28 & \normalfont 26.20 & \cellcolor{green!71!red!25!white}\normalfont 70.74 \\
only\_vla & linear\_only\_vla\_tie\_regression & \normalfont 48.72 & \normalfont 28.46 & \cellcolor{green!71!red!25!white}\normalfont 70.57 \\
only\_text & linear\_only\_text\_regression & \normalfont 52.80 & \normalfont 35.94 & \cellcolor{green!70!red!25!white}\normalfont 70.44 \\
only\_image & ovr\_only\_image & \normalfont 48.30 & \normalfont 27.37 & \cellcolor{green!70!red!25!white}\normalfont 70.13 \\
concat & ovr\_concat\_tie\_regression & \normalfont 49.57 & \normalfont 31.61 & \cellcolor{green!70!red!25!white}\normalfont 70.08 \\
normalize\_concat & ovr\_normalize\_concat\_tie\_regression & \normalfont 49.57 & \normalfont 31.61 & \cellcolor{green!70!red!25!white}\normalfont 70.08 \\
normalize\_concat & ovr\_normalize\_concat & \normalfont 48.12 & \normalfont 27.25 & \cellcolor{green!70!red!25!white}\normalfont 69.69 \\
concat & ovr\_concat & \normalfont 47.98 & \normalfont 27.18 & \cellcolor{green!69!red!25!white}\normalfont 69.33 \\
only\_image & kmeans\_only\_image & \normalfont 47.60 & \normalfont 25.65 & \cellcolor{green!69!red!25!white}\normalfont 69.07 \\
concat & mlp\_concat\_h2048\_1024\_rank\_regression & \normalfont 53.33 & \normalfont 37.01 & \cellcolor{green!69!red!25!white}\normalfont 68.68 \\
normalize\_concat & mlp\_normalize\_concat\_h2048\_1024\_rank\_regression & \normalfont 53.50 & \normalfont 37.52 & \cellcolor{green!67!red!25!white}\normalfont 67.47 \\
only\_vla & linear\_only\_vla\_soft$\lambda$0.1 & \normalfont 47.21 & \normalfont 27.45 & \cellcolor{green!67!red!25!white}\normalfont 66.97 \\
concat & knn\_k5\_concat\_regression & \normalfont 53.34 & \normalfont 37.75 & \cellcolor{green!66!red!25!white}\normalfont 66.46 \\
concat & prknn\_k5\_concat\_regression & \normalfont 53.34 & \normalfont 37.75 & \cellcolor{green!66!red!25!white}\normalfont 66.46 \\
normalize\_concat & knn\_k5\_normalize\_concat\_regression & \normalfont 53.34 & \normalfont 37.75 & \cellcolor{green!66!red!25!white}\normalfont 66.46 \\
normalize\_concat & prknn\_k5\_normalize\_concat\_regression & \normalfont 53.34 & \normalfont 37.75 & \cellcolor{green!66!red!25!white}\normalfont 66.46 \\
average & knn\_k5\_average\_regression & \normalfont 53.30 & \normalfont 37.76 & \cellcolor{green!66!red!25!white}\normalfont 66.38 \\
average & prknn\_k5\_average\_regression & \normalfont 53.30 & \normalfont 37.76 & \cellcolor{green!66!red!25!white}\normalfont 66.38 \\
only\_text & knn\_k5\_only\_text\_regression & \normalfont 53.25 & \normalfont 37.75 & \cellcolor{green!66!red!25!white}\normalfont 66.34 \\
weighted\_average\_text075 & knn\_k5\_weighted\_average\_w0.75\_regression & \normalfont 53.43 & \normalfont 37.84 & \cellcolor{green!66!red!25!white}\normalfont 66.34 \\
weighted\_average\_text075 & prknn\_k5\_weighted\_average\_regression & \normalfont 53.43 & \normalfont 37.84 & \cellcolor{green!66!red!25!white}\normalfont 66.34 \\
only\_text & prknn\_k5\_only\_text\_regression & \normalfont 53.31 & \normalfont 37.84 & \cellcolor{green!66!red!25!white}\normalfont 66.15 \\
only\_image & mlp\_only\_image\_h2048\_1024\_tie\_regression & \normalfont 53.36 & \normalfont 37.90 & \cellcolor{green!66!red!25!white}\normalfont 66.02 \\
weighted\_average\_text025 & knn\_k5\_weighted\_average\_w0.25\_regression & \normalfont 53.40 & \normalfont 38.10 & \cellcolor{green!65!red!25!white}\normalfont 65.42 \\
weighted\_average\_text025 & prknn\_k5\_weighted\_average\_regression & \normalfont 53.40 & \normalfont 38.10 & \cellcolor{green!65!red!25!white}\normalfont 65.42 \\
concat & mlp\_concat\_h2048\_1024\_regression & \normalfont 53.43 & \normalfont 38.26 & \cellcolor{green!65!red!25!white}\normalfont 64.91 \\
only\_vla & linear\_only\_vla\_regression & \normalfont 52.84 & \normalfont 38.00 & \cellcolor{green!65!red!25!white}\normalfont 64.86 \\
only\_vla & mlp\_only\_vla\_h2048\_1024\_rank\_regression & \normalfont 52.09 & \normalfont 37.63 & \cellcolor{green!65!red!25!white}\normalfont 64.72 \\
average & mlp\_average\_h2048\_1024\_rank\_regression & \normalfont 53.29 & \normalfont 38.34 & \cellcolor{green!64!red!25!white}\normalfont 64.41 \\
average & mlp\_average\_h2048\_1024\_regression & \normalfont 53.47 & \normalfont 38.52 & \cellcolor{green!64!red!25!white}\normalfont 64.02 \\
concat & ovr\_concat\_rank\_regression & \normalfont 51.22 & \normalfont 37.39 & \cellcolor{green!64!red!25!white}\normalfont 63.89 \\
normalize\_concat & ovr\_normalize\_concat\_rank\_regression & \normalfont 51.22 & \normalfont 37.39 & \cellcolor{green!64!red!25!white}\normalfont 63.89 \\
concat & ovr\_concat\_regression & \normalfont 51.55 & \normalfont 37.62 & \cellcolor{green!64!red!25!white}\normalfont 63.84 \\
normalize\_concat & ovr\_normalize\_concat\_regression & \normalfont 51.55 & \normalfont 37.62 & \cellcolor{green!64!red!25!white}\normalfont 63.84 \\
only\_vla & knn\_k5\_only\_vla\_regression & \normalfont 53.46 & \normalfont 38.60 & \cellcolor{green!64!red!25!white}\normalfont 63.71 \\
only\_vla & prknn\_k5\_only\_vla\_regression & \normalfont 53.46 & \normalfont 38.60 & \cellcolor{green!64!red!25!white}\normalfont 63.71 \\
only\_image & knn\_k5\_only\_image\_regression & \normalfont 53.15 & \normalfont 38.49 & \cellcolor{green!64!red!25!white}\normalfont 63.66 \\
only\_vla & mlp\_only\_vla\_h2048\_1024\_regression & \normalfont 52.37 & \normalfont 38.18 & \cellcolor{green!64!red!25!white}\normalfont 63.50 \\
normalize\_concat & mlp\_normalize\_concat\_h2048\_1024\_regression & \normalfont 53.63 & \normalfont 38.80 & \cellcolor{green!63!red!25!white}\normalfont 63.20 \\
only\_image & prknn\_k5\_only\_image\_regression & \normalfont 53.35 & \normalfont 38.69 & \cellcolor{green!63!red!25!white}\normalfont 63.20 \\
only\_vla & kmeans\_only\_vla\_regression & \normalfont 46.88 & \normalfont 32.17 & \cellcolor{green!63!red!25!white}\normalfont 62.71 \\
only\_text & kmeans\_only\_text\_regression & \normalfont 46.27 & \normalfont 30.36 & \cellcolor{green!63!red!25!white}\normalfont 62.53 \\
weighted\_average\_text075 & kmeans\_weighted\_average\_regression & \normalfont 46.17 & \normalfont 30.41 & \cellcolor{green!62!red!25!white}\normalfont 62.21 \\
concat & linear\_concat\_rank\_regression & \normalfont 52.52 & \normalfont 38.81 & \cellcolor{green!62!red!25!white}\normalfont 61.52 \\
normalize\_concat & linear\_normalize\_concat\_rank\_regression & \normalfont 52.52 & \normalfont 38.81 & \cellcolor{green!62!red!25!white}\normalfont 61.52 \\
only\_image & mlp\_only\_image\_h2048\_1024\_rank\_regression & \normalfont 53.88 & \normalfont 39.41 & \cellcolor{green!61!red!25!white}\normalfont 61.02 \\
average & kmeans\_average\_regression & \normalfont 45.37 & \normalfont 29.65 & \cellcolor{green!60!red!25!white}\normalfont 60.49 \\
only\_image & mlp\_only\_image\_h2048\_1024\_regression & \normalfont 53.86 & \normalfont 39.65 & \cellcolor{green!60!red!25!white}\normalfont 59.84 \\
weighted\_average\_text075 & ovr\_weighted\_average\_w0.75\_rank\_regression & \normalfont 52.18 & \normalfont 39.16 & \cellcolor{green!60!red!25!white}\normalfont 59.68 \\
weighted\_average\_text025 & ovr\_weighted\_average\_w0.25\_regression & \normalfont 52.90 & \normalfont 39.49 & \cellcolor{green!59!red!25!white}\normalfont 59.35 \\
only\_vla & kmeans\_only\_vla & \normalfont 44.03 & \normalfont 22.76 & \cellcolor{green!59!red!25!white}\normalfont 59.29 \\
weighted\_average\_text025 & ovr\_weighted\_average\_w0.25\_rank\_regression & \normalfont 52.53 & \normalfont 39.39 & \cellcolor{green!59!red!25!white}\normalfont 59.27 \\
average & linear\_average\_rank\_regression & \normalfont 52.51 & \normalfont 39.39 & \cellcolor{green!59!red!25!white}\normalfont 59.22 \\
average & ovr\_average\_rank\_regression & \normalfont 52.41 & \normalfont 39.36 & \cellcolor{green!59!red!25!white}\normalfont 59.19 \\
only\_image & linear\_only\_image\_rank\_regression & \normalfont 52.83 & \normalfont 39.59 & \cellcolor{green!59!red!25!white}\normalfont 58.78 \\
weighted\_average\_text025 & linear\_weighted\_average\_rank\_regression & \normalfont 52.64 & \normalfont 39.59 & \cellcolor{green!59!red!25!white}\normalfont 58.53 \\
only\_text & ovr\_only\_text\_rank\_regression & \normalfont 52.58 & \normalfont 39.76 & \cellcolor{green!58!red!25!white}\normalfont 57.70 \\
average & ovr\_average\_regression & \normalfont 52.73 & \normalfont 39.88 & \cellcolor{green!57!red!25!white}\normalfont 57.34 \\
weighted\_average\_text075 & ovr\_weighted\_average\_w0.75\_regression & \normalfont 52.44 & \normalfont 39.87 & \cellcolor{green!57!red!25!white}\normalfont 56.99 \\
only\_text & ovr\_only\_text\_regression & \normalfont 52.72 & \normalfont 40.02 & \cellcolor{green!57!red!25!white}\normalfont 56.63 \\
weighted\_average\_text075 & linear\_weighted\_average\_rank\_regression & \normalfont 52.61 & \normalfont 40.20 & \cellcolor{green!56!red!25!white}\normalfont 55.64 \\
only\_vla & ovr\_only\_vla\_rank\_regression & \normalfont 53.25 & \normalfont 40.40 & \cellcolor{green!55!red!25!white}\normalfont 55.34 \\
only\_text & linear\_only\_text\_rank\_regression & \normalfont 52.56 & \normalfont 40.38 & \cellcolor{green!55!red!25!white}\normalfont 54.63 \\
only\_vla & ovr\_only\_vla\_regression & \normalfont 53.51 & \normalfont 40.65 & \cellcolor{green!54!red!25!white}\normalfont 54.23 \\
concat & kmeans\_concat\_regression & \normalfont 42.79 & \normalfont 26.96 & \cellcolor{green!54!red!25!white}\normalfont 54.12 \\
normalize\_concat & kmeans\_normalize\_concat\_regression & \normalfont 42.79 & \normalfont 26.96 & \cellcolor{green!54!red!25!white}\normalfont 54.12 \\
only\_image & ovr\_only\_image\_regression & \normalfont 53.53 & \normalfont 40.96 & \cellcolor{green!52!red!25!white}\normalfont 52.37 \\
only\_vla & kmeans\_only\_vla\_rank\_regression & \normalfont 41.65 & \normalfont 23.93 & \cellcolor{green!51!red!25!white}\normalfont 51.39 \\
weighted\_average\_text025 & kmeans\_weighted\_average\_regression & \normalfont 41.66 & \normalfont 25.97 & \cellcolor{green!51!red!25!white}\normalfont 50.89 \\
only\_image & kmeans\_only\_image\_regression & \normalfont 41.00 & \normalfont 24.75 & \cellcolor{green!49!red!25!white}\normalfont 49.10 \\
weighted\_average\_text025 & mlp\_weighted\_average\_h2048\_1024\_rank\_regression & \normalfont 53.00 & \normalfont 41.40 & \cellcolor{green!49!red!25!white}\normalfont 48.97 \\
only\_text & knn\_k5\_only\_text\_rank\_regression & \normalfont 53.18 & \normalfont 41.47 & \cellcolor{green!49!red!25!white}\normalfont 48.66 \\
only\_text & prknn\_k5\_only\_text\_rank\_regression & \normalfont 53.18 & \normalfont 41.47 & \cellcolor{green!49!red!25!white}\normalfont 48.66 \\
only\_image & ovr\_only\_image\_rank\_regression & \normalfont 53.35 & \normalfont 41.55 & \cellcolor{green!48!red!25!white}\normalfont 48.29 \\
only\_vla & linear\_only\_vla\_rank\_regression & \normalfont 52.96 & \normalfont 41.56 & \cellcolor{green!48!red!25!white}\normalfont 47.83 \\
only\_text & kmeans\_only\_text\_rank\_regression & \normalfont 40.11 & \normalfont 19.41 & \cellcolor{green!47!red!25!white}\normalfont 47.07 \\
average & kmeans\_average\_rank\_regression & \normalfont 40.08 & \normalfont 19.51 & \cellcolor{green!47!red!25!white}\normalfont 46.94 \\
weighted\_average\_text075 & kmeans\_weighted\_average\_rank\_regression & \normalfont 39.91 & \normalfont 18.94 & \cellcolor{green!46!red!25!white}\normalfont 46.45 \\
weighted\_average\_text075 & knn\_k5\_weighted\_average\_w0.75\_rank\_regression & \normalfont 53.20 & \normalfont 41.82 & \cellcolor{green!46!red!25!white}\normalfont 46.15 \\
weighted\_average\_text075 & prknn\_k5\_weighted\_average\_rank\_regression & \normalfont 53.20 & \normalfont 41.82 & \cellcolor{green!46!red!25!white}\normalfont 46.15 \\
only\_text & mlp\_only\_text\_h2048\_1024\_rank\_regression & \normalfont 53.61 & \normalfont 41.96 & \cellcolor{green!45!red!25!white}\normalfont 45.35 \\
only\_image & knn\_k5\_only\_image\_rank\_regression & \normalfont 53.20 & \normalfont 42.02 & \cellcolor{green!44!red!25!white}\normalfont 44.50 \\
only\_image & prknn\_k5\_only\_image\_rank\_regression & \normalfont 53.20 & \normalfont 42.02 & \cellcolor{green!44!red!25!white}\normalfont 44.50 \\
weighted\_average\_text075 & mlp\_weighted\_average\_h2048\_1024\_rank\_regression & \normalfont 53.44 & \normalfont 42.15 & \cellcolor{green!44!red!25!white}\normalfont 43.62 \\
average & knn\_k5\_average\_rank\_regression & \normalfont 53.29 & \normalfont 42.16 & \cellcolor{green!43!red!25!white}\normalfont 43.45 \\
average & prknn\_k5\_average\_rank\_regression & \normalfont 53.29 & \normalfont 42.16 & \cellcolor{green!43!red!25!white}\normalfont 43.45 \\
concat & knn\_k5\_concat\_rank\_regression & \normalfont 53.35 & \normalfont 42.41 & \cellcolor{green!41!red!25!white}\normalfont 41.28 \\
concat & prknn\_k5\_concat\_rank\_regression & \normalfont 53.35 & \normalfont 42.41 & \cellcolor{green!41!red!25!white}\normalfont 41.28 \\
normalize\_concat & knn\_k5\_normalize\_concat\_rank\_regression & \normalfont 53.35 & \normalfont 42.41 & \cellcolor{green!41!red!25!white}\normalfont 41.28 \\
normalize\_concat & prknn\_k5\_normalize\_concat\_rank\_regression & \normalfont 53.35 & \normalfont 42.41 & \cellcolor{green!41!red!25!white}\normalfont 41.28 \\
weighted\_average\_text025 & kmeans\_weighted\_average\_rank\_regression & \normalfont 38.32 & \normalfont 17.76 & \cellcolor{green!41!red!25!white}\normalfont 41.08 \\
only\_image & kmeans\_only\_image\_rank\_regression & \normalfont 38.26 & \normalfont 18.29 & \cellcolor{green!41!red!25!white}\normalfont 40.83 \\
concat & kmeans\_concat\_rank\_regression & \normalfont 38.23 & \normalfont 17.67 & \cellcolor{green!41!red!25!white}\normalfont 40.77 \\
normalize\_concat & kmeans\_normalize\_concat\_rank\_regression & \normalfont 38.23 & \normalfont 17.67 & \cellcolor{green!41!red!25!white}\normalfont 40.77 \\
weighted\_average\_text025 & mlp\_weighted\_average\_h2048\_1024\_regression & \normalfont 53.51 & \normalfont 42.79 & \cellcolor{green!38!red!25!white}\normalfont 37.63 \\
only\_text & mlp\_only\_text\_h2048\_1024\_regression & \normalfont 53.90 & \normalfont 42.83 & \cellcolor{green!37!red!25!white}\normalfont 37.41 \\
weighted\_average\_text025 & knn\_k5\_weighted\_average\_w0.25\_rank\_regression & \normalfont 53.51 & \normalfont 42.93 & \cellcolor{green!36!red!25!white}\normalfont 36.20 \\
weighted\_average\_text025 & prknn\_k5\_weighted\_average\_rank\_regression & \normalfont 53.51 & \normalfont 42.93 & \cellcolor{green!36!red!25!white}\normalfont 36.20 \\
weighted\_average\_text075 & mlp\_weighted\_average\_h2048\_1024\_regression & \normalfont 53.63 & \normalfont 43.12 & \cellcolor{green!34!red!25!white}\normalfont 34.11 \\
only\_vla & knn\_k5\_only\_vla\_rank\_regression & \normalfont 53.74 & \normalfont 43.41 & \cellcolor{green!31!red!25!white}\normalfont 30.67 \\
only\_vla & prknn\_k5\_only\_vla\_rank\_regression & \normalfont 53.74 & \normalfont 43.41 & \cellcolor{green!31!red!25!white}\normalfont 30.67 \\
only\_text & kmeans\_only\_text\_tie\_regression & \normalfont 29.02 & \normalfont 3.36 & \cellcolor{green!7!red!25!white}\normalfont 6.96 \\
weighted\_average\_text075 & kmeans\_weighted\_average\_tie\_regression & \normalfont 28.96 & \normalfont 3.34 & \cellcolor{green!7!red!25!white}\normalfont 6.74 \\
average & kmeans\_average\_tie\_regression & \normalfont 28.93 & \normalfont 3.39 & \cellcolor{green!7!red!25!white}\normalfont 6.63 \\
only\_vla & kmeans\_only\_vla\_tie\_regression & \normalfont 28.89 & \normalfont 4.33 & \cellcolor{green!6!red!25!white}\normalfont 6.47 \\
only\_image & kmeans\_only\_image\_tie\_regression & \normalfont 27.65 & \normalfont 7.00 & \cellcolor{green!2!red!25!white}\normalfont 1.64 \\
weighted\_average\_text025 & kmeans\_weighted\_average\_tie\_regression & \normalfont 27.49 & \normalfont 6.73 & \cellcolor{green!1!red!25!white}\normalfont 0.98 \\
concat & kmeans\_concat\_tie\_regression & \normalfont 27.40 & \normalfont 6.26 & \cellcolor{green!1!red!25!white}\normalfont 0.66 \\
normalize\_concat & kmeans\_normalize\_concat\_tie\_regression & \normalfont 27.40 & \normalfont 6.26 & \cellcolor{green!1!red!25!white}\normalfont 0.66 \\
\midrule
\multicolumn{5}{@{}l}{\normalfont\bfseries Fixed baselines and oracles}\\
\textrm{---} & expensive\_global & \normalfont 54.02 & \normalfont 45.10 & \cellcolor{green!0!red!25!white}\normalfont 0.00 \\
\textrm{---} & strongest\_global & \normalfont 54.02 & \normalfont 45.10 & \cellcolor{green!0!red!25!white}\normalfont 0.00 \\
\textrm{---} & strongest\_per\_category & \normalfont 54.02 & \normalfont 45.10 & \cellcolor{green!0!red!25!white}\normalfont 0.00 \\
\textrm{---} & random & \normalfont 40.02 & \normalfont 22.39 & \cellcolor{green!46!red!25!white}\normalfont 46.33 \\
\textrm{---} & cheapest\_global & \normalfont 27.24 & \normalfont 2.11 & \cellcolor{green!0!red!25!white}\normalfont 0.00 \\
\textrm{---} & accuracy\_oracle & \normalfont 55.17 & \normalfont 27.96 & \cellcolor{green!88!red!25!white}\normalfont 87.94 \\
\textrm{---} & $\eta$ oracle & \normalfont 55.17 & \normalfont 33.32 & \cellcolor{green!81!red!25!white}\normalfont 80.59 \\
\midrule
\multicolumn{5}{@{}l}{\normalfont\bfseries OOD baseline sweep --- all fusion methods $\times$ $\gamma\in\{0,0.1,\dots,1\}$}\\
concat & ood\_baseline\_success\_000 & \normalfont 28.29 & \normalfont 8.59 & \cellcolor{green!4!red!25!white}\normalfont 4.14 \\
concat & ood\_baseline\_success\_010 & \normalfont 28.29 & \normalfont 8.59 & \cellcolor{green!4!red!25!white}\normalfont 4.14 \\
concat & ood\_baseline\_success\_020 & \normalfont 28.42 & \normalfont 8.77 & \cellcolor{green!5!red!25!white}\normalfont 4.64 \\
concat & ood\_baseline\_success\_030 & \normalfont 34.03 & \normalfont 16.52 & \cellcolor{green!26!red!25!white}\normalfont 25.81 \\
concat & ood\_baseline\_success\_040 & \normalfont 37.04 & \normalfont 21.70 & \cellcolor{green!36!red!25!white}\normalfont 36.28 \\
concat & ood\_baseline\_success\_050 & \normalfont 49.83 & \normalfont 39.60 & \cellcolor{green!55!red!25!white}\normalfont 54.52 \\
concat & ood\_baseline\_success\_060 & \normalfont 49.83 & \normalfont 39.60 & \cellcolor{green!55!red!25!white}\normalfont 54.52 \\
concat & ood\_baseline\_success\_070 & \normalfont 49.83 & \normalfont 39.60 & \cellcolor{green!55!red!25!white}\normalfont 54.52 \\
concat & ood\_baseline\_success\_080 & \normalfont 53.66 & \normalfont 43.75 & \cellcolor{green!26!red!25!white}\normalfont 26.01 \\
concat & ood\_baseline\_success\_090 & \normalfont 53.66 & \normalfont 43.75 & \cellcolor{green!26!red!25!white}\normalfont 26.01 \\
concat & ood\_baseline\_success\_100 & \normalfont 54.02 & \normalfont 45.10 & \cellcolor{green!0!red!25!white}\normalfont 0.00 \\
normalize\_concat & ood\_baseline\_success\_000 & \normalfont 28.29 & \normalfont 8.59 & \cellcolor{green!4!red!25!white}\normalfont 4.14 \\
normalize\_concat & ood\_baseline\_success\_010 & \normalfont 28.29 & \normalfont 8.59 & \cellcolor{green!4!red!25!white}\normalfont 4.14 \\
normalize\_concat & ood\_baseline\_success\_020 & \normalfont 28.42 & \normalfont 8.77 & \cellcolor{green!5!red!25!white}\normalfont 4.64 \\
normalize\_concat & ood\_baseline\_success\_030 & \normalfont 34.03 & \normalfont 16.52 & \cellcolor{green!26!red!25!white}\normalfont 25.81 \\
normalize\_concat & ood\_baseline\_success\_040 & \normalfont 37.04 & \normalfont 21.70 & \cellcolor{green!36!red!25!white}\normalfont 36.28 \\
normalize\_concat & ood\_baseline\_success\_050 & \normalfont 49.83 & \normalfont 39.60 & \cellcolor{green!55!red!25!white}\normalfont 54.52 \\
normalize\_concat & ood\_baseline\_success\_060 & \normalfont 49.83 & \normalfont 39.60 & \cellcolor{green!55!red!25!white}\normalfont 54.52 \\
normalize\_concat & ood\_baseline\_success\_070 & \normalfont 49.83 & \normalfont 39.60 & \cellcolor{green!55!red!25!white}\normalfont 54.52 \\
normalize\_concat & ood\_baseline\_success\_080 & \normalfont 53.66 & \normalfont 43.75 & \cellcolor{green!26!red!25!white}\normalfont 26.01 \\
normalize\_concat & ood\_baseline\_success\_090 & \normalfont 53.66 & \normalfont 43.75 & \cellcolor{green!26!red!25!white}\normalfont 26.01 \\
normalize\_concat & ood\_baseline\_success\_100 & \normalfont 54.02 & \normalfont 45.10 & \cellcolor{green!0!red!25!white}\normalfont 0.00 \\
average & ood\_baseline\_success\_000 & \normalfont 27.24 & \normalfont 2.11 & \cellcolor{green!0!red!25!white}\normalfont 0.00 \\
average & ood\_baseline\_success\_010 & \normalfont 27.24 & \normalfont 2.11 & \cellcolor{green!0!red!25!white}\normalfont 0.00 \\
average & ood\_baseline\_success\_020 & \normalfont 27.24 & \normalfont 2.11 & \cellcolor{green!0!red!25!white}\normalfont 0.00 \\
average & ood\_baseline\_success\_030 & \normalfont 27.24 & \normalfont 2.11 & \cellcolor{green!0!red!25!white}\normalfont 0.00 \\
average & ood\_baseline\_success\_040 & \normalfont 27.24 & \normalfont 2.11 & \cellcolor{green!0!red!25!white}\normalfont 0.00 \\
average & ood\_baseline\_success\_050 & \normalfont 27.24 & \normalfont 2.11 & \cellcolor{green!0!red!25!white}\normalfont 0.00 \\
average & ood\_baseline\_success\_060 & \normalfont 27.24 & \normalfont 2.11 & \cellcolor{green!0!red!25!white}\normalfont 0.00 \\
average & ood\_baseline\_success\_070 & \normalfont 27.24 & \normalfont 2.11 & \cellcolor{green!0!red!25!white}\normalfont 0.00 \\
average & ood\_baseline\_success\_080 & \normalfont 27.24 & \normalfont 2.11 & \cellcolor{green!0!red!25!white}\normalfont 0.00 \\
average & ood\_baseline\_success\_090 & \normalfont 27.24 & \normalfont 2.11 & \cellcolor{green!0!red!25!white}\normalfont 0.00 \\
average & ood\_baseline\_success\_100 & \normalfont 54.02 & \normalfont 45.10 & \cellcolor{green!0!red!25!white}\normalfont 0.00 \\
weighted\_average\_text025 & ood\_baseline\_success\_000 & \normalfont 27.24 & \normalfont 2.11 & \cellcolor{green!0!red!25!white}\normalfont 0.00 \\
weighted\_average\_text025 & ood\_baseline\_success\_010 & \normalfont 27.24 & \normalfont 2.11 & \cellcolor{green!0!red!25!white}\normalfont 0.00 \\
weighted\_average\_text025 & ood\_baseline\_success\_020 & \normalfont 27.24 & \normalfont 2.11 & \cellcolor{green!0!red!25!white}\normalfont 0.00 \\
weighted\_average\_text025 & ood\_baseline\_success\_030 & \normalfont 27.24 & \normalfont 2.11 & \cellcolor{green!0!red!25!white}\normalfont 0.00 \\
weighted\_average\_text025 & ood\_baseline\_success\_040 & \normalfont 27.24 & \normalfont 2.11 & \cellcolor{green!0!red!25!white}\normalfont 0.00 \\
weighted\_average\_text025 & ood\_baseline\_success\_050 & \normalfont 27.24 & \normalfont 2.11 & \cellcolor{green!0!red!25!white}\normalfont 0.00 \\
weighted\_average\_text025 & ood\_baseline\_success\_060 & \normalfont 27.24 & \normalfont 2.11 & \cellcolor{green!0!red!25!white}\normalfont 0.00 \\
weighted\_average\_text025 & ood\_baseline\_success\_070 & \normalfont 27.24 & \normalfont 2.11 & \cellcolor{green!0!red!25!white}\normalfont 0.00 \\
weighted\_average\_text025 & ood\_baseline\_success\_080 & \normalfont 27.24 & \normalfont 2.11 & \cellcolor{green!0!red!25!white}\normalfont 0.00 \\
weighted\_average\_text025 & ood\_baseline\_success\_090 & \normalfont 27.24 & \normalfont 2.11 & \cellcolor{green!0!red!25!white}\normalfont 0.00 \\
weighted\_average\_text025 & ood\_baseline\_success\_100 & \normalfont 54.02 & \normalfont 45.10 & \cellcolor{green!0!red!25!white}\normalfont 0.00 \\
weighted\_average\_text075 & ood\_baseline\_success\_000 & \normalfont 27.24 & \normalfont 2.11 & \cellcolor{green!0!red!25!white}\normalfont 0.00 \\
weighted\_average\_text075 & ood\_baseline\_success\_010 & \normalfont 27.24 & \normalfont 2.11 & \cellcolor{green!0!red!25!white}\normalfont 0.00 \\
weighted\_average\_text075 & ood\_baseline\_success\_020 & \normalfont 27.24 & \normalfont 2.11 & \cellcolor{green!0!red!25!white}\normalfont 0.00 \\
weighted\_average\_text075 & ood\_baseline\_success\_030 & \normalfont 27.24 & \normalfont 2.11 & \cellcolor{green!0!red!25!white}\normalfont 0.00 \\
weighted\_average\_text075 & ood\_baseline\_success\_040 & \normalfont 27.24 & \normalfont 2.11 & \cellcolor{green!0!red!25!white}\normalfont 0.00 \\
weighted\_average\_text075 & ood\_baseline\_success\_050 & \normalfont 27.24 & \normalfont 2.11 & \cellcolor{green!0!red!25!white}\normalfont 0.00 \\
weighted\_average\_text075 & ood\_baseline\_success\_060 & \normalfont 27.24 & \normalfont 2.11 & \cellcolor{green!0!red!25!white}\normalfont 0.00 \\
weighted\_average\_text075 & ood\_baseline\_success\_070 & \normalfont 27.24 & \normalfont 2.11 & \cellcolor{green!0!red!25!white}\normalfont 0.00 \\
weighted\_average\_text075 & ood\_baseline\_success\_080 & \normalfont 28.13 & \normalfont 3.92 & \cellcolor{green!4!red!25!white}\normalfont 3.50 \\
weighted\_average\_text075 & ood\_baseline\_success\_090 & \normalfont 28.13 & \normalfont 3.92 & \cellcolor{green!4!red!25!white}\normalfont 3.50 \\
weighted\_average\_text075 & ood\_baseline\_success\_100 & \normalfont 54.02 & \normalfont 45.10 & \cellcolor{green!0!red!25!white}\normalfont 0.00 \\
only\_text & ood\_baseline\_success\_000 & \normalfont 27.24 & \normalfont 2.11 & \cellcolor{green!0!red!25!white}\normalfont 0.00 \\
only\_text & ood\_baseline\_success\_010 & \normalfont 27.24 & \normalfont 2.11 & \cellcolor{green!0!red!25!white}\normalfont 0.00 \\
only\_text & ood\_baseline\_success\_020 & \normalfont 27.24 & \normalfont 2.11 & \cellcolor{green!0!red!25!white}\normalfont 0.00 \\
only\_text & ood\_baseline\_success\_030 & \normalfont 27.24 & \normalfont 2.11 & \cellcolor{green!0!red!25!white}\normalfont 0.00 \\
only\_text & ood\_baseline\_success\_040 & \normalfont 27.24 & \normalfont 2.11 & \cellcolor{green!0!red!25!white}\normalfont 0.00 \\
only\_text & ood\_baseline\_success\_050 & \normalfont 27.24 & \normalfont 2.11 & \cellcolor{green!0!red!25!white}\normalfont 0.00 \\
only\_text & ood\_baseline\_success\_060 & \normalfont 27.24 & \normalfont 2.11 & \cellcolor{green!0!red!25!white}\normalfont 0.00 \\
only\_text & ood\_baseline\_success\_070 & \normalfont 27.24 & \normalfont 2.11 & \cellcolor{green!0!red!25!white}\normalfont 0.00 \\
only\_text & ood\_baseline\_success\_080 & \normalfont 37.28 & \normalfont 22.40 & \cellcolor{green!37!red!25!white}\normalfont 37.03 \\
only\_text & ood\_baseline\_success\_090 & \normalfont 37.28 & \normalfont 22.40 & \cellcolor{green!37!red!25!white}\normalfont 37.03 \\
only\_text & ood\_baseline\_success\_100 & \normalfont 54.02 & \normalfont 45.10 & \cellcolor{green!0!red!25!white}\normalfont 0.00 \\
only\_image & ood\_baseline\_success\_000 & \normalfont 27.24 & \normalfont 2.11 & \cellcolor{green!0!red!25!white}\normalfont 0.00 \\
only\_image & ood\_baseline\_success\_010 & \normalfont 27.24 & \normalfont 2.11 & \cellcolor{green!0!red!25!white}\normalfont 0.00 \\
only\_image & ood\_baseline\_success\_020 & \normalfont 27.24 & \normalfont 2.11 & \cellcolor{green!0!red!25!white}\normalfont 0.00 \\
only\_image & ood\_baseline\_success\_030 & \normalfont 27.24 & \normalfont 2.11 & \cellcolor{green!0!red!25!white}\normalfont 0.00 \\
only\_image & ood\_baseline\_success\_040 & \normalfont 27.24 & \normalfont 2.11 & \cellcolor{green!0!red!25!white}\normalfont 0.00 \\
only\_image & ood\_baseline\_success\_050 & \normalfont 27.24 & \normalfont 2.11 & \cellcolor{green!0!red!25!white}\normalfont 0.00 \\
only\_image & ood\_baseline\_success\_060 & \normalfont 27.24 & \normalfont 2.11 & \cellcolor{green!0!red!25!white}\normalfont 0.00 \\
only\_image & ood\_baseline\_success\_070 & \normalfont 27.24 & \normalfont 2.11 & \cellcolor{green!0!red!25!white}\normalfont 0.00 \\
only\_image & ood\_baseline\_success\_080 & \normalfont 28.63 & \normalfont 3.36 & \cellcolor{green!5!red!25!white}\normalfont 5.44 \\
only\_image & ood\_baseline\_success\_090 & \normalfont 28.63 & \normalfont 3.36 & \cellcolor{green!5!red!25!white}\normalfont 5.44 \\
only\_image & ood\_baseline\_success\_100 & \normalfont 54.02 & \normalfont 45.10 & \cellcolor{green!0!red!25!white}\normalfont 0.00 \\
only\_vla & ood\_baseline\_success\_000 & \normalfont 54.02 & \normalfont 45.10 & \cellcolor{green!0!red!25!white}\normalfont 0.00 \\
only\_vla & ood\_baseline\_success\_010 & \normalfont 54.02 & \normalfont 45.10 & \cellcolor{green!0!red!25!white}\normalfont 0.00 \\
only\_vla & ood\_baseline\_success\_020 & \normalfont 54.02 & \normalfont 45.10 & \cellcolor{green!0!red!25!white}\normalfont 0.00 \\
only\_vla & ood\_baseline\_success\_030 & \normalfont 54.02 & \normalfont 45.10 & \cellcolor{green!0!red!25!white}\normalfont 0.00 \\
only\_vla & ood\_baseline\_success\_040 & \normalfont 54.02 & \normalfont 45.10 & \cellcolor{green!0!red!25!white}\normalfont 0.00 \\
only\_vla & ood\_baseline\_success\_050 & \normalfont 54.02 & \normalfont 45.10 & \cellcolor{green!0!red!25!white}\normalfont 0.00 \\
only\_vla & ood\_baseline\_success\_060 & \normalfont 54.02 & \normalfont 45.10 & \cellcolor{green!0!red!25!white}\normalfont 0.00 \\
only\_vla & ood\_baseline\_success\_070 & \normalfont 54.02 & \normalfont 45.10 & \cellcolor{green!0!red!25!white}\normalfont 0.00 \\
only\_vla & ood\_baseline\_success\_080 & \normalfont 54.02 & \normalfont 45.10 & \cellcolor{green!0!red!25!white}\normalfont 0.00 \\
only\_vla & ood\_baseline\_success\_090 & \normalfont 54.02 & \normalfont 45.10 & \cellcolor{green!0!red!25!white}\normalfont 0.00 \\
only\_vla & ood\_baseline\_success\_100 & \normalfont 54.02 & \normalfont 45.10 & \cellcolor{green!0!red!25!white}\normalfont 0.00 \\
\midrule
\multicolumn{5}{@{}l}{\normalfont\bfseries Budget-oracle sweep --- $x\in\{0,5,\dots,100\}\%$}\\
\textrm{---} & accuracy\_oracle\_budget\_000pct & \normalfont 27.24 & \normalfont 2.11 & \cellcolor{green!0!red!25!white}\normalfont 0.00 \\
\textrm{---} & accuracy\_oracle\_budget\_005pct & \normalfont 28.34 & \normalfont 2.52 & \cellcolor{green!4!red!25!white}\normalfont 4.30 \\
\textrm{---} & accuracy\_oracle\_budget\_010pct & \normalfont 45.46 & \normalfont 13.97 & \cellcolor{green!66!red!25!white}\normalfont 65.83 \\
\textrm{---} & accuracy\_oracle\_budget\_015pct & \normalfont 51.99 & \normalfont 21.26 & \cellcolor{green!84!red!25!white}\normalfont 84.04 \\
\textrm{---} & accuracy\_oracle\_budget\_020pct & \normalfont 53.75 & \normalfont 24.05 & \cellcolor{green!87!red!25!white}\normalfont 87.45 \\
\textrm{---} & accuracy\_oracle\_budget\_025pct & \normalfont 54.56 & \normalfont 25.79 & \cellcolor{green!88!red!25!white}\normalfont 88.36 \\
\textrm{---} & accuracy\_oracle\_budget\_030pct & \normalfont 54.73 & \normalfont 26.41 & \cellcolor{green!88!red!25!white}\normalfont 88.27 \\
\textrm{---} & accuracy\_oracle\_budget\_035pct & \normalfont 55.10 & \normalfont 27.58 & \cellcolor{green!88!red!25!white}\normalfont 88.16 \\
\textrm{---} & accuracy\_oracle\_budget\_040pct & \normalfont 55.10 & \normalfont 27.58 & \cellcolor{green!88!red!25!white}\normalfont 88.16 \\
\textrm{---} & accuracy\_oracle\_budget\_045pct & \normalfont 55.17 & \normalfont 27.96 & \cellcolor{green!88!red!25!white}\normalfont 87.94 \\
\textrm{---} & accuracy\_oracle\_budget\_050pct & \normalfont 55.17 & \normalfont 27.96 & \cellcolor{green!88!red!25!white}\normalfont 87.94 \\
\textrm{---} & accuracy\_oracle\_budget\_055pct & \normalfont 55.17 & \normalfont 27.96 & \cellcolor{green!88!red!25!white}\normalfont 87.94 \\
\textrm{---} & accuracy\_oracle\_budget\_060pct & \normalfont 55.17 & \normalfont 27.96 & \cellcolor{green!88!red!25!white}\normalfont 87.94 \\
\textrm{---} & accuracy\_oracle\_budget\_065pct & \normalfont 55.17 & \normalfont 27.96 & \cellcolor{green!88!red!25!white}\normalfont 87.94 \\
\textrm{---} & accuracy\_oracle\_budget\_070pct & \normalfont 55.17 & \normalfont 27.96 & \cellcolor{green!88!red!25!white}\normalfont 87.94 \\
\textrm{---} & accuracy\_oracle\_budget\_075pct & \normalfont 55.17 & \normalfont 27.96 & \cellcolor{green!88!red!25!white}\normalfont 87.94 \\
\textrm{---} & accuracy\_oracle\_budget\_080pct & \normalfont 55.17 & \normalfont 27.96 & \cellcolor{green!88!red!25!white}\normalfont 87.94 \\
\textrm{---} & accuracy\_oracle\_budget\_085pct & \normalfont 55.17 & \normalfont 27.96 & \cellcolor{green!88!red!25!white}\normalfont 87.94 \\
\textrm{---} & accuracy\_oracle\_budget\_090pct & \normalfont 55.17 & \normalfont 27.96 & \cellcolor{green!88!red!25!white}\normalfont 87.94 \\
\textrm{---} & accuracy\_oracle\_budget\_095pct & \normalfont 55.17 & \normalfont 27.96 & \cellcolor{green!88!red!25!white}\normalfont 87.94 \\
\textrm{---} & accuracy\_oracle\_budget\_100pct & \normalfont 55.17 & \normalfont 27.96 & \cellcolor{green!88!red!25!white}\normalfont 87.94 \\
\end{longtable}}

\end{document}